\PassOptionsToPackage{table}{xcolor}

\documentclass{fairmeta}


\usepackage[utf8]{inputenc}

\usepackage{url}

\usepackage{amsmath,amssymb}
\usepackage{amsfonts}
\usepackage{bbm}

\usepackage{float}
\usepackage{tabularx}
\usepackage{makecell}

\usepackage{algorithm}
\usepackage{algorithmic}

\usepackage{enumitem}
\usepackage{nicefrac}
\usepackage{pifont}
\usepackage{fontawesome5}

\definecolor{forestgreen}{rgb}{0.13, 0.55, 0.13}
\definecolor{indiagreen}{rgb}{0.07, 0.53, 0.03}
\definecolor{darkred}{rgb}{0.75, 0.05, 0.05}
\definecolor{goldenyellow}{rgb}{0.95, 0.63, 0.05}

\definecolor{promptblue}{RGB}{232,242,255}
\definecolor{promptframe}{RGB}{62,113,196}
\definecolor{outputgreen}{RGB}{239,248,240}
\definecolor{outputframe}{RGB}{80,150,95}

\newcommand{\cmark}{\textcolor{forestgreen}{\faCheck}}
\newcommand{\xmark}{\textcolor{darkred}{\faTimes}}
\newcommand{\pmark}{\textcolor{goldenyellow}{\faIcon[regular]{circle}}}

\graphicspath{{figure/}}

\newtcolorbox{promptbox}[1]{
  colback=promptblue,
  colframe=promptframe,
  title={#1},
  fonttitle=\bfseries\sffamily,
  fontupper=\small\ttfamily,
  breakable,
  enhanced,
  sharp corners,
  boxrule=0.7pt,
  left=6pt,
  right=6pt,
  top=5pt,
  bottom=5pt
}

\newtcolorbox{outputbox}[1]{
  colback=outputgreen,
  colframe=outputframe,
  title={#1},
  fonttitle=\bfseries\sffamily,
  fontupper=\small\sffamily,
  breakable,
  enhanced,
  sharp corners,
  boxrule=0.7pt,
  left=6pt,
  right=6pt,
  top=5pt,
  bottom=5pt
}

\title{{\setstretch{1.08}\fontsize{17}{20}\selectfont AnnotateAnything: Automatic Annotation of 3D Assets for Robot Manipulation\par}}

\author[1,*]{Haoran Lu}
\author[1,*]{Mutian Shen}
\author[1]{Shuyang Yu}
\author[1]{Yu Xiao}
\author[3,4]{Songling Liu}
\author[1]{Jianshu Zhang}
\author[1]{Shang Wu}
\author[2]{Yue Chen}
\author[1]{Guo Ye}
\author[1]{Jiayi Wang}
\author[1]{Zhaoran Wang}
\author[1]{Han Liu}

\affiliation[1]{Northwestern~University}
\affiliation[2]{Peking~University}
\affiliation[3]{Institute~of~Automation,~Chinese~Academy~of~Sciences,~Beijing~100190}
\affiliation[4]{University~of~Chinese~Academy~of~Sciences}

\contribution[*]{Equal contribution}

\abstract{
Simulation enables scalable robot data collection, but raw 3D assets provide only geometry, lacking the semantic, interactive, and physical knowledge needed to specify where and how robots should act. In this work, we present \textbf{AnnotateAnything}, a general automatic annotation framework that converts passive 3D assets into manipulation-ready assets with structured, diverse, and executable manipulation labels. AnnotateAnything is built around two complementary pipelines. First, a unified visual-language annotation pipeline using vision-language reasoning to infer object semantics, interaction constraints, and 3D-grounded cues, providing human-prior guidance for identifying meaningful interaction regions. Second, a fully automatic and massively parallel physics annotation pipeline grounds these priors in each asset's geometry and physical constraints through candidate generation, geometry optimization and trajectory generation. This pipeline produces diverse and executable action annotations, including grasp poses, dexterous contacts, articulation waypoints, insertion directions, hanging affordances, and navigation targets. Using the generated annotations, we further build an asynchronous parallel simulation data-collection system across diverse objects, tasks, and robot embodiments. Experiments demonstrate that AnnotateAnything achieves superior annotation efficiency, data-collection efficiency, and task success rates over existing annotation and data-generation pipelines, while also supporting downstream tasks such as affordance detection, robotic VQA, and visual instruction finetuning. We provide project materials on the project page and plan to release the full code, annotations, and benchmark to facilitate future research. \textbf{Videos, code, demo assets, and annotations are provided in supplementary materials}. Project page: \texttt{https://tourmaline-caramel-169490.netlify.app/}.
}

\begin{document}

\maketitle

\section{Introduction}
\label{sec:intro}

\begin{figure}[htbp]
  \centering
  \includegraphics[width=0.98\linewidth]{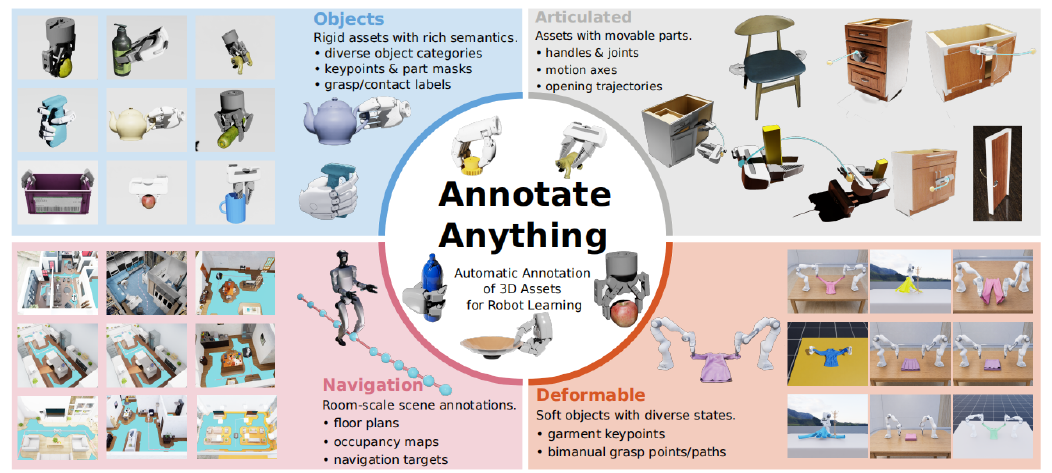}
  \caption{Overview of AnnotateAnything, a unified framework that converts raw 3D assets into manipulation-ready annotations across four asset families: rigid objects, articulated objects, deformable objects, and room-scale navigation scenes.}
  \label{fig:teaser}
\vspace{-2em}
\end{figure}
Training general-purpose robotic agents requires large-scale and diverse robot data across objects, embodiments and tasks~\cite{Intelligence2026pi07AS,Nvidia2025GR00TNA,generalist2025gen0,Ye2026WorldAM,Zheng2026EgoScaleSD,Intelligence202505AV,chen2026learningpartawaredense3d,11128651,Shen2025BiAssembleLC,Li2024BroadcastingSR,Li2026GarmentPileAC,Wu2024UniGarmentManipAU}. 
Simulation offers a scalable alternative by enabling automatic generation~\cite{Ye2025LearningTF,Geng2025RoboVerseTA,HaoranLuUniGarmentAU,Lu2024GarmentLabAU} and recent asset collections and simulation benchmarks provide rich geometric and visual substrates for such data generation~\cite{lightwheel_kitchen_2025,mittal2025isaac,Lightwheel_Team_LW-BenchHub_Lightwheel_s_End-to-End,Li2024BEHAVIOR1KAH}. 
Yet raw 3D assets typically encode only shape and appearance, lacking the semantic, interactive, and physical knowledge needed for robot learning, such as where to grasp, which parts can move, how objects articulate, and where insertion or hanging is feasible. 
Thus, the key bottleneck is not simply acquiring more assets, but automatically converting passive 3D geometry into manipulation-ready assets with actionable robot interaction annotations.

Generating manipulation annotations for raw 3D assets raises four key challenges. 
At the semantic level, annotations must encode human interaction priors so that generated trajectories are natural, safe, and likely to succeed; \textit{automatic annotation must incorporate human-prior reasoning and interaction understanding} \textbf{(C1)}~\cite{tang2025mindhandpurposefulrobotic,Qu2025SpatialVLAES}. 
These priors must be grounded in each asset's geometry and kinematics, since feasible actions depend on physical properties. \textit{Abstract interaction intent must be converted into executable, asset-specific labels} \textbf{(C2)}~\cite{niu2026scalable,Deshpande2026MolmoB0TLS}. 
Beyond correctness, annotations should preserve multiple valid solutions across interaction strategies; \textit{diverse feasible labels should not collapse to a single canonical solution} \textbf{(C3)}~\cite{Wang2026ManiTwinSD,chen2026univtac}. 
At scale, the process must cover many assets, categories, and interactions while retaining geometric precision and interaction specificity; \textit{this requires automation without per-asset manual design or task-specific engineering} \textbf{(C4)}.

Producing annotations that meet these requirements remains difficult: existing methods often rely on human annotation for semantic or physical knowledge~\cite{Chen2025RoboTwin2A,chen2026univtac,Chen2026RMBenchMR,Tian2025InternDataA1PH}, which is costly, labor-intensive, hard to scale, and limited in diversity, while RL-based alternatives reduce manual labeling but require reward engineering, complex training, and substantial computation, and may produce behaviors misaligned with human priors~\cite{wang2023robogen,wang2026mobilemanibench}. 
Therefore, we present \textbf{AnnotateAnything}, a general automatic annotation framework that converts 3D assets into manipulation-ready assets with actionable labels. 
AnnotateAnything provides complementary language annotations for interaction-level reasoning, visual annotations for 3D understanding, and action annotations specifying executable manipulation parameters such as grasp poses, articulation waypoints, and navigation trajectories.

To address these challenges, \textbf{AnnotateAnything} combines VLM-driven human-prior extraction with physics-grounded annotation. 
It first uses VLMs to bridge high-level language reasoning and visual grounding: language annotations provide multi-granularity descriptions from object semantics, while visual annotations translate these priors into 3D-grounded cues such as affordances, keypoints, and part segmentation. 
Together, these annotations localize meaningful interaction regions and guide downstream action-label generation, thereby injecting human interaction understanding into automatic annotation \textbf{(tackling C1)}. 
We then ground these priors in each asset through geometry-based optimization and asset-specific physical reasoning. 
To preserve diversity, AnnotateAnything generates annotations at multiple levels: within each task, it produces feasible candidates with different contacts, approach directions, orientations, and interaction configurations; across tasks, it covers grasping, dexterous manipulation, bimanual interaction, articulation, folding, hanging, and mobile manipulation; and across embodiments, it supports parallel-jaw grippers, dexterous hands, dual-arm systems, mobile manipulators, and humanoids \textbf{(tackling C3)}~\cite{ben2024homie,luo2025sonic}. 
Finally, AnnotateAnything enables scalability through a fully parallelized physics annotation pipeline covering candidate generation, geometry optimization, label selection, trajectory generation, physical validation, and data augmentation, without per-asset manual design or task-specific engineering. 
With CUDA-accelerated optimization and validation~\cite{curobo_v2}, the pipeline scales across assets and tasks while preserving asset-specific accuracy, interaction feasibility, and execution success \textbf{(tackling C4)}.

Based on the generated annotations, we set up an asynchronous parallel data-collection system for efficient data collection. Beyond data collection, our integrated visual-language-action annotations further support downstream applications, including affordance and keypoint detection~\cite{hou2024keygridunsupervised3dkeypoints,shi2021skeleton,Ning2023Where2ExploreFA}, robotics-oriented VQA and reasoning, and visual instruction finetuning for multimodal robot models. Under matched or framework-adapted evaluation protocols, experiments show that AnnotateAnything achieves substantially higher annotation efficiency, data-collection efficiency, and task success rates than the evaluated annotation or data-generation baselines, demonstrating the practical value of automatic, diverse, and asset-specific manipulation annotations.

In summary, our contributions are as follows:
\begin{itemize}[topsep=0pt,itemsep=1pt,parsep=0pt,partopsep=0pt,leftmargin=*]
    \item We present \textbf{AnnotateAnything}, a general automatic annotation framework that converts passive 3D assets into manipulation-ready assets through structured, actionable, and executable annotations.

    \item We design a \textbf{unified visual-language-action annotation pipeline} that integrates language reasoning, 3D visual grounding, and asset-specific action label generation, enabling executable manipulation specification across diverse objects, tasks, and robot embodiments.

    \item We develop a fully automatic and massively parallel \textbf{physics annotation pipeline}, covering candidate generation, trajectory generation and optimization, physical validation and augmentation.

    \item Under matched or framework-adapted evaluations, we demonstrate \textbf{superior annotation efficiency}, data-collection efficiency, and task success rates over the evaluated baselines, together with utility across several downstream tasks.
\end{itemize}

\section{Related Work}
\begin{table*}[t]
\centering
\caption{
Comparison of automatic data collection and annotation generation methods.
\cmark: explicitly supported or reported;
\pmark: partially supported;
\xmark: not supported or not reported.
}
\label{tab:related_work_comparison}
\scriptsize
\setlength{\tabcolsep}{2.25pt}
\renewcommand{\arraystretch}{1.15}
\resizebox{\textwidth}{!}{%
\begin{tabular}{l c c c c c c c c c c c c c}
\toprule
\multirow{2}{*}{\makecell[c]{\textbf{Method}}}
& \multicolumn{7}{c}{\textbf{Skill / Interaction Coverage}}
& \multicolumn{3}{c}{\textbf{Scenario / Embodiment}}
& \multicolumn{3}{c}{\textbf{Pipeline Property}} \\
\cmidrule(lr){2-8}
\cmidrule(lr){9-11}
\cmidrule(lr){12-14}
& \makecell{\textbf{Grasp}}
& \makecell{\textbf{DexGrasp}}
& \makecell{\textbf{BiGrasp}}
& \makecell{\textbf{BiDexGrasp}}
& \makecell{\textbf{Articulation}}
& \makecell{\textbf{Deformable}}
& \makecell{\textbf{Navigation}}
& \makecell{\textbf{Tabletop}}
& \makecell{\textbf{Mobile}}
& \makecell{\textbf{Humanoid}}
& \makecell{\textbf{Physics}\\\textbf{Validation}}
& \makecell{\textbf{Parallel}\\\textbf{Generation}}
& \makecell{\textbf{Data}\\\textbf{Augmentation}} \\
\midrule

\rowcolor{gray!8}
InternData-A1~\cite{Tian2025InternDataA1PH}
& \cmark
& \xmark
& \cmark
& \xmark
& \pmark
& \pmark
& \xmark
& \cmark
& \xmark
& \xmark
& \xmark
& \xmark
& \xmark \\

\rowcolor{gray!8}
RoboTwin2.0~\cite{Chen2025RoboTwin2A}
& \cmark
& \xmark
& \cmark
& \xmark
& \pmark
& \xmark
& \xmark
& \cmark
& \xmark
& \xmark
& \xmark
& \xmark
& \xmark \\

MolmoBot~\cite{Deshpande2026MolmoB0TLS}
& \cmark
& \xmark
& \xmark
& \xmark
& \pmark
& \xmark
& \pmark
& \cmark
& \pmark
& \xmark
& \cmark
& \xmark
& \cmark \\

\rowcolor{gray!8}
RoboGen~\cite{wang2023robogen}
& \cmark
& \xmark
& \xmark
& \xmark
& \cmark
& \xmark
& \xmark
& \cmark
& \pmark
& \xmark
& \pmark
& \cmark
& \xmark \\

GenManip~\cite{gao2025genmanip}
& \cmark
& \xmark
& \xmark
& \xmark
& \pmark
& \xmark
& \xmark
& \cmark
& \xmark
& \xmark
& \xmark
& \xmark
& \xmark \\

\rowcolor{gray!8}

GenieSim 3.0~\cite{yin2026geniesim30}
& \cmark
& \pmark
& \pmark
& \pmark
& \pmark
& \xmark
& \xmark
& \cmark
& \pmark
& \xmark
& \xmark
& \xmark
& \xmark \\

\rowcolor{indiagreen!10}
\textbf{AnnotateAnything}
& \textbf{\cmark}
& \textbf{\cmark}
& \textbf{\cmark}
& \textbf{\cmark}
& \textbf{\cmark}
& \textbf{\cmark}
& \textbf{\cmark}
& \textbf{\cmark}
& \textbf{\cmark}
& \textbf{\cmark}
& \textbf{\cmark}
& \textbf{\cmark}
& \textbf{\cmark} \\

\bottomrule
\end{tabular}%
}
\vspace{-0.5em}
\vspace{-1em}
\end{table*}

Table~\ref{tab:related_work_comparison} summarizes capabilities explicitly supported or reported by each system rather than a protocol-matched head-to-head benchmark. In particular, InternData-A1 is primarily a pretraining dataset, while RoboTwin2.0 and GenieSim 3.0 are task or data-generation platforms and do not expose the same per-asset annotation interface as AnnotateAnything.

\textbf{Automatic Data Collection for Robot Learning.}
Automatic robot data collection mainly follows annotation-based methods~\cite{Tian2025InternDataA1PH,Chen2025RoboTwin2A,Mu_2025_CVPR} or RL-based methods~\cite{wang2026mobilemanibench,wang2023robogen,yang2026ultradexgrasplearninguniversaldexterous,wan2023unidexgrasp++,wang2023dexgraspnet,bao2023dexart}. These approaches require manual task design, reward engineering, or costly training, whereas AnnotateAnything automatically turns raw 3D assets into reusable skill-consumable annotations for scalable data collection.

\textbf{Automatic Annotation for Manipulation.}
Prior automatic annotation methods often focus on task-specific labels such as grasps, contacts, affordances, or waypoints, limiting their scale, object diversity, embodiment coverage, or task generality~\cite{Wang2026ManiTwinSD,gao2025genmanip,yang2026ultradexgrasplearninguniversaldexterous,wan2023unidexgrasp++,chen2025bodexscalableefficientrobotic,lin2026bidexgraspcoordinatedbimanualdexterous,Wang2025ArticuBotLU,bao2023dexart}. AnnotateAnything unifies diverse skill-consumable labels across grasping, dexterous and bimanual manipulation, articulation, insertion, and hanging. BODex~\cite{chen2025bodexscalableefficientrobotic} focuses on scalable dexterous-grasp generation for a fixed hand interface and serves as the protocol-matched DexGrasp baseline in Sec.~\ref{sec:exp_physics_quality}. Additional related work is discussed in Appendix~\ref{app:related_work}.
\section{Method}
\label{sec:method}

AnnotateAnything takes a raw 3D asset, from a single object to a room, and converts it into a manipulation-ready asset with hierarchical language, visual, and action annotations. 
First, AnnotateAnything uses a visual-language annotation pipeline to inject human interaction priors and ground them in asset- and scene-level annotations (Secs.~\ref{sec:vl_pipeline}). 
Second, a physics-based action annotation pipeline converts these cues into executable annotations for rigid, articulated, deformable, garment, and room-scale interactions (Sec.~\ref{sec:action_generation_pipeline}), using a unified action schema, a generation-and-validation pipeline, and primitive-specific instantiations.

\subsection{Hierarchical Visual-Language Annotation Pipeline}
\label{sec:vl_pipeline}

\begin{figure}[t]
  \centering
  \includegraphics[width=0.98\linewidth]{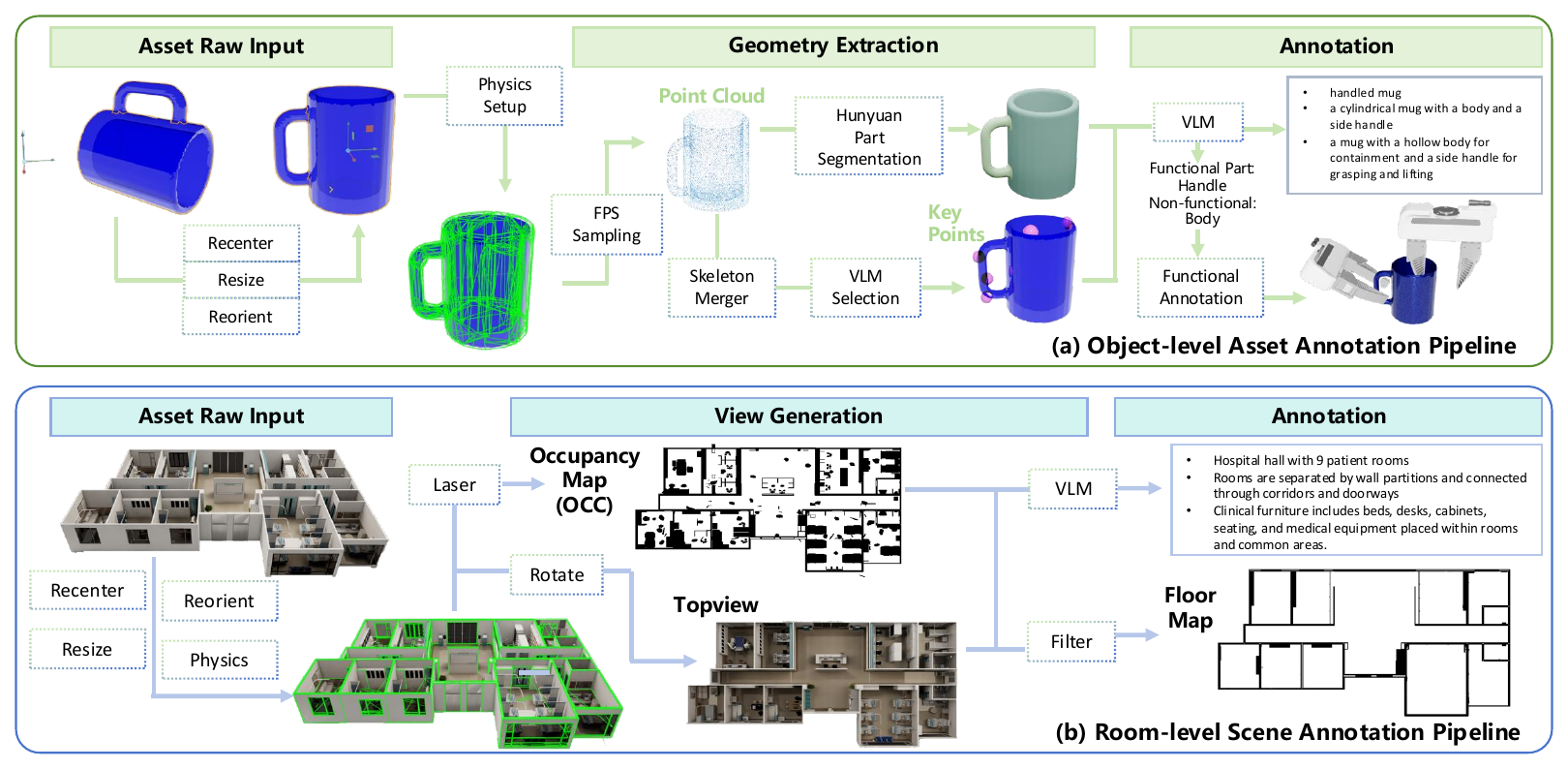}
  \caption{Hierarchical visual-language annotation pipeline. AnnotateAnything generates asset-level language and visual annotations, including descriptions, keypoints, part segmentation, and affordance regions, and composes them into room-level annotations.}
  \label{fig:vl-pipeline}
\vspace{-1.5em}
\end{figure}

\subsubsection{Hierarchical Language Annotation}
\label{sec:language_annotation}

\textbf{Asset-level language annotation.}
For each object asset, we feed multi-view RGB observations to Qwen3.5-VL~\cite{qwenteam2026qwen35omnitechnicalreport} to generate three-level descriptions: a semantic phrase, a functional sentence, and a part-aware paragraph. These annotations summarize object identity, manipulable parts, and feasible interactions as priors for visual grounding and action generation.

\textbf{Room-level language annotation.}
Given selected room views and a labeled top-view occupancy map, we query Qwen3.5-VL~\cite{qwenteam2026qwen35omnitechnicalreport} for three-level room descriptions: a layout summary, a furniture-and-zone description, and dense scene context. They capture object relations, navigable regions, and task-relevant interaction areas, complementing asset-level annotations.

\subsubsection{Hierarchical Visual Annotation}
\label{sec:visual_annotation}

\textbf{Asset-level visual annotation.}
For each object asset, we reconstruct a fused RGB-D point cloud and annotate semantic keypoints by VLM selection over FPS candidates. We obtain native 3D part proposals through Hunyuan3D-style decomposition with P3-SAM~\cite{ma2025p3sam} and X-Part~\cite{yan2025xpart}; the VLM assigns semantic labels, identifies task-relevant parts, and grounds affordance regions on these proposals. Together, the selected keypoints, part regions, and affordance regions provide 3D anchors for physics-based action annotation.

\textbf{Room-level visual annotation.}
For each room-scale scene, we construct multi-height occupancy maps from simulated LiDAR and ray-cast observations, and derive floor-plan and wall-structure annotations by filtering out movable objects. These visual annotations provide global geometric context for navigation, exploration, and scene-level action annotation.

\textbf{Cross-level Composition and Consistency.} For each room-scale scene, we isolate object instances, annotate each asset independently with object-centric annotation and then transform these annotations back to the global frame using scene instance identities. 
The resulting representation associates each room object with its semantic, visual, and action-relevant annotations, while aligning them with room-level annotation and descriptions through shared identifiers.
Structured-output parsing and segmentation-identifier checks reject malformed or ungrounded outputs, and anchors with unresolved interaction-defining attributes are not used for skill instantiation.

\subsection{Physics-based Action Annotation Pipeline}
\label{sec:action_generation_pipeline}

The physics-based action annotation pipeline grounds interaction priors in robot embodiment and physical feasibility, converting them into structured, executable, and validation-aware action labels.  It consists of candidate target generation, trajectory generation, trajectory optimization, physics validation, and physics-aware augmentation, producing manipulation-ready assets with diverse labels such as grasp poses, dexterous contacts, articulation waypoints, insertion directions, hanging poses, deformable-object trajectories, and navigation targets.

\begin{figure}[t]
  \centering
  \includegraphics[width=0.98\linewidth]{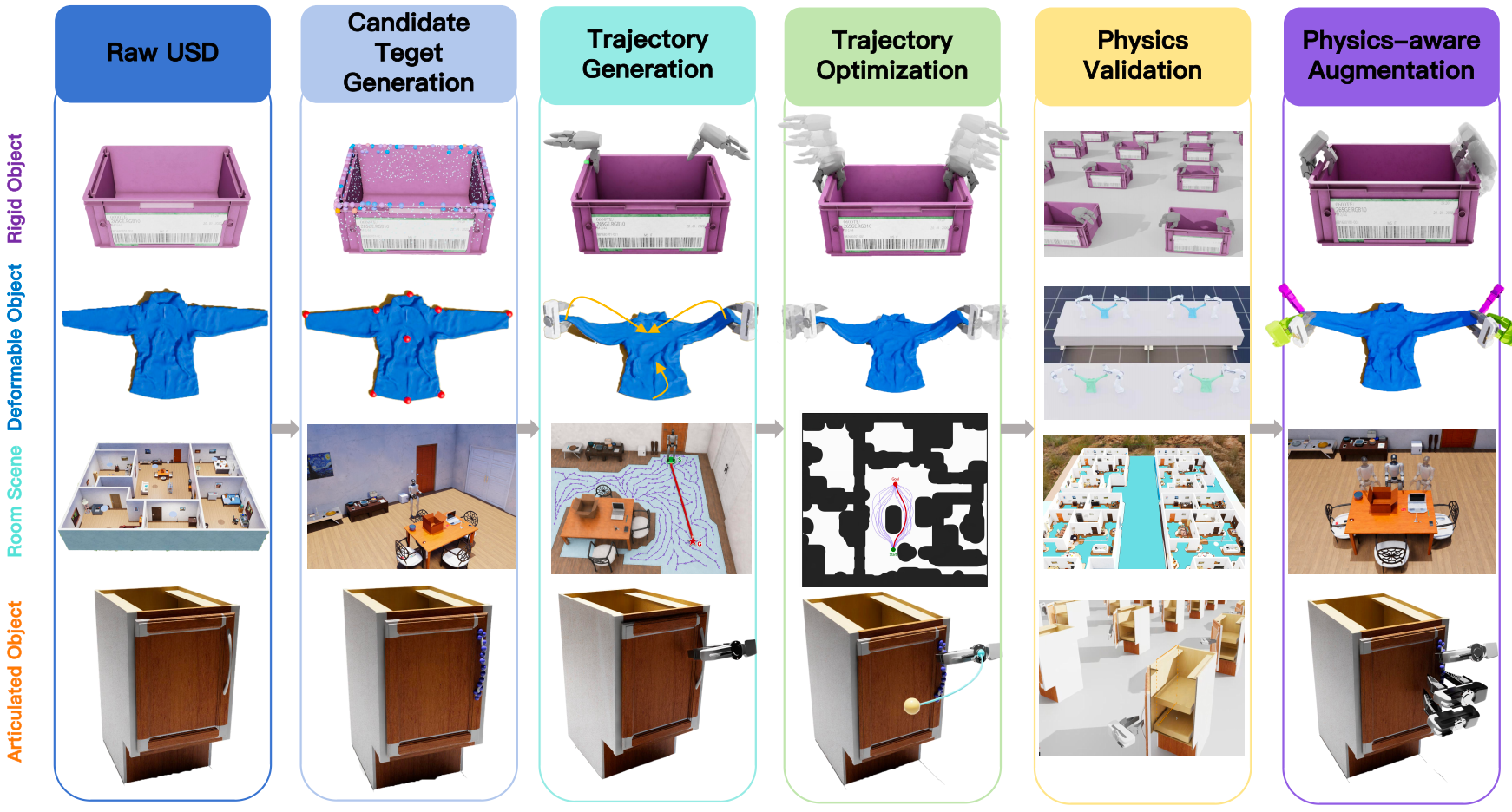}
  \caption{Physics-based action annotation pipeline. Grounded visual-language priors are converted into executable action annotations through candidate target generation, trajectory generation, trajectory optimization, physics validation, and physics-aware augmentation.}
  \label{fig:physics-annotation}
\vspace{-1.5em}
\end{figure}

\subsubsection{Unified Action Annotation Schema}
\label{sec:action_schema}

Our action schema preserves \textbf{action diversity} while maintaining \textbf{functional consistency}. 
Rather than assigning one canonical action to each object or part, we organize annotations as a hierarchical candidate bank built on visual annotations. 
Given an asset $\mathcal{A}$, the visual annotation stage provides grounded interaction anchors
\[
\mathcal{H}(\mathcal{A}) =
\mathcal{P}(\mathcal{A}) \cup
\mathcal{K}(\mathcal{A}) \cup
\mathcal{R}^{\mathrm{aff}}(\mathcal{A}) \cup
\mathcal{R}^{\mathrm{scene}}(\mathcal{A}),
\]
where $\mathcal{P}$, $\mathcal{K}$, $\mathcal{R}^{\mathrm{aff}}$, and $\mathcal{R}^{\mathrm{scene}}$ denote part regions, semantic keypoints, affordance regions, and scene-level regions. 
Each anchor $h$ is associated with a functional affordance $\phi(h)$, so local anchors define manipulation targets while scene-level anchors define navigation or approach targets.

For each $h \in \mathcal{H}(\mathcal{A})$, we infer compatible skills $\mathcal{S}(h)$ from $\phi(h)$, language priors, geometry, and task constraints. 
This ensures that an action is not only physically feasible but also supports the intended downstream interaction, e.g., grasping a mug handle for pouring or a cabinet handle for opening. 
For every anchor-skill pair, we maintain
\[
\mathcal{B}_{h,s}=\{a_{h,s}^{(i)}\}_{i=1}^{N_{h,s}},\quad s\in\mathcal{S}(h),\qquad
a_{h,s}^{(i)}=(s,o(h),h,\phi(h),x^{(i)},\theta^{(i)},\tau^{(i)},v^{(i)}, ,d^{(i)}).
\]
Here, $N_{h,s}$ is the number of feasible candidates; $s$ is the skill type; $o(h)$ is the associated object; and $\phi(h)$ conditions generation on functional affordance. 
The target $x^{(i)}$ instantiates the anchor as a concrete interaction target, while $\theta^{(i)}$ stores skill-specific parameters such as poses, contacts, directions, or embodiment states. 
The optional trajectory $\tau^{(i)}$ stores waypoints, while $v^{(i)}$ records feasibility metadata including collision, task success, and trajectory statistics.
The auxiliary descriptor $d^{(i)}$ records diversity attributes such as
target location, approach direction, contact mode, trajectory family, or
embodiment profile.

This schema defines action annotation as a \textbf{one-to-many mapping} from grounded anchors to function-conditioned executable candidates. 
Since each $\mathcal{B}_{h,s}$ stores multiple validated candidates across poses, contacts, directions, and trajectories, AnnotateAnything provides downstream modules with diverse physically feasible and functionally meaningful labels instead of a single heuristic annotation.

\subsubsection{General Physics-based Action Generation Pipeline}
\label{sec:general_action_pipeline}

\textbf{Candidate target generation.}
Given a skill type $s$, we localize skill-compatible regions $\mathcal{C}_s$ using language and visual grounding for object-centric skills, and a navigation mesh for room-level navigation. 
For example, functional grasping selects task-relevant grasp regions, articulation selects handles or movable parts, and navigation samples reachable regions around the target object on the navigation mesh. 
We then sample concrete targets from $\mathcal{C}_s$ via FPS on object or navmesh, instantiating targets such as contact points, articulated handle points, and interaction-ready base poses. 
The sampled targets are filtered by geometric and embodiment constraints, including curvature, visibility, collision margins, reachability, and traversability, before being used to populate the candidate banks.

\textbf{Trajectory generation.}
Given a retained target $x^{(i)}$ for skill $s$, we generate an initial waypoint sequence $\tau_{0}^{(i)}=\mathcal{G}_{s}\!\left(h,\phi(h),x^{(i)},\theta^{(i)};\mathcal{A}\right)$, where $\mathcal{G}_{s}$ is selected by the skill's constraint source. 
Object-property-conditioned generators use static geometry or garment keypoints for grasping, dexterous grasping, fling, and folding; object-trajectory-conditioned generators follow articulated motion while preserving the end-effector--part relation; object-vector-conditioned generators align actions with insertion, hanging, pouring, or placement directions; and scene-trajectory-conditioned generators sample interaction-ready base poses on the navigation mesh and compute coarse A* paths. 
These trajectories serve as template seeds for optimization and validation.

\textbf{Trajectory optimization.}
We refine each template by optimizing action parameters and waypoint sequences under task, contact, geometry, embodiment, and smoothness constraints, $(\theta_*^{(i)},\tau_*^{(i)})=\arg\min_{\theta,\tau} E_{\mathrm{task}}+E_{\mathrm{contact}}+E_{\mathrm{coll}}+E_{\mathrm{kin}}+E_{\mathrm{smooth}}$. 
Here, the terms preserve functional affordances, improve contact stability, penalize collision and unsafe clearance, enforce robot kinematics, and regularize motion. 
For dexterous grasping, we optimize hand pose, finger configuration, and contact assignment on grounded functional regions; for adaptive skills, we adjust waypoints to asset geometry such as sleeve length or hem-to-shoulder distance; and for navigation, the A* path is locally optimized with DWB to satisfy embodiment dynamics, turning constraints, and manipulation reachability.

\textbf{Physics validation.}
Optimized candidates are validated with parallel physics simulation. 
For most object-centric skills, we use floating end-effector, e.g., a floating gripper or dexterous hand, to test contact, collision, and stability without tying to a specific robot pose. 
For strongly embodiment-dependent skills, such as garment folding and entangled-object retrieval, we validate with the full manipulator to capture arm collision and entanglement constraints. 
We further apply disturbances and vary gravity magnitude or direction to test robustness, especially for grasp candidates. 
Floating validation serves as an object-centric candidate filter rather than the final embodied execution test; retained candidates are subsequently evaluated through embodiment-specific inverse kinematics, collision-aware planning, and full-robot execution as described in Sec.~\ref{sec:downstream-data-collection}.

\textbf{Physics-aware augmentation.}
To increase diversity, we augment validated candidates with \textbf{local perturbation} and \textbf{symmetry-aware augmentation}. 
Local perturbation samples bounded variations around anchors and action parameters, such as contact centers while preserving the same functional affordance $\phi(h)$ and satisfying object-robot constraints. 
Symmetry-aware augmentation expands candidates using annotated object symmetries by consistently transforming targets, pose parameters, and trajectory waypoints, e.g., rotating grasps around a bottle's symmetry axis. 
Perturbation- and symmetry-derived candidates are retained only after
passing the corresponding geometry and physics validation checks.

\section{Downstream Tasks}

\label{sec:downstream}
\begin{figure}[htbp]
  \centering
  \includegraphics[width=0.98\linewidth]{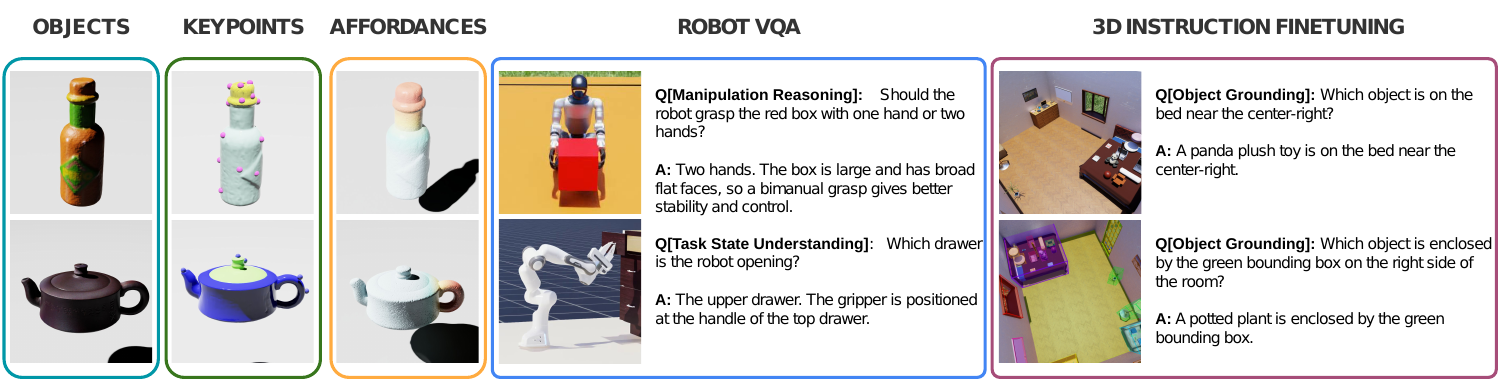}
  \caption{Downstream applications supported by AnnotateAnything. The same visual-language-action annotations can be converted into supervision for affordance and keypoint detection, robot reasoning VQA, and 3D VLM instruction finetuning. Affordance and keypoint supervision is quantitatively evaluated, while VQA and instruction tuning are presented as illustrative repurposings.}
  \label{fig:downstream-task}

\end{figure}

AnnotateAnything turns passive 3D assets into reusable visual-language-action annotations for robot learning. 
As shown in Fig.~\ref{fig:downstream-task}, we study two downstream uses: using executable action annotations for large-scale simulation data collection, and converting the same annotations into supervision for affordance and keypoint detection, robot reasoning VQA, and 3D VLM instruction finetuning. We quantitatively evaluate the perception uses, including an additional matched evaluation of part annotations, while robot reasoning VQA and 3D VLM instruction finetuning are presented as illustrative repurposings.

\begin{figure}[htbp]
  \centering
  \includegraphics[width=0.98\linewidth]{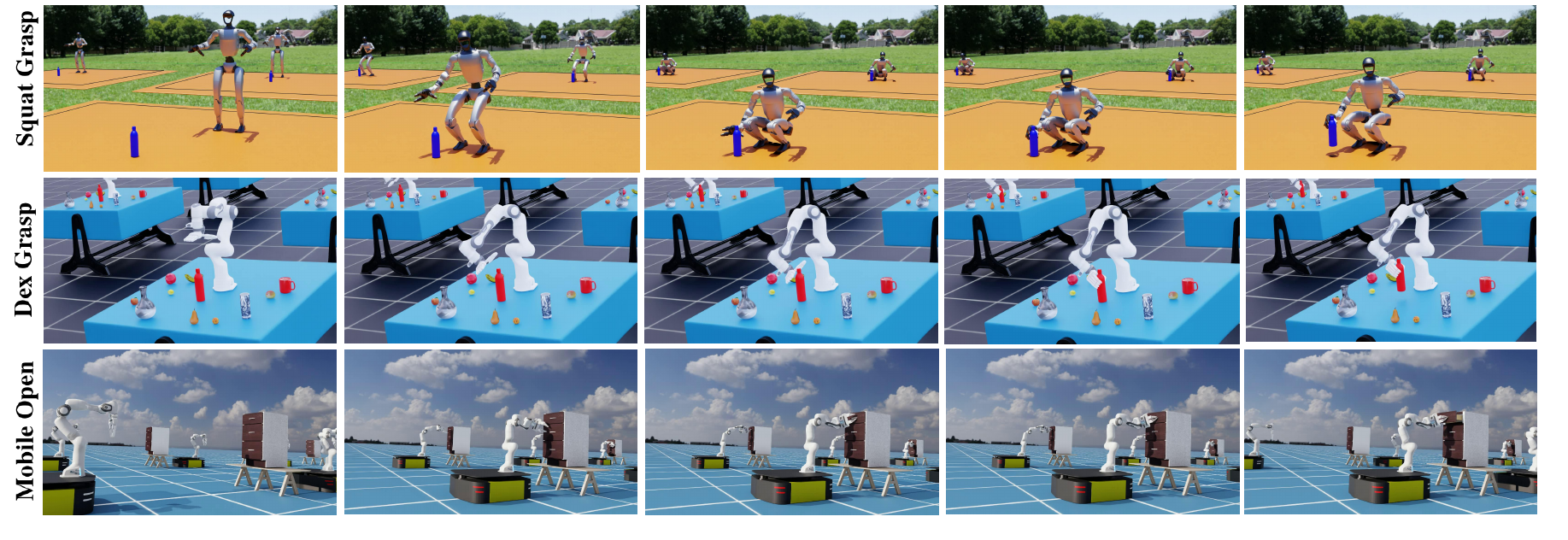}
  \caption{Representative atomic skills: covering tabletop, bimanual, whole-body, dexterous, and mobile manipulation.}
  \label{fig:downstream-atomic-skill}
  \vspace{-1.0em}
\end{figure}

\subsection{Large-scale Robot Data Collection}

\label{sec:downstream-data-collection}
We use the generated action annotations as executable interfaces for large-scale simulation-based robot data collection. Although this system is not the core contribution of the paper, we include it to demonstrate that AnnotateAnything annotations can be directly used for downstream data-collection pipelines.

\textbf{Atomic skill interface.}
We build an atomic-skill library aligned with our annotation format, where skills consume action parameters such as grasp poses, target parts, waypoints, insertion directions, hanging anchors, and navigation goals. 
The library supports tabletop manipulation, bimanual manipulation, whole-body control, humanoids, dexterous hands, and mobile manipulation, and can compose atomic skills into long-horizon tasks.

\textbf{Parallel rollout generation.}
We run heterogeneous parallel simulation environments, where each environment independently samples assets, tasks, object poses, and scene layouts. 
For each rollout, asynchronous cuRobo-v2 planning is used for goal reaching and obstacle avoidance~\cite{curobo_v2,sundaralingam2023curoboparallelizedcollisionfreeminimumjerk}, with domain randomization over pose, lighting, material, camera, and scene configuration.

\textbf{Trajectory validation.}
For each randomized scene, we retrieve ranked candidates from the annotation bank. For each candidate, we solve goal-set IK using the complete task-compatible robot embodiment and select the feasible IK solution with the lowest planning cost. If the highest-ranked retained candidate is infeasible under full-robot IK or planning constraints, we evaluate subsequent stored candidates until a feasible solution is found or the candidate set is exhausted.
We also remove action candidates that repeatedly fail simulation validation, ensuring that the retained annotations remain executable under diverse robot--object configurations.

\subsection{Annotation-derived Downstream Applications}
\label{sec:downstream-applications}

\textbf{Keypoint Generation and Affordance Grounding.}
We reuse visual annotations as keypoint supervision and physics-validated action annotations as affordance supervision. 
Visual labels provide VLM-selected 3D functional anchors, while action labels provide executable cues such as grasp directions, articulation trajectories, garment pick-and-place points, and navmesh-grounded interaction regions. 
These labels can be projected to images or represented on point clouds for robot-centric perception training~\cite{chen2026pa3ff,wu2024unigarmentmanip,ning2023where2explore,11128651,Li2024BroadcastingSR,li2023imagemanip}.

\textbf{Validated downstream evaluation.}
We keep each established downstream pipeline fixed and replace only its original supervision or input with AnnotateAnything outputs. On a matched $247$-object PartNeXt subset~\cite{wang2025partnext}, SAMPart3D~\cite{yang2024sampart3d} obtains $43.9$ object-macro mIoU with our part annotations, compared with $46.70$ using reference supervision. On a matched $200$-object LASO subset~\cite{liu2024laso}, the released LASO pipeline obtains $40.7/54.1$ mIoU/F1 with our affordance annotations, compared with $43.6/57.2$ using reference supervision. Skeleton Merger~\cite{shi2021skeleton} obtains $74.5/73.6$ DAS on the KeypointNet~\cite{you2020keypointnet} Airplane/Chair categories using our keypoint annotations, compared with $77.7/76.8$ using reference supervision. These results demonstrate downstream usability with modest gaps to reference supervision, rather than parity.

\textbf{Robot Reasoning VQA.}
As an illustrative repurposing, robot reasoning VQA is generated as a by-product of simulation rollouts.
During execution, we record grounded labels such as the selected object, part anchor, IK goal-set solution, and approach side. 
Because these labels depend on the runtime robot--object pose and the selected candidate from the annotation bank, they provide execution-level QA supervision that cannot be derived from static assets alone~\cite{zhang2026spacenum,zhang2026progresslm,pan2025advevo_marl}.

\textbf{3D VLM Instruction-tuning Data.}
As another illustrative repurposing, AnnotateAnything can be converted into lightweight instruction-tuning data for 3D VLMs.
Simulation assets and rollouts provide dense grounding labels, including 3D boxes, projected 2D boxes, and part segmentations, while language annotations and cross-level composition provide object relations, room layouts, and task contexts. 
Together, these signals yield grounding-style instruction--response pairs for 3D spatial reasoning, object localization, and scene-level QA, serving as a downstream demonstration rather than a full 3D VLM benchmark.
\section{Experiment}

\subsection{Experimental Setup}
\label{sec:exp_setup}

\textbf{Evaluation scope.}
We evaluate AnnotateAnything as an automatic annotation pipeline for converting heterogeneous raw 3D assets into high-quality language, visual, and action annotations for robot manipulation. 
Rather than treating it as a dataset release or a standalone data-generation system, our experiments assess annotation quality from four aspects: scale and conversion, visual-language and visual annotation quality, physics-grounded executability, and downstream rollout utility.

\textbf{Audited evaluation suite.}
All simulation-based action-quality and rollout success-rate results are reported on an audited evaluation suite sampled from a larger heterogeneous asset pool.
The pool spans multiple asset sources and asset types, including rigid objects, articulated objects, deformable or garment assets, and room-scale scenes. 
Since our goal is annotation conversion rather than curated asset release, we report source and scale statistics separately, while simulation pass, readiness, execution, and rollout success rates, together with action and trajectory human scores, are computed on the audited suite.
The suite contains $1{,}160$ assets and $3{,}698$ valid asset--skill pairs, including $480/1{,}700$ rigid assets/pairs, $320/740$ articulated assets/pairs, $280/760$ deformable or garment assets/pairs, and $80/498$ room-scale assets/pairs. It was sampled before annotation outcomes were computed and manually checked only for asset validity, simulator compatibility, and independently defined skill applicability, without success- or failure-based filtering. The suite is stratified by valid asset--skill pairs, covering grasping, dexterous grasping, bimanual grasping, bimanual dexterous grasping, articulation, insertion, hanging, deformable manipulation, and navigation or approach-target generation. Visual-language quality is evaluated on a separate balanced subset of $360$ manually referenced examples, with $90$ examples from each asset family.

\textbf{Human evaluation protocol.}
Generation-based evaluations use three independent generation seeds. Each human-evaluated item is independently scored by five non-author raters under blinded and randomized presentation. The complete sampling procedure, rating protocol, inter-rater agreement, and uncertainty estimation are provided in Appendices~H.1 and H.9.

\subsection{Annotation Coverage over Heterogeneous Assets}
\label{sec:exp_coverage}

\begin{figure*}[htbp]
  \centering
  \begin{minipage}[t]{0.44\textwidth}
    \centering
    \includegraphics[width=\linewidth]{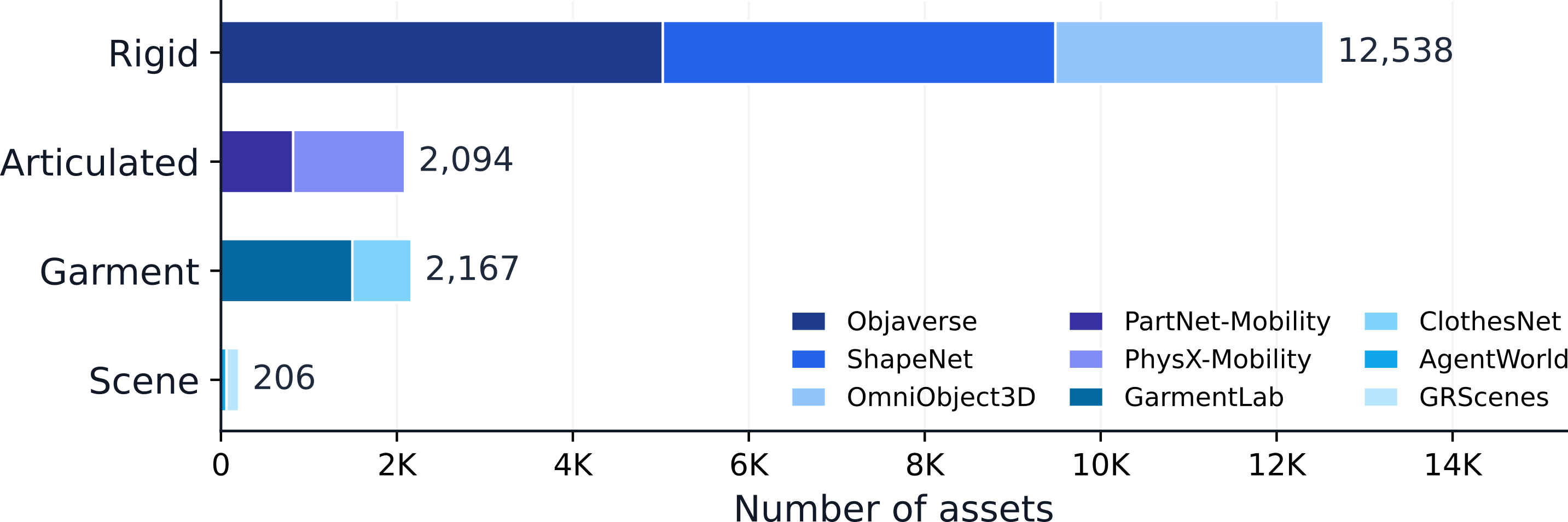}\\[-0.25em]
    {\scriptsize \textbf{(A)} Processed asset pool by asset type and source.}
  \end{minipage}\hfill
  \begin{minipage}[t]{0.39\textwidth}
    \centering
    \includegraphics[width=\linewidth]{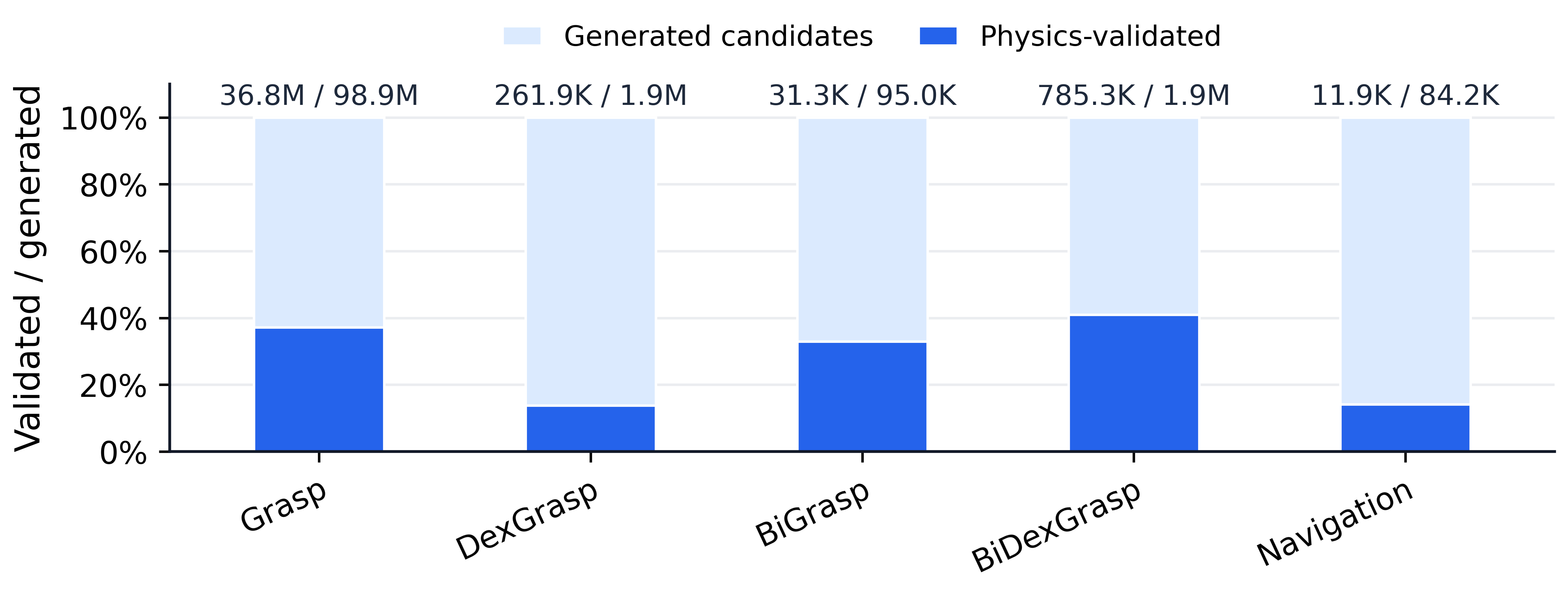}\\[-0.25em]
    {\scriptsize \textbf{(B)} Physics-validated candidates by skill family, with raw counts shown above each bar.}
  \end{minipage}\hfill
  \begin{minipage}[t]{0.16\textwidth}
    \centering
    \includegraphics[width=\linewidth]{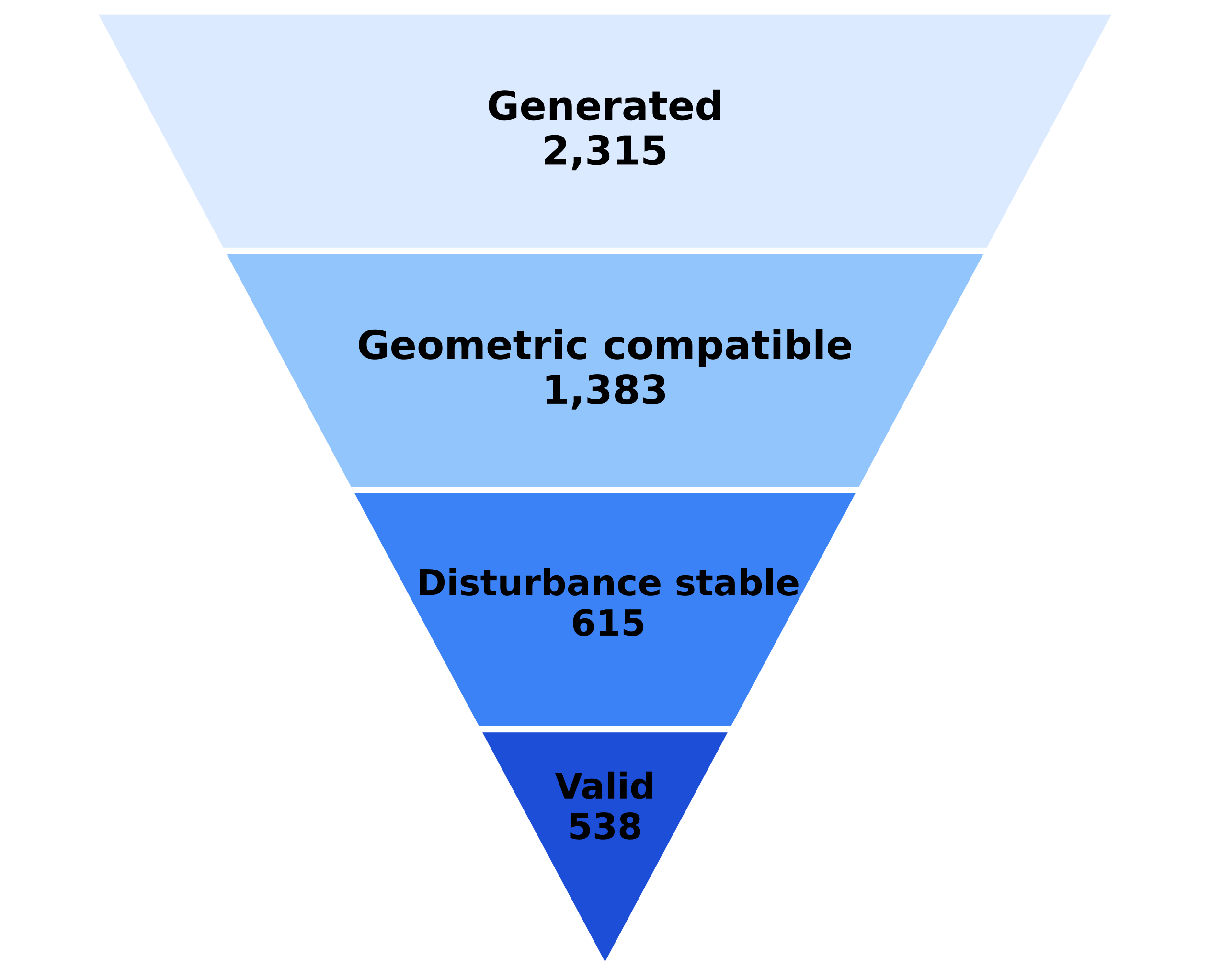}\\[-0.25em]
    {\scriptsize \textbf{(C)} Conversion funnels}
  \end{minipage}
  \caption{
  \textbf{Annotation coverage over heterogeneous assets.}
  Condensed coverage statistics:
  (A) processed asset pool grouped by asset type and source family,
  (B) physics validation outcomes grouped by skill family,
  and (C) candidate-to-annotation conversion, showing how generated candidates are filtered into feasible, physics-validated, and retained action annotations.
  }
  \label{fig:annotation_coverage}
  \vspace{-1em}
\end{figure*}

\begin{figure*}[t]
  \centering
  \includegraphics[width=0.98\linewidth]{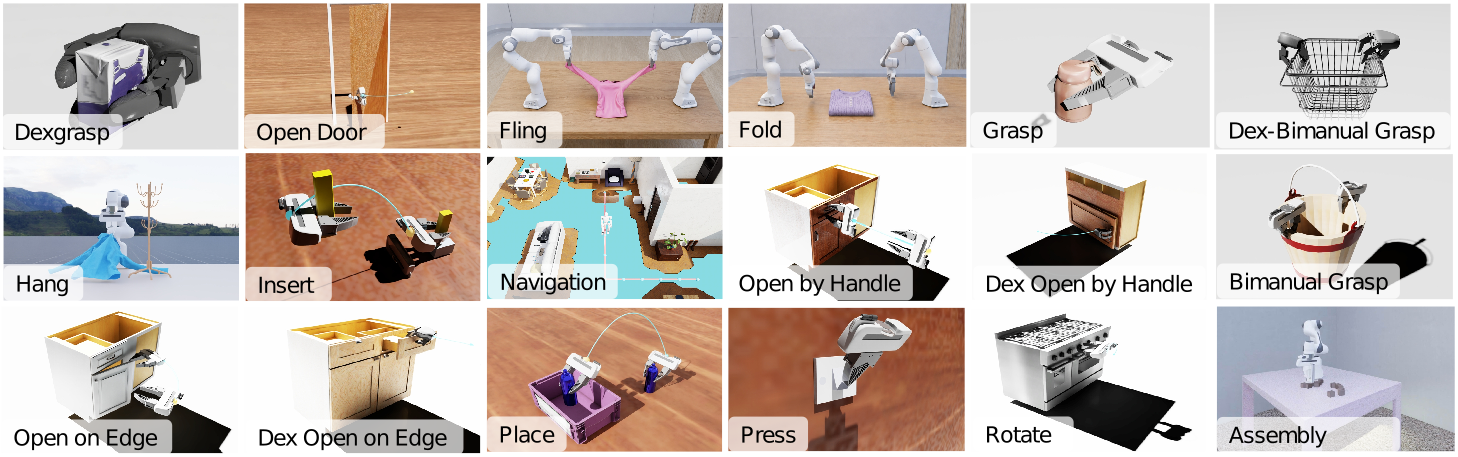}
  \caption{Atomic skill taxonomy with qualitative annotation examples for the 18 atomic skills.}
  \label{fig:atomic_skill_taxonomy}
  \vspace{-1em}
\end{figure*}

We first evaluate whether raw assets from diverse sources can be converted into manipulation-ready annotation banks. 
As shown in Fig.~\ref{fig:annotation_coverage}, we process $17{,}005$ assets from $9$ source families, spanning rigid objects, articulated objects, deformable or garment assets, and room-scale scenes. 
AnnotateAnything produces $100$M physics-validated action annotations over $18$ atomic skills, grouped in Fig.~\ref{fig:atomic_skill_taxonomy}. The count includes original candidates and validated perturbation- and symmetry-derived variants, while failed raw proposals are excluded; it therefore measures validated annotation instances rather than $100$M distinct manipulation strategies.
On the audited evaluation suite, the pipeline generates $2{,}315$ candidates per attempted asset--skill pair on average; $615$ pass physics validation and $538$ are retained in the final annotation bank, corresponding to a $26.6\%$ physics validation pass rate and a $23.2\%$ retained-candidate rate. 
Macro-averaged pass rates, readiness rates, execution success rates, and annotation throughput are reported in Table~\ref{tab:physics_quality_and_collection}.

\paragraph{Candidate-bank diversity.}
To quantify candidate diversity beyond raw candidate counts, we cluster candidates within each asset--skill pair. On independently sampled sets of $4{,}185$ Grasp, $400$ DexGrasp, and $300$ Articulation assets, the median numbers of distinct candidate clusters per asset are $176$, $153$, and $21$, respectively. Clustering criteria and complete diversity statistics are provided in Appendix~H.2.

\subsection{Visual-Language Annotation Quality}
\label{sec:exp_vl_quality}

The visual-language stage provides semantic and spatial priors for physics annotation. 
We evaluate it on the separate balanced $360$-example manual-reference
set described in Appendix~H.1, with references for language descriptions,
functional parts, keypoints, affordance regions, and scene-level spatial
cues~\cite{mo2019partnet,wang2025partnext,you2020keypointnet,liu2024laso}.
We evaluate complete visual-language bundles using human ratings on a $0$--$100$ scale, reported as mean $\pm$ standard error, and compare against three variants: \textbf{Direct VLM}, \textbf{Aff.-only + heur.}, and \textbf{w/o 3D refine}. 
We also evaluate generated part, affordance, and keypoint annotations using aggregate matching scores against manual references derived from the corresponding annotation sources~\cite{wang2025partnext,liu2024laso,you2020keypointnet}.

\begin{table*}[htbp]
\centering
\caption{
Visual-language and visual annotation quality on the separate $360$-example manual-reference set. 
The left block reports mean $\pm$ standard error human ratings for complete visual-language annotation bundles. 
The right block reports aggregate matching scores for generated part, affordance, and keypoint annotations under a unified evaluation protocol.
}
\label{tab:vl_quality}
\scriptsize
\setlength{\tabcolsep}{2.2pt}
\renewcommand{\arraystretch}{1.08}

\begin{tabular}{lccccc@{\hspace{0.8em}}ccc}
\toprule
\multicolumn{6}{c}{\textbf{Visual-Language Bundle Quality}} 
& \multicolumn{3}{c}{\textbf{Visual Annotation Quality}} \\
\cmidrule(lr){1-6}
\cmidrule(lr){7-9}
Method 
& Semantic 
& \begin{tabular}[c]{@{}c@{}}3D\\Grounding\end{tabular}
& Coverage 
& Action
& Overall 
& Part 
& Afford. 
& Keypoint \\
\midrule

Direct VLM 
& $84.2{\pm}1.1$ 
& $61.5{\pm}1.9$ 
& $68.7{\pm}1.7$ 
& $64.1{\pm}1.8$ 
& $69.6{\pm}1.4$ 
& 72.8
& 68.5
& 70.2 \\

\begin{tabular}[c]{@{}l@{}}Aff.-only\\+ heur.\end{tabular}
& $76.8{\pm}1.5$ 
& $70.2{\pm}1.8$ 
& $73.1{\pm}1.6$ 
& $67.5{\pm}1.7$ 
& $71.9{\pm}1.4$ 
& 70.5
& 79.3
& 68.8 \\

w/o 3D refine 
& $87.5{\pm}0.9$ 
& $73.4{\pm}1.6$ 
& $77.8{\pm}1.4$ 
& $75.2{\pm}1.5$ 
& $78.5{\pm}1.2$ 
& 82.6
& 78.1
& 80.8 \\

Ours 
& $\mathbf{93.1{\pm}0.6}$ 
& $\mathbf{89.4{\pm}0.8}$ 
& $\mathbf{90.2{\pm}0.7}$ 
& $\mathbf{91.0{\pm}0.8}$ 
& $\mathbf{90.9{\pm}0.6}$ 
& \textbf{91.5}
& \textbf{89.0}
& \textbf{90.3} \\

\bottomrule
\end{tabular}
\vspace{-1em}
\end{table*}

Table~\ref{tab:vl_quality} shows that the full pipeline achieves the best bundle-level quality. 
Direct VLM is competitive in semantics but weaker in 3D grounding and actionability; affordance-only heuristics improve affordance localization but lack reliable part and keypoint grounding; removing 3D refinement weakens all three visual annotation types. 
These results show that both language reasoning and explicit 3D grounding are needed to produce reliable priors for physics-based action annotation. The overall human-score improvement over the strongest ablation is $12.4$ points, with a paired $95\%$ confidence interval of $[9.8,15.0]$.

\paragraph{Grounding reliability.}
Beyond the aggregate matching scores in Table~\ref{tab:vl_quality}, interface-specific evaluation yields $944/1{,}080$ correct part regions ($87.4\%$ at label-compatible IoU $\geq 0.5$), $687/810$ correct affordance regions ($84.8\%$), $90.3\%$ PCK@0.05 for keypoints, and $300/360$ exact compatible-skill sets ($83.3\%$; macro precision/recall/F1 of $88.4/84.6/86.5$). On matched external subsets, our annotations achieve $44.1$ object-macro mIoU on PartNeXt~\cite{wang2025partnext}, compared with $46.70$ for SAMPart3D~\cite{yang2024sampart3d} and $41.01$ for P3-SAM~\cite{ma2025p3sam}, and $41.0/54.3$ mIoU/F1 on LASO~\cite{liu2024laso}, compared with $43.6/57.2$ for the released LASO baseline and $35.8/48.9$ for Direct VLM. These results are matched re-runs using identical subsets and metrics rather than numbers copied from the original papers.

\paragraph{Failure detection and validation filtering.}
Across $3{,}240$ asset- and part-level VLM queries, $38$ fail structured parsing and $56$ cannot be linked to a valid segmentation identifier; all $94$ detectable failures are rejected before action generation. These checks do not identify every possible semantic error. In an additional $18$-object audit, physics and geometric validation increase candidate-level intended-part selectivity from $50.0\%$ to $70.0\%$, with improvements on $16/18$ objects.

\subsection{Physics-grounded Action Annotation Quality}
\label{sec:exp_physics_quality}

The physics stage converts grounded visual-language priors into executable action banks. 
For each valid asset--skill pair, it generates candidates, applies geometry and embodiment constraints, optimizes poses or trajectories, and validates the results in simulation. Floating end effectors are used for object-centric candidate filtering, while the reported \emph{Exec.} metric evaluates goal-set inverse kinematics, collision-aware planning, and execution with the complete task-compatible robot embodiment.
Table~\ref{tab:physics_quality_and_collection} reports action annotation quality and annotation-enabled rollout collection across the atomic skills in Fig.~\ref{fig:atomic_skill_taxonomy}. 
Accepted annotations per ready asset--skill pair, embodiment mappings, and detailed validation criteria are provided in the appendix.

The full pipeline produces validated annotations across all evaluated skills and achieves the strongest aggregate performance.
Ablations show that geometry-only sampling lacks semantic precision, VL-only annotation lacks physical executability, and removing physics validation reduces execution and rollout reliability. 
Rollout results further show that the physics-validated annotation bank improves both data success and collection efficiency over annotation-free and VL-only variants.

\begin{table*}[htbp]
\centering
\caption{
Physics-grounded action annotation quality and annotation-enabled rollout collection on the audited evaluation suite. 
Upper: atomic-skill results of the full pipeline;
lower: aggregate baselines and ablations.
Human scores are mean $\pm$ standard error on a $0$--$100$ scale. Pass/Ready denotes candidate-level physics pass rate / asset--skill-pair readiness rate, while Exec. denotes execution with the complete task-compatible robot embodiment.
}
\label{tab:physics_quality_and_collection}
\scriptsize
\setlength{\tabcolsep}{3.0pt}
\renewcommand{\arraystretch}{1.05}

\begin{tabular}{lcccc|cccc}
\toprule
\multirow{2}{*}{Setting}
&
\multicolumn{4}{c|}{\textbf{Action Annotation Quality}}
&
\multicolumn{4}{c}{\textbf{Annotation-enabled Rollout Collection}}
\\
\cmidrule(lr){2-5}
\cmidrule(lr){6-9}
&
Pass/Ready (\%/\%)
& Exec. (\%)
& Ann./min
& Ann. Human
&
Data Succ. (\%)
& Traj./h
& Att./Succ.
& Traj. Human
\\
\midrule
\multicolumn{9}{l}{\textit{Atomic-skill results of the full pipeline}} \\
Grasp                 
& 60.3 / 96.4 & 98.1 & 800 & $91.2{\pm}0.7$
& 90.5 & 300 & 1.11 & $90.1{\pm}0.8$ \\

DexGrasp              
& 43.8 / 88.7 & 93.9 & 200 & $85.7{\pm}1.0$
& 86.7 & 170 & 1.15 & $86.0{\pm}1.0$ \\

BiGrasp               
& 38.5 / 90.5 & 95.0 & 105 & $87.0{\pm}0.9$
& 87.8 & 145 & 1.14 & $86.5{\pm}0.9$ \\

BiDexGrasp            
& 27.4 / 82.4 & 88.5 & 95 & $82.4{\pm}1.3$
& 82.4 & 115 & 1.21 & $82.0{\pm}1.2$ \\

Articulation          
& 41.6 / 89.6 & 93.2 & 110 & $87.2{\pm}0.9$
& 88.6 & 150 & 1.13 & $87.5{\pm}0.9$ \\

Insertion             
& 33.2 / 84.1 & 89.3 & 70 & $84.1{\pm}1.2$
& 84.2 & 125 & 1.19 & $84.0{\pm}1.1$ \\

Hanging               
& 31.5 / 81.7 & 88.7 & 60 & $83.0{\pm}1.2$
& 83.4 & 110 & 1.20 & $83.4{\pm}1.2$ \\

Deformable            
& 30.1 / 78.9 & 84.1 & 45 & $80.2{\pm}1.5$
& 80.3 & 90 & 1.25 & $80.4{\pm}1.4$ \\

Nav. / Approach       
& 58.7 / 94.2 & 96.5 & 300 & $89.6{\pm}0.8$
& 90.1 & 280 & 1.11 & $89.4{\pm}0.8$ \\

\midrule
Ours, aggregate      
& 40.6 / 87.4 & 91.9 & 198 & $85.6{\pm}1.0$
& 86.0 & 165 & 1.17 & $85.5{\pm}1.0$ \\

\midrule
\multicolumn{9}{l}{\textit{Aggregate baselines and ablations}} \\
No annotation / random
& -- & -- & -- & --
& 24.0 & 55 & 4.17 & $51.8{\pm}2.1$ \\

Geometry-only         
& 28.1 / 58.5 & 58.4 & 150 & $61.5{\pm}1.8$
& 48.2 & 90 & 2.07 & $60.8{\pm}1.7$ \\

VL-only               
& 31.7 / 65.0 & 67.5 & 130 & $70.2{\pm}1.5$
& 61.8 & 115 & 1.62 & $69.4{\pm}1.5$ \\

w/o phys. validation$^\dagger$ 
& 27.0 / 74.0 & 63.2 & 350 & $65.4{\pm}1.6$
& 57.6 & 110 & 1.74 & $65.5{\pm}1.6$ \\

Ours full             
& \textbf{40.6 / 87.4} & \textbf{91.9} & 198 & $\mathbf{85.6{\pm}1.0}$
& \textbf{86.0} & \textbf{165} & \textbf{1.17} & $\mathbf{85.5{\pm}1.0}$ \\

\bottomrule
\end{tabular}
\end{table*}

The human-score improvements over the strongest matched ablations are $15.4$ points with a paired $95\%$ confidence interval of $[11.9,18.9]$ for action annotations and $16.1$ points with an interval of $[12.6,19.6]$ for rollout trajectories.

\paragraph{Floating-to-full-robot audit.}
On $520$ audited DexGrasp asset--skill pairs with $10$ candidates per pair, $2{,}278/5{,}200$ candidates ($43.8\%$) pass floating-hand validation, leaving $461/520$ ready pairs ($88.7\%$). On the complete G1--Dex3-1 embodiment, the first-ranked floating-valid annotation succeeds for $402/461$ pairs ($87.2\%$); allowing fallback to another stored candidate increases success to $433/461$ pairs ($93.9\%$). This exact result is reported as the $93.9\%$ DexGrasp Exec. value in Table~\ref{tab:physics_quality_and_collection}.

\begin{table*}[t]
\centering
\caption{
\textbf{Framework-adapted external baseline estimates.}
\textbf{(a)} Conversion of each method's native output into our common candidate interface.
\textbf{(b)} Results under the common validation and rollout protocol.
Pass/Ready denotes physics pass rate / asset readiness rate; Exec. and Data Succ. are percentages; Ann./min and Traj./h are throughput counts; Att./Succ. is attempts per successful trajectory; Ann. H and Traj. H are mean human scores on a $0$--$100$ scale. For each \emph{Ours (matched)} row, summary metrics are aggregated over the listed skill families, with Att./Succ. recomputed as $100/\mathrm{Data\ Succ.}$. \emph{Ours full} is an all-family reference and is not directly compared with subset-specific baselines.
}
\label{tab:external_adapted_comparison}
\label{tab:app_external_adaptation}
\label{tab:app_external_adapted}
\scriptsize
\setlength{\tabcolsep}{3pt}
\renewcommand{\arraystretch}{1.08}

\textbf{(a) Adaptation protocol}
\vspace{0.25em}

\begin{tabular}{p{0.145\linewidth}p{0.205\linewidth}p{0.47\linewidth}p{0.13\linewidth}}
\toprule
Method & Native Output & Adaptation in Our Framework & Evaluation Scope \\
\midrule
ManiTwin-adapted
& Verified grasp poses and functional / grasp points
& Store the grasp annotations as candidates, then apply our grasp validator and rollout protocol.
& Grasp \\

MOPS-affordance + heuristic
& Pixel-level affordance labels
& Treat affordance masks as visual anchors, sample heuristic action candidates, then apply our validator and rollout protocol.
& Visual affordance \\

RoboGen-adapted
& Generated tasks, scenes, and task programs
& Convert generated task targets or scripted parameters into candidates, then apply our validator and rollout protocol.
& Grasp, Articulation \\

MolmoBot-adapted
& VLM-guided interaction proposals or actions
& Use proposed interaction regions as anchors, instantiate the closest skill templates, then apply our validator and rollout protocol.
& Grasp, Articulation, Navigation \\

GenManip-adapted
& Generated manipulation programs or trajectories
& Convert generated targets and waypoints into candidate annotations, then apply our validator and rollout protocol.
& Grasp, Articulation, Insertion \\
\bottomrule
\end{tabular}

\vspace{0.55em}
\textbf{(b) Results under the common protocol}
\vspace{0.25em}

\resizebox{\textwidth}{!}{%
\begin{tabular}{llcccccccc}
\toprule
Method
& Evaluation Scope
& Pass/Ready
& Exec.
& Ann./min
& Ann. H
& Data Succ.
& Traj./h
& Att./Succ.
& Traj. H \\
\midrule
\multicolumn{10}{l}{\textit{Action-producing baselines with skill-family-matched rows}} \\
ManiTwin-adapted
& Grasp
& 55.8 / 82.6
& 87.4
& 95
& 80.5
& 74.0
& 135
& 1.35
& 78.8 \\
Ours (matched)
& Grasp
& \textbf{60.3 / 96.4}
& \textbf{98.1}
& \textbf{800}
& \textbf{91.2}
& \textbf{90.5}
& \textbf{300}
& \textbf{1.11}
& \textbf{90.1} \\
\midrule
RoboGen-adapted
& Grasp, Articulation
& 22.5 / 45.2
& 54.8
& 18
& 59.6
& 39.5
& 42
& 2.53
& 57.0 \\
Ours (matched)
& Grasp, Articulation
& \textbf{51.0 / 93.0}
& \textbf{95.7}
& \textbf{455}
& \textbf{89.2}
& \textbf{89.6}
& \textbf{225}
& \textbf{1.12}
& \textbf{88.8} \\
\midrule
MolmoBot-adapted
& Grasp, Articulation, Nav.
& 29.4 / 58.0
& 64.1
& 34
& 68.2
& 53.5
& 82
& 1.87
& 66.8 \\
Ours (matched)
& Grasp, Articulation, Nav.
& \textbf{53.5 / 93.4}
& \textbf{95.9}
& \textbf{403}
& \textbf{89.3}
& \textbf{89.7}
& \textbf{243}
& \textbf{1.11}
& \textbf{89.0} \\
\midrule
GenManip-adapted
& Grasp, Articulation, Insertion
& 27.8 / 55.4
& 61.5
& 28
& 66.5
& 50.2
& 74
& 1.99
& 64.0 \\
Ours (matched)
& Grasp, Articulation, Insertion
& \textbf{45.0 / 90.0}
& \textbf{93.5}
& \textbf{327}
& \textbf{87.5}
& \textbf{87.8}
& \textbf{192}
& \textbf{1.14}
& \textbf{87.2} \\
\midrule
\multicolumn{10}{l}{\textit{Perception-derived heuristic; no directly matched action-annotation row}} \\
MOPS-affordance + heuristic
& Visual affordance
& --
& --
& --
& 71.0
& 42.6
& 60
& 2.35
& 62.5 \\
\midrule
\multicolumn{10}{l}{\textit{All-family reference}} \\
Ours full
& All skill families
& 40.6 / 87.4
& 91.9
& 198
& 85.6
& 86.0
& 165
& 1.17
& 85.5 \\
\bottomrule
\end{tabular}%
}
\end{table*}

\paragraph{Framework-adapted external comparisons.}
Existing systems expose different native outputs and skill coverage, so their published metrics are not directly comparable. We convert each method's closest compatible output into our candidate interface and evaluate the converted candidates with the same physics validator and rollout criteria used for our method. Table~\ref{tab:external_adapted_comparison} therefore reports \emph{framework-adapted estimates}, not official numbers or end-to-end reproductions of the original systems. For action-producing baselines, each \textbf{Ours (matched)} row is recomputed over the same reported skill families. Across these matched subsets, our complete pipeline consistently improves readiness, execution success, human-rated annotation quality, and rollout utility. Because the adapters add a common candidate representation and evaluator, part of each gap may reflect conversion fidelity. The MOPS row evaluates affordance masks followed by our heuristic action sampler and is consequently reported as a separate perception-derived comparison.

\paragraph{Protocol-matched DexGrasp comparison.}
We additionally compare with BODex~\cite{chen2025bodexscalableefficientrobotic} using the same Dex3-1 hand, $520$ audited asset--skill pairs, $10$ candidates per pair, floating-hand validator, and G1 full-arm execution protocol, without output conversion or hand retargeting. BODex obtains $40.1/84.8/90.2\%$ Pass/Ready/Exec., compared with $43.8/88.7/93.9\%$ for AnnotateAnything. For the matched ManiTwin Grasp comparison, the paired $95\%$ confidence intervals for our improvements are $+4.5$ percentage points $[3.6,5.4]$ in Pass, $+13.8$ $[8.1,19.5]$ in Ready, and $+10.7$ $[8.8,12.6]$ in Exec. These comparisons are restricted to the stated protocol-matched or framework-adapted settings.

\subsection{Real-World Experiments}
\label{sec:real_world}

As a lightweight real-world sanity check, we train policies only from annotation-enabled simulation rollouts with domain randomization and evaluate them zero-shot on representative real-world tasks. A Franka Panda with its stock two-finger gripper performs Grasp, Insertion, Open Drawer, Close Lid, and long-horizon Pick-and-Place, while the same arm with an XHand attached through a custom mount performs DexGrasp. For each evaluated object or configuration, we fine-tune the pretrained $\pi_{0.5}$ policy~\cite{Intelligence202505AV} on $1{,}000$ automatically collected simulation demonstrations generated with the corresponding simulated embodiment. We use no real-world demonstrations, real-world fine-tuning, or test-time adaptation.

The controlled evaluation contains $540$ single-task trials and $100$ long-horizon episodes. Each evaluated object, configuration, or long-horizon object set receives $20$ independently reset physical runs.

The long-horizon task requires five objects to be manipulated and placed sequentially within each episode, so an episode is counted as successful only when the complete sequence succeeds. Since we do not include a matched physical baseline, these results provide evidence of zero-shot sim-to-real feasibility rather than comparative real-world superiority. Additional qualitative hardware demonstrations are excluded from the controlled trial totals.

\begin{table}[htbp]
\centering
\caption{
Zero-shot real-world transfer of policies trained from annotation-enabled simulation rollouts. All entries report physical-robot evaluations: the first five task columns report single-task trials, while the Long-horizon column reports episodes. Wilson $95\%$ confidence intervals are provided in Appendix~H.9.
}
\label{tab:real_world}
\scriptsize
\setlength{\tabcolsep}{2.8pt}
\renewcommand{\arraystretch}{1.05}
\resizebox{\columnwidth}{!}{%
\begin{tabular}{lcccccc}
\toprule
Metric 
& Grasp 
& DexGrasp 
& Insertion 
& Open Drawer
& Close Lid
& Long-horizon \\
\midrule
Success
& 186 / 200
& 154 / 200
& 28 / 40
& 33 / 40
& 53 / 60
& 61 / 100 \\
Success (\%)
& 93.0
& 77.0
& 70.0
& 82.5
& 88.3
& 61.0 \\
\bottomrule
\end{tabular}%
}
\vspace{-1em}
\end{table}

\section{Conclusion}
\label{sec:conclusion}
We presented \textbf{AnnotateAnything}, a unified framework for automatically converting raw 3D assets into manipulation-ready annotations. By integrating VLM-driven human-prior reasoning with physics-grounded optimization and scalable parallel validation, AnnotateAnything generates diverse, executable, and asset-specific labels across objects, tasks, and robot embodiments. 

\clearpage

\bibliographystyle{unsrtnat}
\bibliography{main}

\clearpage
\appendix
\begin{center}
{\LARGE\bfseries Appendix}\par
\vspace{1em}
\end{center}
\addcontentsline{toc}{section}{Appendix}
\suppressfloats[t] 
\section{Appendix Overview}
\label{app:overview}

This appendix collects the detailed prompts, schemas, implementation details, supporting experiments, governance notes, and qualitative examples that complement the main paper. \textbf{Videos, code, demo assets, and annotations are provided in supplementary materials.}

\begin{itemize}[leftmargin=*,itemsep=0.3em,topsep=0.3em]
    \item \textbf{Appendix~\ref{app:qualitative_example} (Qualitative Example).} Provides an end-to-end qualitative walkthrough of the annotation pipeline on representative assets and scenes.
    \item \textbf{Appendix~\ref{app:language_pipeline} (Language Annotation Pipeline).} Details the asset-level and room-level language prompts, output schemas, and example annotations used in the visual-language stage.
    \item \textbf{Appendix~\ref{app:visual_annotation} (Visual Annotation Pipeline).} Describes the visual grounding procedures, including keypoints, part segmentation, occupancy maps, and room-level geometric annotations.
    \item \textbf{Appendix~\ref{app:action_annotation} (Details of Physics Annotation Pipeline).} Expands the unified action schema and the general generation, optimization, validation, and augmentation pipeline for executable action annotations.
    \item \textbf{Appendix~\ref{sec:primitive_specific_annotation} (Primitive-specific Action Annotation).} Summarizes skill-specific instantiations for grasping, articulation, insertion, hanging, garment manipulation, and navigation-related actions.
    \item \textbf{Appendix~\ref{supp:robot-data-collection} (Additional Details of Large-scale Robot Data Collection).} Covers downstream data generation, affordance supervision, robot reasoning VQA, and 3D VLM instruction-tuning data derived from the annotations.
    \item \textbf{Appendix~\ref{app:exp_details} (Additional Experimental Details).} Reports audited-suite details, coverage statistics, extra analysis, and supplementary experiment breakdowns.
    \item \textbf{Appendix~\ref{app:related_work} (Related Work).} Provides a broader discussion of adjacent literature on simulation data generation, affordance annotation, assets, and skill-centric learning.
    \item \textbf{Appendix~\ref{app:governance} (Limitations, Broader Impact, and Asset Licensing).} Documents compute resources, limitations, broader impacts, external asset licensing, and release considerations for new derived assets.
\end{itemize}
\section{Qualitative Example}
\label{app:qualitative_example}
\begin{figure*}[ht]
  \centering
  \includegraphics[width=0.98\linewidth]{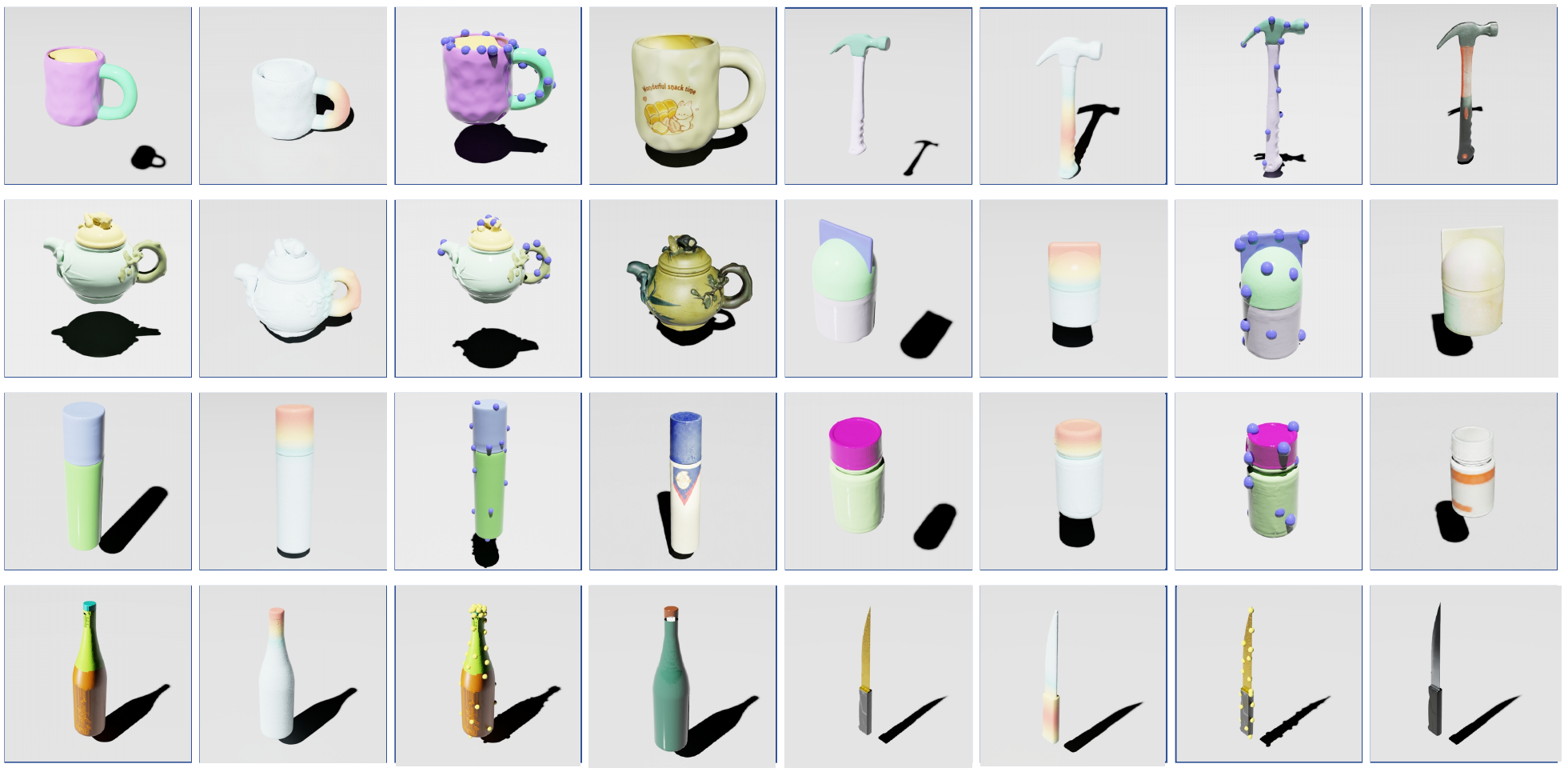}
  \caption{Qualitative gallery of asset-level annotations across diverse object categories. Each row shows multiple instances within a category (mugs, hammers, teapots, kettles, lipsticks, bottles, knives),
  illustrating the geometric and appearance diversity covered by our automatic annotation pipeline at the object-asset level.}
  \label{fig:supp_asset_gallery}
\end{figure*}

\begin{figure}[H]
  \centering
  \includegraphics[width=0.98\linewidth]{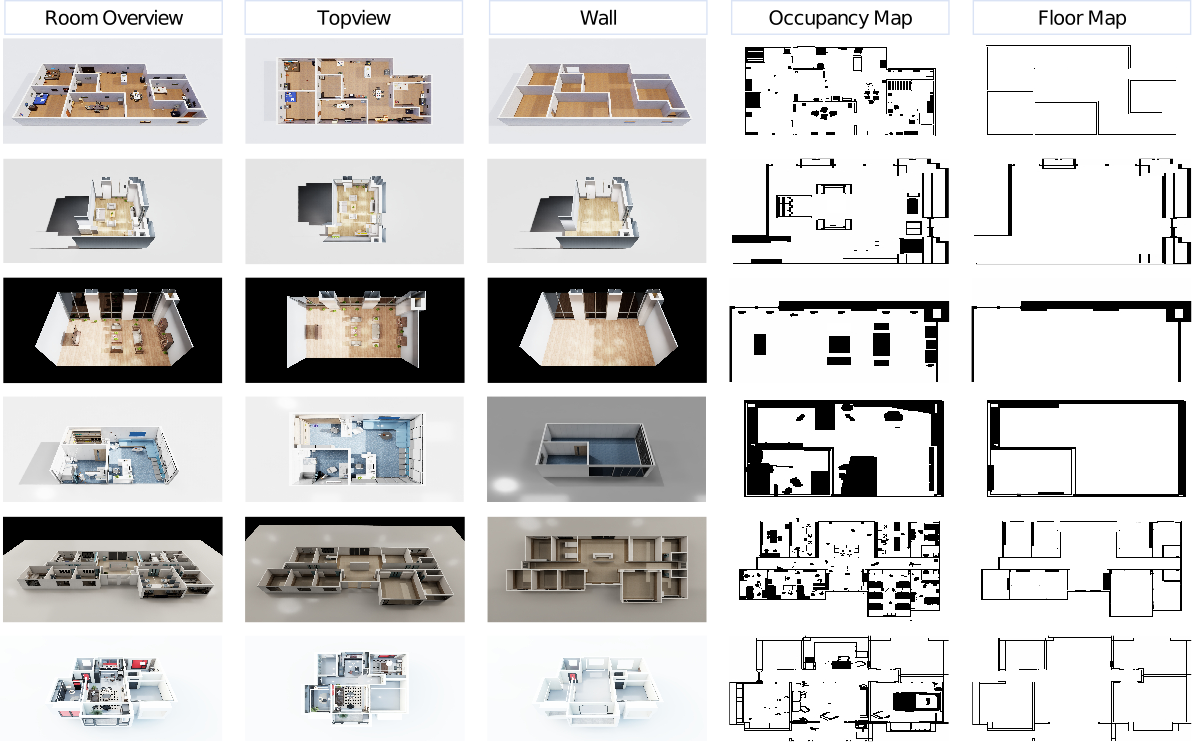}
  \caption{Qualitative examples of room-level visual annotations. From left to right: full 3D room overview, top-view rendering, wall-only structural map, multi-height occupancy map, and cleaned floor plan. Each
  row corresponds to a different scene, demonstrating that the pipeline produces consistent structural annotations across rooms with varying layouts and clutter.}
  \label{fig:supp_room_visual_annotation}
\end{figure}

\begin{figure}[H]
  \centering
  \includegraphics[width=0.98\linewidth]{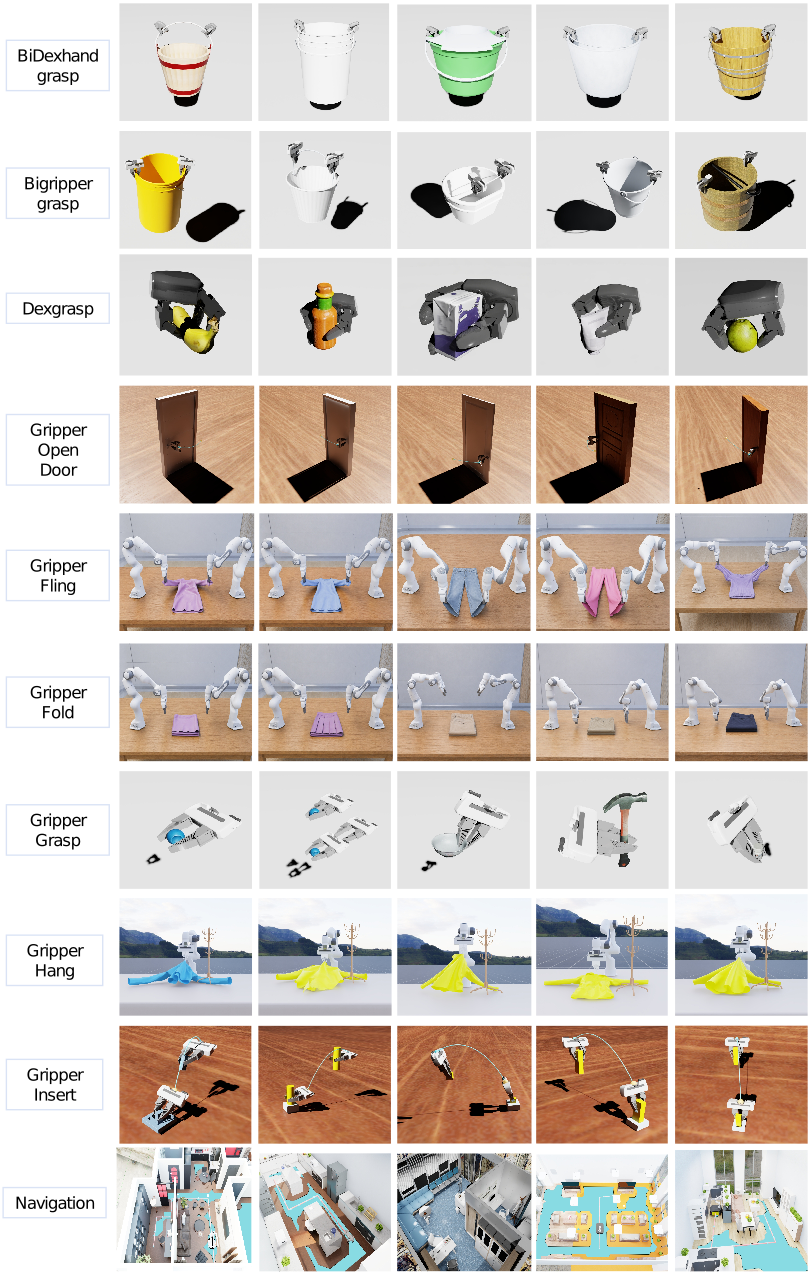}
  \caption{Qualitative examples of action annotations across manipulation skills. Each row shows several instances of one skill family, including bidexterous grasp, bigripper grasp, dexterous grasp, gripper
  open-door, fling, fold, grasp, hang, insert, and navigation. The examples illustrate that the unified action schema produces functionally meaningful candidate trajectories across both rigid and deformable
  objects.}
  \label{fig:supp_action_skills_1}
\end{figure}

\begin{figure}[H]
  \centering
  \includegraphics[width=0.98\linewidth]{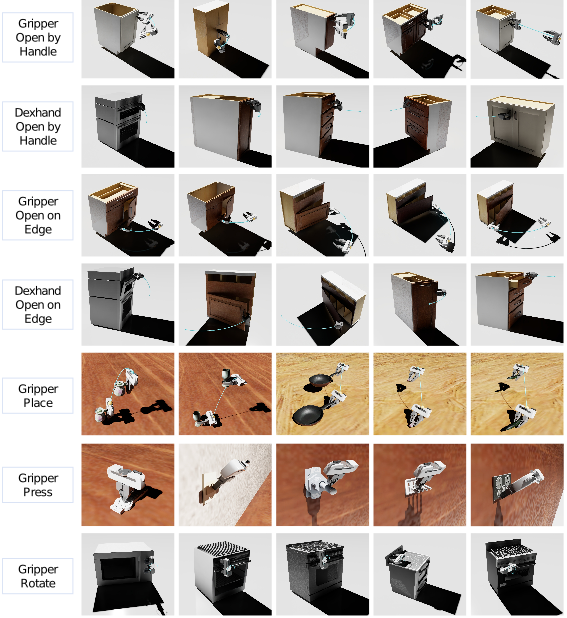}
  \caption{Additional qualitative examples of action annotations for articulated and contact-rich skills. Each row shows several instances of one skill, including gripper/dexterous opening by handle,
  gripper/dexterous opening by edge, place, press, and rotate. The samples highlight how the same articulated object admits multiple feasible interaction strategies, parameterized by different anchor types and
  end-effector embodiments.}
  \label{fig:supp_action_skills_2}
\end{figure}
\begin{table}[htbp]
\centering
\caption{Hierarchical annotation schema of AnnotateAnything.}
\label{tab:annotation_schema}
\resizebox{\linewidth}{!}{
\begin{tabular}{lll}
\toprule
Level & Annotation Type & Example Outputs \\
\midrule
Asset & Language & Sparse tags; dense object-, part- and task-level descriptions \\
Asset & Visual & 3D keypoints; part segmentation \\
Room & Language & Scene descriptions; object relations; task contexts \\
Room & Visual & Occupancy maps; floor plans; top-view/BEV maps; object layouts \\
Action & Rigid object & Parallel-jaw grasps; dexterous contacts; insertion and hanging poses \\
Action & Articulated object & Handles; motion axes; articulation waypoints; opening and closing trajectories \\
Action & Deformable object & Garment keypoints; bimanual grasp points; folding and hanging trajectories \\
Action & Room-scale scene & Navigation targets; approach poses; interaction-ready candidate base poses \\
\bottomrule
\end{tabular}
}
\end{table}

\section{Language Annotation Pipeline}
\label{app:language_pipeline}

\subsection{Asset-level Language Annotation Details}
\label{supp:asset_language_annotation}

\paragraph{Multi-view and hierarchical annotation.}
For each object-level asset, we normalize camera placement using the center $\mathbf{c}$ and diagonal length $d$ of its axis-aligned bounding box. Eight RGB views are rendered from evenly spaced azimuths on a horizontal circle, with camera distance proportional to $d$. Qwen3.5-VL~\cite{qwenteam2026qwen35omnitechnicalreport} then produces a phrase-level label, a sentence-level functional description, and a paragraph-level manipulation description. For part-aware descriptions, each segmented part is highlighted in turn and queried with the remaining object as context, linking the output to its segmentation ID.

\paragraph{Output and filtering.}
The model returns structured JSON. Prompts restrict descriptions to visible evidence and require uncertain attributes to be marked as \texttt{unknown}. We discard unparseable outputs and part descriptions that cannot be matched to a valid segmentation ID. Retained annotations provide semantic priors for visual grounding and physics-based action generation.

\begin{figure*}[t]
  \centering
  \includegraphics[width=0.98\linewidth]{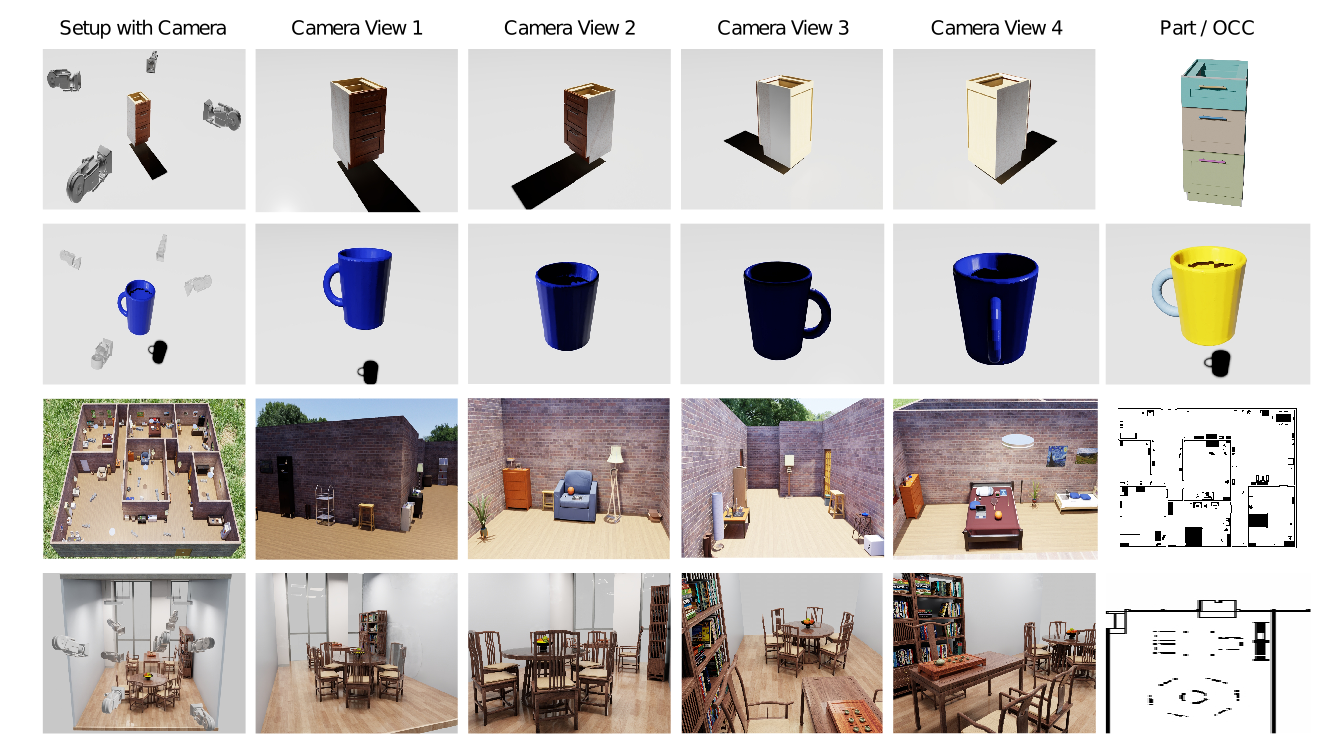}
  \caption{Asset-level language annotation. Eight views from normalized camera placements support global descriptions, while highlighted part renderings produce part-aware descriptions grounded in segmentation regions.}
  \label{fig:supp_asset_language_pipeline}
\end{figure*}

\paragraph{Global asset-level prompt.}
\begin{promptbox}{Global asset-level language prompt}
System: You are a careful annotation engine for robot manipulation assets.
Your task is to describe the visible 3D object for robot reasoning and manipulation.
Use only visual evidence from the provided views. Do not invent hidden mechanisms or unsupported functions.
If an attribute is unclear, write "unknown".

User: You are given eight RGB renderings of the same 3D object from surrounding viewpoints.
Please analyze the object as a robot manipulation asset and output a structured annotation with three levels of detail.

Level 1: phrase-level label.
Write a short phrase, no more than 6 words, that identifies the object category or functional identity.

Level 2: sentence-level description.
Write one sentence, no more than 30 words, describing the object's main function and major manipulable parts.

Level 3: paragraph-level description.
Write one dense paragraph for robot manipulation reasoning.
Describe the object's visible parts, their functions, likely affordances, feasible interactions, and manipulation constraints.
Focus on what a robot can grasp, pull, push, support, open, insert into, hang on, or avoid.

Return the result in JSON format with the following fields:\\
phrase, sentence, paragraph, \texttt{major\_parts}, \texttt{uncertain\_attributes}.
\end{promptbox}

\paragraph{Part-aware prompt.}
\begin{promptbox}{Part-aware language prompt}
System: You are a robot manipulation annotation engine.
You will be given an image of a 3D object where one segmented part is highlighted.
The highlighted region is the target part. The rest of the object is shown only as context.
Describe only the highlighted part.

User: Please analyze the highlighted part for robot manipulation.
Do not describe unrelated parts unless they are needed to explain the target part.
If the highlighted part's function is uncertain, write "unknown" instead of guessing.

Please answer the following questions:
1. What is the name of the highlighted part?
2. What visual evidence supports this part name?
3. What is the function of this part?
4. What robot interactions are feasible with this part?
5. What constraints should a robot consider when interacting with this part?
6. How is this part related to the whole object?

Return the result in JSON format with the following fields:\\
\texttt{part\_name}, \texttt{visual\_evidence}, \texttt{part\_function}, \texttt{feasible\_robot\_interactions}, \texttt{manipulation\_constraints}, \texttt{relation\_to\_whole\_object}, \texttt{confidence}.
\end{promptbox}

\paragraph{Example output.}
\begin{outputbox}{Global asset-level output: mug}
\textbf{Phrase-level label:} Blue handled mug.

\textbf{Sentence-level description:}
A blue mug with a hollow cup body and a side handle, designed for containing liquid and being grasped by the handle.

\textbf{Paragraph-level description:}
The object is a mug composed of a cylindrical cup body, an open rim, a hollow interior, a side handle, and a bottom base. The cup body provides the main container volume and can be used for placement and containment reasoning. The side handle is the most important manipulation part because it supports stable grasping, lifting, and carrying. The open rim and hollow interior indicate that the mug can contain liquid or small objects, but they should be avoided as forceful grasp targets. The bottom base supports stable placement on flat surfaces.
\end{outputbox}

\begin{outputbox}{Part-aware output: mug handle}
\textbf{Highlighted part:} Mug handle.

The highlighted part is the mug handle. It is a curved loop attached to the side of the cup body. Its function is to provide a safe and stable grasping region for lifting or carrying the mug. Feasible robot interactions include grasping the outer handle, inserting fingers or gripper tips through the loop, and pulling or lifting the mug through the handle. The robot should avoid colliding with the cup body near the handle attachment points and should align the grasp with the handle opening.
\end{outputbox}

\subsection{Room-level Language Annotation Details}
\label{supp:room_language_annotation}

\paragraph{Room-view and layout annotation.}
For each room-scale scene, we sample candidate camera poses within individual rooms and use simulator segmentation as a quality signal. We retain eight views with diverse positions and directions after rejecting views with too few visible non-structural objects or views dominated by walls and floors. The VLM receives these RGB views together with a labeled top-view occupancy map containing free space, obstacles, room boundaries, and object instances. Qwen3.5-VL~\cite{qwenteam2026qwen35omnitechnicalreport} generates short-sentence, long-sentence, and paragraph-level descriptions covering room type, furniture and functional zones, spatial relations, navigation structure, interaction regions, and possible robot tasks.

\paragraph{Output and filtering.}
The VLM returns structured JSON, using RGB views for local evidence and the occupancy map for global layout. Prompts prohibit unsupported object names or hidden room functions and require uncertain attributes to be marked as \texttt{unknown}. We discard unparseable outputs and object references that cannot be matched to labeled occupancy-map instances.

\paragraph{Room-level prompt.}
\begin{promptbox}{Room-level language prompt}
System: You are a careful annotation engine for robot navigation and manipulation scenes.
Your task is to describe the visible room or indoor scene for robot reasoning.
Use the provided RGB views for local visual evidence and the labeled top-view occupancy map for global layout, spatial relations, and navigable structure.
Do not invent hidden rooms, unsupported objects, or functions that are not visible or labeled.
If an attribute is unclear, write "unknown".

User: You are given multiple RGB renderings sampled inside a room-scale scene, together with a top-view occupancy map.
The occupancy map marks free space, obstacles, room boundaries, and object instances with labels.
Please analyze the scene as a robot interaction environment and output a structured room-level annotation with three levels of detail.

Level 1: short-sentence description.
Write one short sentence, no more than 15 words, summarizing the room type and global layout.

Level 2: long-sentence description.
Write one to two long sentences describing the major furniture, functional zones, spatial arrangement, and navigation structure.

Level 3: paragraph-level scene description.
Write one dense paragraph for robot navigation and manipulation reasoning.
Describe the room layout, object relations, navigable regions, interaction-relevant areas, and possible long-horizon robot tasks.
Focus on what a robot can navigate to, avoid, approach, pick, place, open, retrieve, or interact with.

Return the result in JSON format with the following fields:
\texttt{short\_sentence}, \texttt{long\_sentence}, paragraph, \texttt{room\_type}, \texttt{functional\_zones}, \texttt{object\_inventory}, \texttt{spatial\_relations}, \texttt{navigable\_regions}, \texttt{interaction\_relevant\_areas}, \texttt{possible\_robot\_tasks}, \texttt{uncertain\_attributes}.
\end{promptbox}

\paragraph{Example output.}
\begin{outputbox}{Room-level output: dining room}
\textbf{Short-sentence description:}
A rectangular dining room with a central table and surrounding chairs.

\textbf{Long-sentence description:}
The room contains a central dining area formed by a round table and multiple chairs, with additional support and storage furniture near the rear and side walls. Open floor space around the furniture provides navigation routes, while the table, chairs, and shelf create local obstacles and approach constraints.

\textbf{Paragraph-level description:}
The room is a rectangular dining or meeting space with most interaction objects concentrated near the center and right wall. A round table with several chairs forms the main functional zone for placing, picking, and tabletop manipulation. A secondary table near the rear wall provides another support surface, while the side bookshelf can be used for object retrieval or placement. The open floor around the furniture provides navigable space, but the chairs and table legs create local obstacles that require careful path planning. A robot can navigate around the table, approach chairs or tables, retrieve objects from the table or shelf, and place objects on available support surfaces.
\end{outputbox}

\section{Visual Annotation Pipeline}
\label{app:visual_annotation}
\subsection{Asset-level Visual Annotation Details}
\label{supp:asset_visual_annotation}

\paragraph{Multi-view point-cloud reconstruction.}
For each object-level asset, we reconstruct the input point cloud from eight initial RGB-D views rather than sampling the mesh directly, keeping the final geometry derived from rendered observations. Cameras are placed at evenly spaced azimuths around the asset, oriented toward the bounding-box center, with distance scaled by object size.

Given depth $D_i$, intrinsics $K_i$, and camera-to-world extrinsics $T^w_{c_i}=[R_i\mid t_i]$, each valid pixel $(u,v)$ is back-projected as
\[
\mathbf{x}^{c_i}_{uv}
=
D_i(u,v)K_i^{-1}
\begin{bmatrix}
u\\v\\1
\end{bmatrix},
\qquad
\mathbf{x}^{w}_{uv}
=
R_i\mathbf{x}^{c_i}_{uv}+t_i.
\]
The fused cloud is
\[
\mathcal{P}_{cam}
=
\bigcup_{i=1}^{N}
\left\{
\mathbf{x}^{w}_{uv}
\mid
D_i(u,v)>0,\;
M_i(u,v)=1
\right\},
\]
where $N=8$ by default and $M_i$ is the simulated foreground mask. Masked points from all views are transformed into a common frame and merged.

\paragraph{Coverage verification with mesh-based reference points.}
Mesh samples are used only to verify camera-view coverage. For reference points $\mathcal{P}_{mesh}$, we compute
\[
\mathrm{Cov}(\mathcal{P}_{cam},\mathcal{P}_{mesh})
=
\frac{1}{|\mathcal{P}_{mesh}|}
\sum_{\mathbf{p}\in\mathcal{P}_{mesh}}
\mathbbm{1}
\left[
\min_{\mathbf{q}\in\mathcal{P}_{cam}}
\|\mathbf{p}-\mathbf{q}\|_2<\epsilon
\right],
\]
where $\epsilon$ is the distance threshold. We also compute voxel-region overlap
\[
\mathrm{IoU}_{roi}
=
\frac{
|\mathcal{V}(\mathcal{P}_{cam})\cap\mathcal{V}(\mathcal{P}_{mesh})|
}{
|\mathcal{V}(\mathcal{P}_{cam})\cup\mathcal{V}(\mathcal{P}_{mesh})|
}.
\]
If either metric is below its threshold, we add views targeting poorly covered regions and repeat reconstruction, improving coverage of thin, concave, or occluded structures without replacing the camera-derived cloud with mesh samples.

\paragraph{Point-cloud downsampling.}
Before segmentation, we downsample $\mathcal{P}_{cam}$ to $M=80{,}000$ points by farthest point sampling (FPS). Starting from an initialized set $\mathcal{S}$, FPS iteratively selects
\[
\mathbf{p}^{*}
=
\arg\max_{\mathbf{p}\in\mathcal{P}_{cam}}
\min_{\mathbf{s}\in\mathcal{S}}
\|\mathbf{p}-\mathbf{s}\|_2,
\]
and updates $\mathcal{S}\leftarrow\mathcal{S}\cup\{\mathbf{p}^{*}\}$ until $|\mathcal{S}|=M$. We denote the result by $\mathcal{P}_{80k}$.

\paragraph{Native 3D part segmentation.}
We segment $\mathcal{P}_{80k}$ using the 182M-parameter Hunyuan3D part-segmentation model, built on native 3D decomposition modules such as P3-SAM~\cite{ma2025p3sam} and X-Part~\cite{yan2025xpart}. It predicts
\[
\mathcal{M}=\{m_k\}_{k=1}^{K},
\qquad
m_k\in\{0,1\}^{|\mathcal{P}_{80k}|},
\]
with one mask per part. When denser geometry is needed, labels are propagated to $\mathcal{P}_{cam}$ by nearest-neighbor assignment:
\[
\ell(\mathbf{p})
=
\ell\left(
\arg\min_{\mathbf{q}\in\mathcal{P}_{80k}}
\|\mathbf{p}-\mathbf{q}\|_2
\right),
\qquad
\mathbf{p}\in\mathcal{P}_{cam}.
\]
Under our setup, segmentation takes approximately 30 seconds per object.

\begin{figure}[h]
    \centering
    \includegraphics[width=0.98\textwidth]{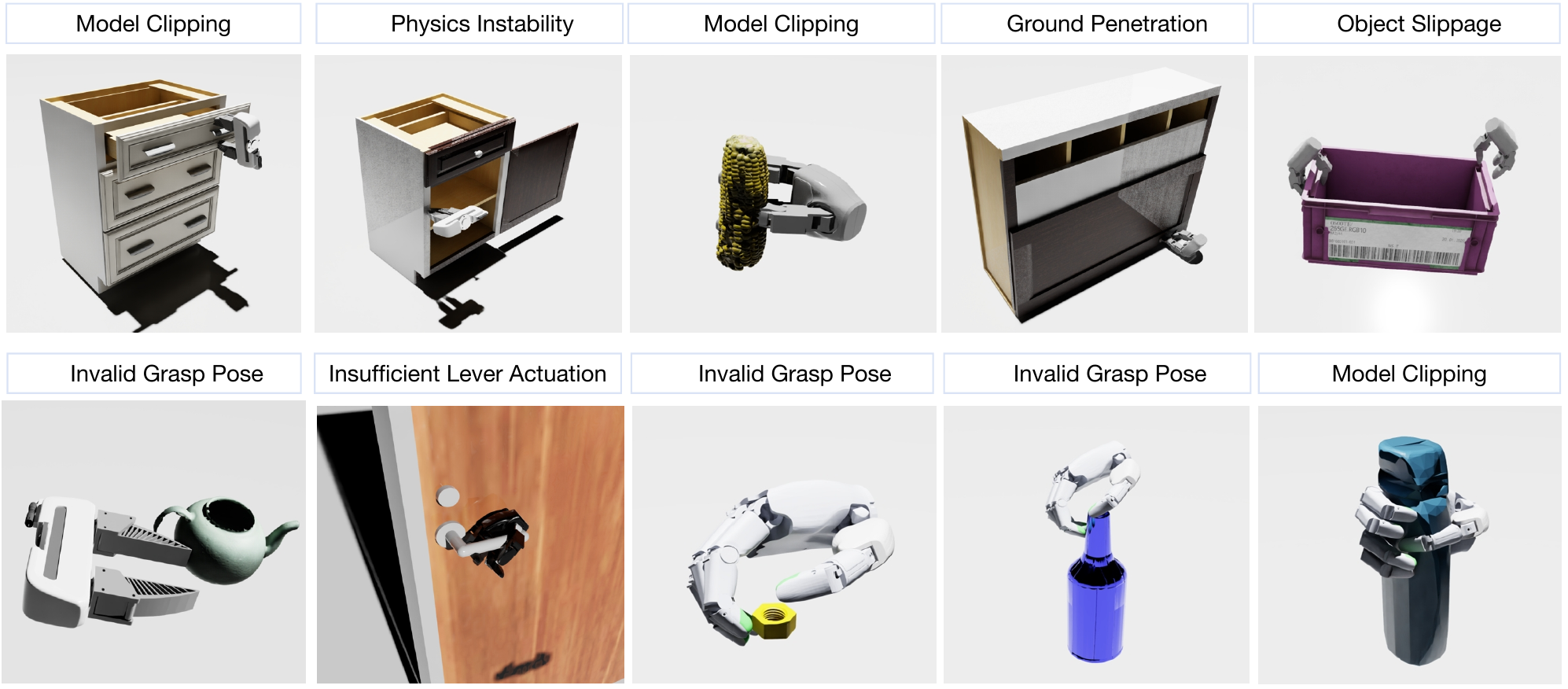}
    \caption{Common failures during physics validation of generated action candidates: model clipping, physics instability, ground penetration, object slippage, invalid grasp poses, and insufficient lever actuation. These cases motivate simulation-based filtering beyond geometric and language priors.}
    \label{fig:supp_physics_failure}
\end{figure}

\paragraph{Segmentation limitation.}
A key limitation of part segmentation is semantic granularity mismatch. For example, a wine glass may be returned as one part although manipulation benefits from separating its bowl, stem, and base. We therefore treat segmentation as a visual prior; task-specific geometric reasoning and physics validation remain necessary for executable action annotations.

\subsection{Room-level Visual Annotation Details}
\label{supp:room_visual_annotation}

\begin{figure}[h]
    \centering
    \includegraphics[width=0.98\textwidth]{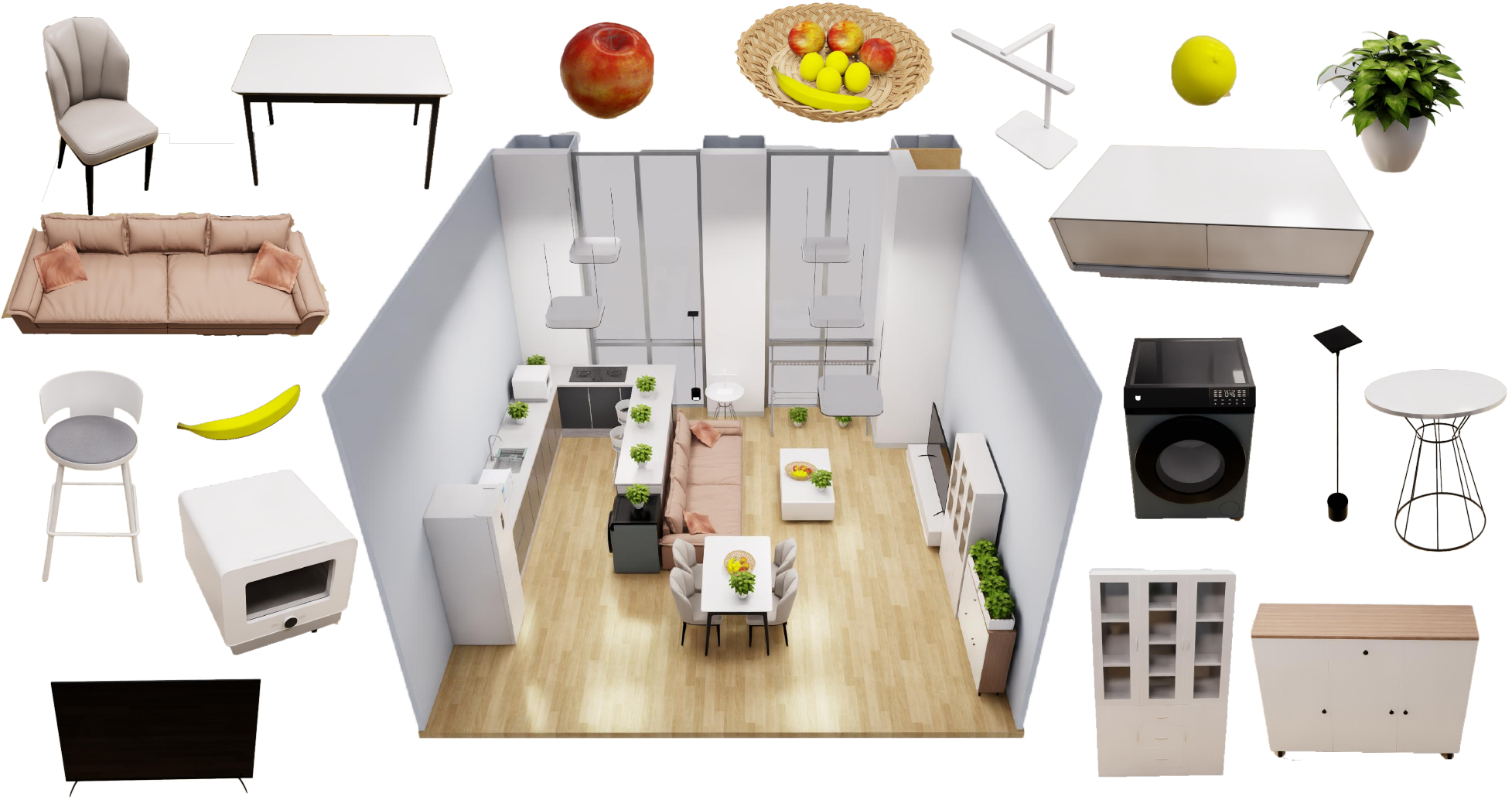}
    \caption{Room-scale scene composition. A populated 3D room (center) is assembled from the diverse object assets shown in the surrounding panels, producing cluttered scenes for scan-pose sampling and occupancy-map construction.}
    \label{fig:supp_room_composition}
\end{figure}

\paragraph{Wall-based room decomposition.}
We decompose each scene into rooms using wall boundaries and layout structure. Each room is represented by a global-frame floor-plan polygon $\mathcal{R}_j$, processed independently before its maps are stitched into the scene representation.

\paragraph{Rule-based scan-pose sampling.}
For each $\mathcal{R}_j$, we randomly sample a candidate set $\Pi_j$ of camera and LiDAR poses inside the room polygon, requiring free-space placement, clearance from walls and large obstacles, and a viewing direction toward the room interior. Simulator instance segmentation is used only to score view quality. For candidate pose $\pi$, the visible object set is
\[
\mathcal{V}(\pi)
=
\left\{
o\in\mathcal{O}_j
\;\middle|\;
\sum_{u,v}\mathbf{1}[S_{\pi}(u,v)=o]>\tau_{\mathrm{area}}
\right\},
\]
where $\mathcal{O}_j$ is the room's object set, $S_{\pi}$ is the rendered segmentation map, and $\tau_{\mathrm{area}}$ removes barely visible instances. We greedily maximize newly covered IDs:
\[
\pi_t
=
\arg\max_{\pi\in\Pi_j}
\left|
\mathcal{V}(\pi)
\setminus
\bigcup_{k<t}\mathcal{V}(\pi_k)
\right|,
\]
until
\[
\mathrm{Cov}_{obj}
=
\frac{
\left|\bigcup_t\mathcal{V}(\pi_t)\right|
}{
|\mathcal{O}_j|
}
\]
reaches its threshold or the pose budget is exhausted. This avoids uninformative wall-facing views and improves object-instance coverage.

\paragraph{Multi-height occupancy from LiDAR and ray tracing.}
For each selected pose $\pi_t$ and height $h_\ell\in\mathcal{H}$, we cast horizontal rays in uniformly sampled directions. If $\rho_\alpha$ is the first-hit distance along direction $\mathbf{d}_\alpha$, cells before the hit are free and the hit cell is occupied:
\[
\mathcal{F}_{t,\ell,\alpha}
=
\left\{
g(\mathbf{x}_t+r\mathbf{d}_{\alpha},h_\ell)
\;\middle|\;
0<r<\rho_\alpha-\delta
\right\},
\quad
\mathcal{O}_{t,\ell,\alpha}
=
g(\mathbf{x}_t+\rho_\alpha\mathbf{d}_\alpha,h_\ell),
\]
where $\mathbf{x}_t$ is the scan position, $g(\cdot)$ maps 3D points to grid cells, and $\delta$ is a safety margin. We aggregate rays with
\[
L_\ell(c)
\leftarrow
L_\ell(c)
+
\begin{cases}
\lambda_{\mathrm{occ}}, & c\in\mathcal{O}_{t,\ell,\alpha},\\
\lambda_{\mathrm{free}}, & c\in\mathcal{F}_{t,\ell,\alpha},\\
0, & \text{otherwise},
\end{cases}
\]
and compute $P_\ell(c)=\sigma(L_\ell(c))$. The multi-height annotation is
\[
\mathcal{G}_j
=
\{G_j^{h_1},G_j^{h_2},\ldots,G_j^{h_L}\},
\]
whose slices capture floor-level free space, low obstacles, furniture geometry, and wall-level structure.

\paragraph{Global map stitching.}
Room maps are aligned using the local-room-to-world transform $T_j^w$. For room map $G_j$,
\[
G^w(c)
=
\mathrm{Fuse}_j\left(
G_j\left((T_j^w)^{-1}c\right)
\right),
\]
where $\mathrm{Fuse}$ combines overlapping evidence from adjacent rooms and shared boundaries. The result includes global occupancy, object-layout, and top-view maps.

\paragraph{VLM-assisted wall and floor-plan filtering.}
Because raw occupancy includes both structure and movable objects, we generate a labeled connected-component map and ask the VLM to retain persistent structural components while removing furniture and clutter. Given $\mathcal{C}=\{C_k\}$ and predicted structural subset $\mathcal{C}_{wall}\subset\mathcal{C}$,
\[
W
=
\bigcup_{C_k\in\mathcal{C}_{wall}}C_k.
\]
The resulting annotations comprise multi-height occupancy, object-layout, cleaned floor-plan, and wall-structure maps, which support exploration, navigation-target generation, and scene-level action annotation.

\paragraph{Prompt for VLM-assisted wall filtering.}
\begin{promptbox}{Wall-structure filtering prompt}
System: You are a careful annotation engine for indoor scene structure. 
Your task is to identify persistent architectural structures in a top-view map. 
Keep walls, room boundaries, door frames, and other fixed structural elements. 
Remove movable objects such as tables, chairs, cabinets, shelves, sofas, lamps, and small objects.

User: You are given a top-view occupancy map with connected components and object labels. 
Each connected component has an ID. 
Please decide which components correspond to permanent wall or room-boundary structures.

Rules:
1. Keep components that form long, continuous room boundaries.
2. Keep components that separate rooms or define door openings.
3. Remove components that correspond to furniture, movable objects, or clutter.
4. If a component is uncertain, mark it as "uncertain" instead of guessing.

Return the result in JSON format with the following fields:
\texttt{wall\_component\_ids}, \texttt{removed\_object\_component\_ids}, \texttt{uncertain\_component\_ids}, reasoning.
\end{promptbox}

\section{Details of the Physics Annotation Pipeline}
\label{app:action_annotation}

\subsection{Unified Action Annotation Schema}
\label{supp:action_schema_details}

Action annotations are organized as a one-to-many mapping from grounded visual anchors to compatible skills and multiple action candidates:
\[
\text{visual anchor}
\rightarrow
\text{functional affordance}
\rightarrow
\text{compatible skills}
\rightarrow
\text{candidate bank}.
\]
For an asset $\mathcal{A}$, the visual pipeline provides grounded anchors
\[
\mathcal{H}(\mathcal{A})=
\mathcal{P}(\mathcal{A})\cup
\mathcal{K}(\mathcal{A})\cup
\mathcal{R}^{\mathrm{aff}}(\mathcal{A})\cup
\mathcal{R}^{\mathrm{scene}}(\mathcal{A}),
\]
where $\mathcal{P}$, $\mathcal{K}$, $\mathcal{R}^{\mathrm{aff}}$, and $\mathcal{R}^{\mathrm{scene}}$ denote part regions, semantic keypoints, affordance regions, and scene-level regions, respectively. Each anchor $h$ has a grounded functional affordance $\phi(h)$ and a compatible skill set $\mathcal{S}(h)$. For each $s\in\mathcal{S}(h)$, we store
\[
\mathcal{B}_{h,s}=\{a_{h,s}^{(i)}\}_{i=1}^{N_{h,s}},
\qquad
a_{h,s}^{(i)}=
\left(s,o(h),h,\phi(h),x^{(i)},\theta^{(i)},\tau^{(i)},v^{(i)},d^{(i)}\right).
\]
Here, $o(h)$ is the object or scene associated with anchor $h$. Candidate fields are populated progressively with targets, parameters, trajectories, optimization outputs, and validation metadata; only candidates that pass physics validation are retained as final action annotations. This schema separates physical feasibility from functional suitability while preserving multiple valid actions for each anchor--skill pair.

\begin{table*}[t]
\centering
\small
\renewcommand{\arraystretch}{1.10}
\caption{Symbols in the unified action annotation schema.}
\label{tab:supp_action_schema_symbols}
\begin{tabularx}{\textwidth}{>{\raggedright\arraybackslash}p{0.17\textwidth}>{\raggedright\arraybackslash}p{0.23\textwidth}>{\raggedright\arraybackslash}X}
\toprule
Symbol & Meaning & Representative contents \\
\midrule
$\mathcal{A}$ & Input asset & Object or room-scale scene. \\
$\mathcal{H}(\mathcal{A})$ & Grounded anchors & Parts, keypoints, affordance regions, and scene regions. \\
$h,\phi(h)$ & Anchor and functional affordance & Mug handle with grasp-for-use; cabinet handle with grasp-for-opening; free space with navigation approach. \\
$\mathcal{S}(h)$ & Compatible skills & Grasp, dexterous grasp, articulation, insertion, hanging, deformable manipulation, or navigation. \\
$\mathcal{B}_{h,s}$ & Candidate bank & Candidates for skill $s$ at anchor $h$; final retained entries have passed validation. \\
$o(h)$ & Associated instance & Object or scene containing anchor $h$. \\
$a_{h,s}^{(i)}$ & Action candidate & One target, parameterization, trajectory, metadata record, and diversity descriptor. \\
$x^{(i)}$ & Concrete target & Contact point, surface patch, support edge, opening frame, handle point, or base pose. \\
$\theta^{(i)}$ & Action parameters & Pose, width, hand configuration, direction, joint range, or base orientation. \\
$\tau^{(i)}$ & Trajectory & Approach-contact-retreat motion, articulation waypoints, or navigation path. \\
$v^{(i)}$ & Validation metadata & Collision, IK, stability, task progress, robustness, and functional success. \\
$d^{(i)}$ & Diversity descriptor & Target location, approach direction, contact mode, trajectory family, or embodiment profile. \\
\bottomrule
\end{tabularx}
\end{table*}

\paragraph{Examples.}
For a mug, the handle is assigned a grasp-for-use affordance and supports parallel-jaw and dexterous grasp banks with varied contacts, orientations, and approach directions. The body can still support physical or stabilizing grasps, but these candidates receive different functional metadata because they may obstruct pouring or access to the opening. For a cabinet, the same handle anchor can support grasp and articulation banks, while a scene-level free-space anchor supports interaction-ready navigation poses. Thus, each asset is represented by structured, function-conditioned candidate banks rather than a single canonical action.

\subsection{General Physics-based Action Generation Pipeline}

\subsubsection{Candidate Target Generation}
\label{supp:candidate_target_generation}

Candidate target generation converts language and visual annotations into concrete anchors:
\[
\text{skill }s+\mathcal{Y}^{\mathrm{lang}}+\mathcal{Y}^{\mathrm{vis}}
\rightarrow
\mathcal{C}_s(\mathcal{A})
\rightarrow
\text{sampled and filtered targets}.
\]
The skill-compatible regions are
\[
\mathcal{C}_s(\mathcal{A})=
\Gamma_s\!\left(
\mathcal{Y}^{\mathrm{lang}},
\mathcal{Y}^{\mathrm{vis}},
\mathcal{G}(\mathcal{A})
\right),
\]
where $\mathcal{Y}^{\mathrm{lang}}$ contains language semantics and functional priors, $\mathcal{Y}^{\mathrm{vis}}$ contains grounded parts, keypoints, affordance regions, and scene maps, $\mathcal{G}(\mathcal{A})$ denotes asset geometry, and $\Gamma_s$ is the skill-specific region-construction operator. Each region $c\in\mathcal{C}_s(\mathcal{A})$ is paired with a functional affordance $\phi(c)$, preventing uniform sampling over semantically irrelevant geometry.

\begin{table*}[htbp]
\centering
\small
\renewcommand{\arraystretch}{1.10}
\caption{Candidate-region sources, sampling methods, and principal filters.}
\label{tab:supp_candidate_region_examples}
\label{tab:supp_sampling_filtering_examples}
\begin{tabularx}{\textwidth}{>{\raggedright\arraybackslash}p{0.15\textwidth}>{\raggedright\arraybackslash}p{0.21\textwidth}>{\raggedright\arraybackslash}p{0.22\textwidth}>{\raggedright\arraybackslash}X}
\toprule
Skill & Region source & Sampling & Main filters \\
\midrule
Functional and physical grasp & Part mask, language, geometry & FPS on eligible surfaces & Part size, contact area, normals, gripper width, collision, reachability, and affordance consistency. \\
Dexterous grasp & Part/affordance region & Contact-set or FPS seeds & Joint limits, self-collision, object penetration, support, and functional-region consistency. \\
Garment manipulation & Semantic keypoints & Keypoint perturbations and bimanual pairs & Visibility, confidence, pair distance, bimanual reachability, and deformation feasibility. \\
Articulation & Handle or movable-part region & FPS or keypoint sampling & Motion-axis consistency, contact maintenance, collision, reachability, and actuation direction. \\
Insertion and hanging & Opening, socket, hook, or support edge & Region and frame sampling & Size, alignment, clearance, support stability, and collision-free approach. \\
Navigation & Occupancy/BEV free space & Grid sampling or 2D FPS & Traversability, clearance, connectivity, visibility, and manipulation reachability. \\
\bottomrule
\end{tabularx}
\end{table*}

For a surface region $c$, spatially diverse anchors are sampled by FPS:
\[
p_1\sim\mathcal{P}_c,
\qquad
p_j=
\arg\max_{p\in\mathcal{P}_c}
\min_{1\leq k<j}\|p-p_k\|_2,
\]
where $\mathcal{P}_c$ contains the surface points inside $c$. Local normals, tangents, curvature, and reference frames are retained for later pose and trajectory generation. Semantic keypoints can be used directly or locally perturbed; bimanual pairs are additionally filtered by distance, symmetry, visibility, and reachability. For scene-level skills, we sample
\[
q_j=(x_j,y_j,\psi_j)
\]
from traversable cells, with $\psi_j$ oriented toward the target object or affordance.

The retained targets are
\[
\mathcal{T}_{c,s}=\{x_{c,s}^{(j)}\}_{j=1}^{M_{c,s}},
\]
and seed the corresponding bank $\mathcal{B}_{h,s}$. At this stage, candidates contain target-level information; later stages instantiate parameters and trajectories, optimize them for the asset and embodiment, and attach physics-validation metadata. For example, mug-handle samples seed functional grasp candidates, garment keypoints seed bimanual fold or fling candidates, and cabinet handles seed both grasp and articulation candidates. Room-scale free-space samples additionally seed approach poses around the cabinet.

\subsection{Trajectory Generation}
\label{supp:trajectory_generation}

Each retained target is converted into a template-level trajectory
\[
\tau_0^{(i)}=
\mathcal{G}_s\!\left(
 h,\phi(h),x^{(i)},\theta^{(i)};\mathcal{A}
\right),
\]
which is subsequently optimized and validated. We group generators by the source of their constraints.

\begin{table*}[t]
\centering
\small
\renewcommand{\arraystretch}{1.10}
\caption{Trajectory-generation families and representative templates.}
\label{tab:supp_trajectory_constraint_families}
\label{tab:supp_garment_templates}
\label{tab:supp_articulation_templates}
\label{tab:supp_vector_templates}
\begin{tabularx}{\textwidth}{>{\raggedright\arraybackslash}p{0.21\textwidth}>{\raggedright\arraybackslash}p{0.22\textwidth}>{\raggedright\arraybackslash}p{0.25\textwidth}>{\raggedright\arraybackslash}X}
\toprule
Constraint family & Representative skills & Main inputs & Template \\
\midrule
Object property & Grasp, dexterous grasp, fold, fling & Geometry, part masks, keypoints, grasp frames, garment scale & Pre-grasp/contact/retreat motion or keypoint-conditioned bimanual motion. \\
Object trajectory & Open, close, rotate, push & Joint axis/state, moving-part transform, handle pose & End-effector waypoints that follow the moving part while preserving grasp or contact. \\
Object vector & Insert, hang, pour, place & Insertion axes, support edge, pour direction, surface normal & Alignment, approach, execution, release, and retreat along semantic/geometric vectors. \\
Scene trajectory & Navigation and approach & Occupancy/BEV map, free space, target region, base model & A* seed followed by embodiment-aware local refinement. \\
\bottomrule
\end{tabularx}
\end{table*}

\paragraph{Object-property-conditioned templates.}
For a parallel-jaw grasp with pose $T_{\mathrm{grasp}}=[R_{\mathrm{grasp}},p_{\mathrm{grasp}}]$ and approach direction $\mathbf{a}_{\mathrm{grasp}}$, we use
\[
T_{\mathrm{pre}}=
\left[R_{\mathrm{grasp}},
 p_{\mathrm{grasp}}-\delta_{\mathrm{pre}}\mathbf{a}_{\mathrm{grasp}}\right],
\qquad
T_{\mathrm{ret}}=
\left[R_{\mathrm{grasp}}\
 p_{\mathrm{grasp}}+\delta_{\mathrm{ret}}\mathbf{r}\right]
 \]
 \[
\tau_0=(T_{\mathrm{pre}},T_{\mathrm{grasp}},T_{\mathrm{close}},T_{\mathrm{ret}}).
\]
Here, $\mathbf{r}$ is a retreat direction and $T_{\mathrm{close}}$ denotes gripper closure at the grasp pose. Dexterous templates additionally specify a palm pose, contact seeds, joint configuration, and closure schedule.

Garment templates are generated from semantic keypoints. For an upper-body garment, a sleeve endpoint is reflected across the shoulder--bottom fold line,
\[
k_{\mathrm{lslv}}^{\star}=
\mathrm{Ref}_{\ell(k_{\mathrm{lsh}},k_{\mathrm{lb}})}
(k_{\mathrm{lslv}}),
\]
and the bimanual sequence lifts, transfers, and places the sleeve near this target. Bottom-up folds move hem keypoints toward the shoulder region. Pants use a side-to-center fold followed by one or more bottom-to-waist folds. Fling templates select two keypoints and scale lift and stretch distances with their separation while enforcing bimanual reachability.

\paragraph{Object-trajectory-conditioned templates.}
For articulated parts, let $T_{\mathrm{part}}(q)$ be the moving-part pose at joint state $q$. After establishing the initial contact,
\[
T_{\mathrm{rel}}=
T_{\mathrm{part}}(q_{\mathrm{init}})^{-1}T_{\mathrm{eef}}^{\mathrm{init}},
\qquad
T_{\mathrm{eef}}(q)=T_{\mathrm{part}}(q)T_{\mathrm{rel}}.
\]
Waypoints are generated by interpolating $q$ from the initial to the target state. This covers revolute and prismatic opening/closing, partial pulls, lid motion, button rotation, and short pushes; push-based actions can instead maintain surface contact without a fixed grasp.

\paragraph{Object-vector-conditioned templates.}
For insertion, both $\mathbf{u}_{\mathrm{act}}$ and
$\mathbf{u}_{\mathrm{pas}}$ denote the positive physical insertion
direction. When the passive-side geometry initially provides an outward
surface normal, we negate it while constructing the passive insertion
frame. The active and passive axes are then aligned as
\[
R\mathbf{u}_{\mathrm{act}}
\approx
\mathbf{u}_{\mathrm{pas}}.
\]
then the trajectory approaches and translates along the passive axis:
\[
p_{\mathrm{pre}}=p_{\mathrm{tar}}-\delta_{\mathrm{pre}}\mathbf{u}_{\mathrm{pas}},
\qquad
p(t)=p_{\mathrm{pre}}+t\delta_{\mathrm{ins}}\mathbf{u}_{\mathrm{pas}}.
\]
Hanging templates approach a support edge or hook, establish support, and lower or release under gravity. Pouring rotates the grasped object toward a receiving region, while placement aligns with a support normal before release and retreat.

\paragraph{Scene-trajectory-conditioned templates.}
For a target base pose $q=(x,y,\psi)$, A* provides a coarse path on the occupancy or BEV grid. Because this seed does not encode turning radius, coupled base velocities, or other embodiment constraints, it is refined by a DWB-style local rollout optimizer before validation.

\subsection{Trajectory Optimization}
\label{supp:trajectory_optimization}

Optimization adapts each template to the asset and robot embodiment $\mathcal{R}$:
\[
(\theta_*^{(i)},\tau_*^{(i)})=
\arg\min_{\theta,\tau}
\mathcal{E}_s\!\left(
\theta,\tau;h,\phi(h),x^{(i)},\mathcal{A},\mathcal{R}
\right),
\]
with
\[
\mathcal{E}_s=
\lambda_{\mathrm{func}}E_{\mathrm{func}}+
\lambda_{\mathrm{task}}E_{\mathrm{task}}+
\lambda_{\mathrm{geom}}E_{\mathrm{geom}}+
\lambda_{\mathrm{contact}}E_{\mathrm{contact}}+
\lambda_{\mathrm{coll}}E_{\mathrm{coll}}+
\lambda_{\mathrm{kin}}E_{\mathrm{kin}}+
\lambda_{\mathrm{smooth}}E_{\mathrm{smooth}}.
\]
The active terms depend on the skill; the result is still a candidate and must pass physics validation.

\begin{table*}[htbp]
\centering
\small
\renewcommand{\arraystretch}{1.10}
\caption{Representative trajectory-optimization variables and objectives.}
\label{tab:supp_traj_opt_families}
\label{tab:supp_articulation_optimization}
\label{tab:supp_vector_optimization}
\label{tab:supp_nav_optimization}
\label{tab:supp_optimization_metadata}
\begin{tabularx}{\textwidth}{>{\raggedright\arraybackslash}p{0.19\textwidth}>{\raggedright\arraybackslash}p{0.23\textwidth}>{\raggedright\arraybackslash}p{0.31\textwidth}>{\raggedright\arraybackslash}X}
\toprule
Skill family & Optimized variables & Main objectives & Stored outputs \\
\midrule
Dexterous/contact & Hand pose, finger joints, contacts, approach & Region consistency, wrench support, collision, joint limits, self-collision & Final hand configuration, contact set, pre-grasp, and objective terms. \\
Garment/adaptive & Waypoints, fold/place targets, timing & Keypoint consistency, scale adaptation, reachability, smoothness, collision & Geometry-adapted bimanual trajectory and task parameters. \\
Articulation & Contact pose, waypoints, joint schedule, pull/push direction & Part tracking, contact maintenance, collision, reachability, joint progress & Refined opening, closing, rotation, or push trajectory. \\
Vector-conditioned & Alignment and execution poses, depth, release & Vector alignment, clearance, support stability, task progress & Refined insertion, hanging, pouring, or placement trajectory. \\
Scene-level & Base path, commands, orientation, approach pose & Path following, local dynamics, clearance, smoothness, manipulation reachability & Embodiment-feasible path and interaction-ready final pose. \\
\bottomrule
\end{tabularx}
\end{table*}

\paragraph{Dexterous and contact-aware optimization.}
Let $q_{\mathrm{hand}}$ and $T_{\mathrm{hand}}$ denote hand joints and root pose, and let $p_c=\mathrm{FK}_c(T_{\mathrm{hand}},q_{\mathrm{hand}})$ be a contact-link position. Contacts are kept on the grounded functional region through
\[
E_{\mathrm{region}}=
\sum_{c\in\mathcal{C}}\mathrm{dist}(p_c,h)^2.
\]
For a selected contact set $\mathcal{C}$, a wrench term can additionally encourage task-relevant support:
\[
E_{\mathrm{wrench}}=
\sum_j\left\|
 w_j-\sum_{c\in\mathcal{C}}G_cf_{c,j}
\right\|_2^2,
\qquad
f_{c,j}\in\mathcal{K}_{\mu_c}.
\]
Here, $w_j$ is a target wrench, $G_c$ maps a contact force to an object wrench, and $\mathcal{K}_{\mu_c}$ is the friction cone at contact $c$. Signed-distance, clearance, joint-limit, and self-collision penalties discourage penetration and infeasible configurations. The approach offset, direction, and intermediate waypoints are jointly refined to obtain a collision-aware approach-contact-close-retreat motion.

\paragraph{Geometry-adaptive waypoint optimization.}
For template waypoints $p_t^0$, we optimize corrections
\[
\{\Delta p_t\}_{t=1}^{T}=
\arg\min_{\{\Delta p_t\}}
\sum_t\|\Delta p_t\|_2^2+
\lambda_{\mathrm{geom}}E_{\mathrm{geom}}+
\lambda_{\mathrm{reach}}E_{\mathrm{reach}}+
\lambda_{\mathrm{smooth}}E_{\mathrm{smooth}},
\]
\[
p_t^*=p_t^0+\Delta p_t.
\]
Garment fold displacements and fling amplitudes are scaled by keypoint distances, bounded, and adjusted for bimanual reachability. This avoids over-folding short garments, under-folding long garments, and bimanual trajectories outside the reachable workspace.

\paragraph{Articulation refinement.}
For object joint state $q_t$, the end-effector tracks the ideal pose induced by the moving part:
\[
E_{\mathrm{art\text{-}track}}=
\sum_t\left\|
T_{\mathrm{eef},t}\ominus\widehat{T}_{\mathrm{eef}}(q_t)
\right\|^2.
\]
Contact-maintenance, collision, reachability, and smoothness terms refine the contact pose, waypoint spacing, pull direction, and joint schedule. Handle-based actions preserve grasp contact, while push-based closing and button actions maintain bounded surface contact along the valid motion direction.

\paragraph{Vector-conditioned refinement.}
For insertion, the alignment cost is
\[
E_{\mathrm{align}}=
1-(R\mathbf{u}_{\mathrm{act}})^{\top}(\mathbf{u}_{\mathrm{pas}}),
\]
and is combined with insertion-depth and collision terms. Hanging optimization adjusts support contact and release so the post-release configuration is gravity stable. Pouring refines rotation toward the receiving region subject to wrist and collision limits, and placement refines support-normal alignment, release height, and retreat direction.

\paragraph{Scene-level optimization.}
DWB-style optimization samples short-horizon controls under the robot's base model. Differential-drive, Ackermann, and holonomic embodiments use $(v,\omega)$, $(v,\delta)$, and $(v_x,v_y,\omega)$ controls, respectively. Rollouts are scored by
\[
J_{\mathrm{DWB}}=
 w_{\mathrm{path}}J_{\mathrm{path}}+
 w_{\mathrm{goal}}J_{\mathrm{goal}}+
 w_{\mathrm{obs}}J_{\mathrm{obs}}+
 w_{\mathrm{clear}}J_{\mathrm{clear}}+
 w_{\mathrm{smooth}}J_{\mathrm{smooth}}+
 w_{\mathrm{reach}}J_{\mathrm{reach}}.
\]
The terms score path tracking, goal error, obstacle collision, clearance, control smoothness, and terminal manipulation reachability. The reachability term favors terminal poses from which the target anchor is manipulable, using reachability maps, IK queries, or workspace heuristics. We store the optimized parameters and trajectory, the objective breakdown, embodiment-feasibility flags, functional-consistency status, and diversity descriptor for subsequent validation and sampling.

\subsection{Physics Validation}
\label{supp:physics_validation}

Physics validation is the final candidate-generation filter before entry into the annotation bank. Full-robot execution feasibility is evaluated separately during embodiment-specific candidate retrieval and execution. For an optimized candidate, we run skill-specific rollouts under validation embodiment $e_{\mathrm{val}}$ and perturbations $\Xi$:
\[
v^{(i)}=
\mathcal{V}_s\!\left(
 a_{h,s}^{(i)},\mathcal{A},e_{\mathrm{val}},\Xi
\right).
\]
The result records success, collisions, contact stability, task progress, robustness, rollout length, perturbations, and embodiment. Failed candidates are rejected.

\paragraph{Validation embodiments.}
Floating grippers or hands validate local object interaction without tying an annotation to one arm configuration. They are used for object-centric grasps, insertion, hanging, placement, pouring, and many handle-based articulation candidates. Non-floating validation is reserved for embodiment-dependent skills: full-manipulator validation checks selected long-horizon deformable interactions for IK, singularities, self-collision, and arm--object or arm--cloth interaction, while base-aware validation checks navigation for local dynamics, clearance, and manipulation reachability.

\begin{table*}[t]
\centering
\small
\renewcommand{\arraystretch}{1.10}
\caption{Validation settings used across skill families.}
\label{tab:supp_validation_embodiments}
\begin{tabularx}{\textwidth}{>{\raggedright\arraybackslash}p{0.20\textwidth}>{\raggedright\arraybackslash}p{0.26\textwidth}>{\raggedright\arraybackslash}p{0.29\textwidth}>{\raggedright\arraybackslash}X}
\toprule
Setting & Representative skills & Principal checks & Purpose \\
\midrule
Floating gripper & Parallel-jaw grasp, insertion, hanging, placement, handle articulation & Contact, penetration, local collision, stability, task progress, release outcome & Tests reusable object-centric interactions without over-constraining them to one arm pose. \\
Floating dexterous hand & Dexterous grasp and articulation & Joint validity, hand-object contact, self-collision, object stability, contact maintenance & Verifies that the hand actually constrains the object. \\
Full manipulator & Selected garment and long-horizon deformable skills & IK, singularity, arm/self-collision, entanglement, task outcome & Captures feasibility that depends on the arm and body. \\
Base-aware robot & Navigation and approach-pose annotation & Traversability, local dynamics, turning radius, clearance, path connectivity, manipulation reachability & Verifies embodiment-feasible paths and useful final base poses. \\
\bottomrule
\end{tabularx}
\end{table*}

\paragraph{Robustness tests.}
For grasp and dexterous-grasp candidates, we sample $K=8$ random gravity directions with increased magnitude,
\[
\mathbf{g}_k=1.5g_0\mathbf{u}_k,
\qquad
\mathbf{u}_k\sim\mathbb{S}^{2},
\qquad
k=1,\ldots,8,
\]
where $g_0$ is the nominal gravity magnitude. Depending on the skill, validation can also perturb object pose, gripper or hand pose, trajectory timing, or apply small external forces. These tests reject candidates that succeed only under a single idealized initialization or load direction.

\begin{table*}[t]
\centering
\small
\renewcommand{\arraystretch}{1.10}
\caption{Skill-specific validation criteria and representative detected failures.}
\label{tab:supp_validation_robustness}
\label{tab:supp_skill_validation_criteria}
\label{tab:supp_common_validation_failures}
\begin{tabularx}{\textwidth}{>{\raggedright\arraybackslash}p{0.18\textwidth}>{\raggedright\arraybackslash}p{0.35\textwidth}>{\raggedright\arraybackslash}X}
\toprule
Skill family & Required checks & Representative failure signal \\
\midrule
Parallel-jaw grasp & Collision-free approach, closure contact, stability, randomized gravity, disturbances & Object drops, slips, or rotates beyond tolerance; pre-grasp penetrates geometry. \\
Dexterous grasp & Joint limits, self-collision, penetration, contact coverage, stability, gravity robustness & Fingers surround the object visually but fail to constrain it, or penetrate/self-collide. \\
Articulation & Contact maintenance, joint progress, collision, reachability, grasp/push validity & Motion stalls, the gripper jams, or the hand loses contact or becomes trapped. \\
Insertion & Axis alignment, entry clearance, insertion depth, post-insertion stability & Collision at the entrance or insufficient depth. \\
Hang/place/pour & Alignment, clearance, release stability, task progress & Object falls after release, settles unstably, or collides during execution. \\
Garment/deformable & Full-arm IK, singularities, arm-cloth collision, entanglement, geometric outcome & Garment wraps around the arm, trajectory becomes unreachable, or intended state is not achieved. \\
Navigation & Traversability, clearance, connectivity, base dynamics, final-pose reachability & A* path is dynamically infeasible or the final pose cannot support manipulation. \\
\bottomrule
\end{tabularx}
\end{table*}

\paragraph{Parallel execution.}
Validation is parallelized across assets, skills, anchors, candidates, and perturbation trials. Jobs with similar simulation settings are batched by skill and embodiment, and failed rollouts terminate early after penetration, contact loss, object drop, stalled task progress, or joint-limit violation. This parallel filtering is necessary because candidate banks can be large and pass rates are low for several contact-rich and deformable skills.

\subsection{Physics-aware Augmentation}
\label{supp:physics_aware_augmentation}

After validation, we expand each bank using bounded local perturbations and symmetry-aware transforms while preserving $s$, $h$, and $\phi(h)$. For a validated candidate,
\[
\mathcal{U}(a_{h,s}^{(i)})=
\{\tilde{a}_{h,s}^{(i,j)}\}_{j=1}^{M_i},
\qquad
\widetilde{\mathcal{B}}_{h,s}=
\mathcal{B}_{h,s}\cup
\bigcup_{a_{h,s}^{(i)}\in\mathcal{B}_{h,s}}
\mathcal{U}(a_{h,s}^{(i)}).
\]

\paragraph{Local perturbations.}
We apply bounded, skill-dependent perturbations to targets, parameters, and waypoints:
\[
\begin{aligned}
\tilde{x}^{(i,j)}&=x^{(i)}+\Delta x^{(j)},\qquad
\tilde{\theta}^{(i,j)}=\theta^{(i)}\oplus\Delta\theta^{(j)},\\
\tilde{\tau}^{(i,j)}&=\{T_t\oplus\Delta T_t^{(j)}\}_{t=1}^{T}.
\end{aligned}
\]
Surface targets remain inside the same part or affordance region, keypoint targets remain inside local confidence regions, and navigation targets remain in traversable reachable space. Perturbed quantities can include grasp center and yaw, palm pose and finger seeds, fold or fling targets, handle contact and pull direction, insertion alignment, support contact, or base pose. The functional affordance and task intent are preserved.

\paragraph{Symmetry-aware augmentation.}
When visual-language annotations indicate a valid symmetry group $\mathcal{G}_{\mathrm{sym}}(o)$, we transform the target, parameters, trajectory, and relevant vectors consistently:
\[
\tilde{a}_{h,s}^{(i,g)}=g\cdot a_{h,s}^{(i)},
\qquad
g\in\mathcal{G}_{\mathrm{sym}}(o),
\]
\[
\tilde{T}=GT,
\qquad
\tilde{\tau}=\{GT_1,\ldots,GT_T\},
\qquad
\tilde{\mathbf{u}}=R_G\mathbf{u}.
\]
Here, $G$ is the rigid transform induced by $g$ and $R_G$ is its rotational component. Examples include rotating bottle grasps around a valid symmetry axis, transferring candidates across repeated equivalent handles, or mirroring bilateral garment folds. Symmetry is applied only when it preserves the functional affordance; a mug handle grasp, for example, is not rotated onto the mug body.

\paragraph{Feasibility rechecks.}
Augmented candidates are accepted only when
\[
\mathbb{I}_{\mathrm{accept}}=
\mathbb{I}\!
\left[
C_{\mathrm{region}}\wedge
C_{\mathrm{func}}\wedge
C_{\mathrm{coll}}\wedge
C_{\mathrm{kin}}\wedge
C_{\mathrm{task}}
\right].
\]
The checks enforce region membership, functional consistency, collision and clearance, embodiment constraints, and task-specific validity. Inexpensive pose-level augmentations use geometry and short-rollout physics
checks, whereas high-risk full-manipulator deformable trajectories are
returned to full physics validation before retention.

\begin{table*}[htbp]
\centering
\small
\renewcommand{\arraystretch}{1.10}
\caption{Representative physics-aware augmentation strategies and required rechecks.}
\label{tab:supp_common_augmentation}
\begin{tabularx}{\textwidth}{>{\raggedright\arraybackslash}p{0.19\textwidth}>{\raggedright\arraybackslash}p{0.27\textwidth}>{\raggedright\arraybackslash}p{0.28\textwidth}>{\raggedright\arraybackslash}X}
\toprule
Skill/object & Perturbation or transform & Constraint & Recheck \\
\midrule
Parallel-jaw grasp & Contact center, yaw, width, pre-grasp offset & Remain on the same graspable region and grasp family & Collision, width, approach, and stability. \\
Dexterous grasp & Palm pose, contact seeds, joint offsets, closure timing & Preserve functional region and feasible hand configuration & Joint limits, self-collision, penetration, and object constraint. \\
Garment fold/fling & Keypoints, fold line, place target, lift/stretch amplitude & Stay within confidence regions and preserve fold/fling intent & Bimanual reachability, cloth clearance, entanglement, and outcome. \\
Articulation & Handle point, grasp orientation, pull direction, waypoint spacing, joint range & Preserve valid joint direction and contact mode & Contact maintenance, joint progress, collision, and jamming. \\
Insertion and hanging & Entry or support pose, depth, axis, release pose & Preserve alignment or gravity-stable support & Clearance, task progress, and post-release stability. \\
Navigation & $(x,y,\psi)$ around an approach pose & Remain in traversable, reachable free space and face the target & Connectivity, DWB feasibility, clearance, and manipulation reachability. \\
Bottle/round knob & Rotation around annotated symmetry axis & Transform lies in a functionally equivalent region & Collision, stability, rotation direction, and wrist feasibility. \\
Repeated handles / bilateral garments & Copy, translation, or mirror transform & Source and target regions are functionally equivalent & Local collision context, reachability, and task validity. \\
\bottomrule
\end{tabularx}
\end{table*}

Physics-aware augmentation therefore increases diversity around validated functional anchors without treating geometric transformations as automatically valid labels.

\section{Primitive-specific Action Annotation}
\label{sec:primitive_specific_annotation}

This section instantiates the unified candidate-bank schema for five primitive families: parallel-jaw and dexterous grasping, articulation, insertion and hanging, garment manipulation, and navigation. Each candidate
\[
a_{h,s}^{(i)}=(s,o(h),h,\phi(h),x^{(i)},\theta^{(i)},\tau^{(i)},v^{(i)},d^{(i)})
\]
stores a grounded target, primitive-specific parameters, optional waypoints, validation metadata, and diversity attributes. All annotators follow the same sequence of target generation, pose or trajectory construction, optional optimization, physics validation, and serialization.

\paragraph{Parallel-jaw and dexterous grasp annotation.}
Both grasp annotators use geometry-based proposals followed by simulation validation; dexterous grasps additionally optimize wrist pose, hand joints, and semantic contacts.

\textbf{Parallel-jaw proposals and poses.}
We convert each USD asset to a triangle mesh, sample $10^5$ area-weighted surface points, and construct each local neighborhood from its $K=64$ nearest samples. Rim-style bimanual candidates are prefiltered from the top $30\%$ of object height and finally selected from the top $20\%$ band. Each point $i$ receives
\[
S_i=\mathbf{1}_{\mathrm{top}}+0.4E_i+1.5B_i+0.8P_i+0.8A_i,
\]
where $E_i$, $B_i$, $P_i$, and $A_i$ measure normal variation, bilateral rim support, XY convex-hull proximity, and outward accessibility, and $\mathbf{1}_{\mathrm{top}}$ indicates membership in the desired upper band. We retain anchors satisfying
\[
P_i>0.1,\qquad A_i>0.05,\qquad w_i\leq0.06\mathrm{m},\qquad
\angle(n_i,\hat z)\leq60^\circ,
\]
where $w_i$ is the estimated local gripper width and $n_i$ is the surface normal. We then apply $0.005\mathrm{m}$ non-maximum suppression and keep at most 500 anchors.

For anchor position $p_a$ and normal $n_a$, the local frame is
\[
z_a=n_a,\qquad
x_a=\operatorname{PCA}_{\mathrm{tangent}}(\mathcal{N}(p_a)),\qquad
y_a=z_a\times x_a,
\]
with $R_a=[x_a,y_a,z_a]$, where $x_a$ is estimated by PCA over the local XY neighborhood. The factor $R_x(\pi)$ below converts this frame to the gripper convention. We evaluate up to 3000 opposite-side pairs and score anchor quality, along-rim consistency, across-object separation, center balance, and tangent consistency. For each retained anchor,
\[
R_g=R_aR_x(\pi),\qquad
p_g=p_a+(0.14\mathrm{m})n_a,\qquad
p_{\mathrm{pre}}=p_g+(0.20\mathrm{m})n_a.
\]
Pairs whose transformed gripper AABBs overlap are rejected before simulation.

\textbf{Parallel-jaw validation.}
The asset is converted to a rigid body and floating grippers are simulated in Isaac Sim with static and dynamic friction set to 2.0. Gravity is disabled during approach and closure. Grippers start at finger target $0.04\mathrm{m}$, approach for 80 steps, close to $0.0\mathrm{m}$ over 32 steps, hold for 50 steps, then lift the object by $1.0\mathrm{m}$ over 200 steps with gravity enabled. Let $z_{\mathrm{close}}$ be the object height after closure and hold, and $z_{\mathrm{lift}}$ its final height after lifting. A pair passes when
\[
z_{\mathrm{lift}}-z_{\mathrm{close}}\geq0.95\mathrm{m}.
\]
Accepted single-gripper poses are stored as $[x,y,z,q_w,q_x,q_y,q_z]$ with anchor and validation metadata; bimanual candidates additionally store paired left/right poses and pairing metadata. The output file is \texttt{Annotation/bi\_gripper\_grasp\_pose.json}.

\textbf{Dexterous proposals and optimization.}
Dexterous proposals are category-conditioned: category 1 targets top rims or edges, category 2 lower-edge support, category 3 bimanual side holds, and category 4 convex side or top holds. Each category uses its own height, normal, width, perimeter, accessibility, and suppression rules. The object is voxelized into an SDF $d:\mathbb{R}^3\rightarrow\mathbb{R}$ at $5\mathrm{mm}$ resolution with 8 padding voxels, using $d(x)>0$ outside the object and $d(x)<0$ for penetration. The hand is represented by semantically labeled collision spheres for thumb, finger pads, palm support, and avoid regions. Category-specific surface patches provide contact targets and desired contact sides.

For each hand, we optimize
\[
x=[t_x,t_y,t_z,\omega_x,\omega_y,\omega_z,q_1,\ldots,q_m],
\qquad R=R_0\operatorname{Exp}(\omega),
\]
under bounded wrist and joint updates, where $t$, $\omega$, and $q$ are the wrist translation, rotation-vector update, and hand joint angles, and $R_0$ is the seed wrist orientation. The staged objective is
\[
J_s=J_{\mathrm{clear}}+J_{\mathrm{target}}+J_{\mathrm{normal}}+J_{\mathrm{opp}}
+J_{\mathrm{coll}}+J_{\mathrm{self}}+J_{\mathrm{prior}}+J_{\mathrm{cat}}.
\]
For contact sphere center $c_i$, radius $r_i$, activity weight $\lambda_i$, and desired clearance $\epsilon_i$,
\[
J_{\mathrm{clear}}=\sum_i w_c\lambda_i(d(c_i)-r_i-\epsilon_i)^2.
\]
Using the SDF projection $\pi(c_i)=c_i-d(c_i)\hat\nabla d(c_i)$, where $\hat\nabla d$ is the normalized SDF gradient, target and normal alignment are
\[
J_{\mathrm{target}}=
\sum_i w_p\lambda_i\left\|D_sR_c^\top(\pi(c_i)-p_i^\star)\right\|^2,
\qquad
J_{\mathrm{normal}}=
\sum_i w_n\lambda_i(1-\hat\nabla d(c_i)^\top n_i^\star)^2.
\]
Here $p_i^\star$ and $n_i^\star$ are the assigned target point and normal, $R_c$ is the local contact frame, and $D_s$ is a stage-dependent diagonal weighting matrix. For paired contacts,
\[
J_{\mathrm{opp}}=
\sum_{(i,j)}w_o\left(1-
\frac{c_j-c_i}{\|c_j-c_i\|}^\top o_{ij}^\star\right)^2,
\]
where $o_{ij}^\star$ is the desired opposition direction. Object and self-collision use hinge penalties,
\[
J_{\mathrm{coll}}=
\sum_j w_{\mathrm{coll},j}[m_j-(d(c_j)-r_j)]_+^2,
\qquad
J_{\mathrm{self}}=
\sum_{(i,j)}w_{\mathrm{self}}[r_i+r_j+m_{\mathrm{self}}-\|c_i-c_j\|]_+^2,
\]
where $m_j$ is the required object-clearance margin and $m_{\mathrm{self}}$ is the self-collision margin. We use $4\mathrm{mm}$ clearance for inactive spheres, $10\mathrm{mm}$ for palm spheres, and a $2\mathrm{mm}$ self-collision margin. Seed regularization relative to $(t_0,R_0,q_0)$ is
\[
J_{\mathrm{prior}}=w_t\|t-t_0\|^2+w_R\|\log(R_0^\top R)\|^2+w_q\|q-q_0\|^2,
\]
and $J_{\mathrm{cat}}$ encodes category-specific contact priors.

We solve with bounded L-BFGS-B for at most 140 iterations, $f_{\mathrm{tol}}=g_{\mathrm{tol}}=10^{-8}$, finite-difference step $10^{-4}$, and at most 80 line-search steps. Pregrasp, grasp, and squeeze stages progressively increase contact, target, and collision weights. Category 3 additionally performs a fast translation and wrist-roll pose search before refinement.

Category 1 and 2 solutions are paired across opposite edge anchors, while category 3 solutions are paired across opposite side anchors with similar height and cross-section. We reject pairs with overlapping hand AABBs, wrist separation below $0.10\mathrm{m}$, inconsistent side-hold directions, or excessive combined penetration. Each pair stores
\[
\theta^{(i)}=(T_L,q_L,\mathcal{C}_L,T_R,q_R,\mathcal{C}_R),
\]
where $T$, $q$, and $\mathcal{C}$ denote wrist pose, hand joints, and semantic contact assignments.
Floating-hand validation executes pregrasp, grasp, squeeze, settle, gravity hold, and a $0.3\mathrm{m}$ lift. A candidate passes if the object remains supported and drops by at most $2\mathrm{cm}$ during lift and hold.

\paragraph{Articulation waypoint annotation.}
We recover the USD joint model, generate a primitive-specific contact pose, analytically map joint motion to end-effector waypoints, and validate the trajectory in simulation.

\textbf{Joint and contact extraction.}
After fixing the asset base to the world, we discover each revolute or prismatic joint and store
\[
\mathcal{J}=(j,\mathrm{type},a,c,q_{\min},q_{\max},b_{\mathrm{parent}},b_{\mathrm{child}}),
\]
where $j$ identifies the joint, $\mathrm{type}$ is revolute or prismatic, $a$ and $c$ are the world-space axis and pivot, and $b_{\mathrm{parent}},b_{\mathrm{child}}$ are the connected bodies; revolute limits are converted to radians. Let $q_0$ denote the initial joint state. Let $q_{\mathrm{open}}\in\{q_{\min},q_{\max}\}$ denote the limit in the
selected opening direction. We set
\[
q^\star=
\operatorname{clip}\!\left(
q_0+0.7(q_{\mathrm{open}}-q_0),
q_{\min},q_{\max}
\right),
\qquad
\Delta q^\star=q^\star-q_0.
\]
Closing analogously moves toward the selected closed limit, optionally with
a small overshoot before the final target is clamped to
$[q_{\min},q_{\max}]$.

Open-by-handle restricts grasp generation to handle geometry and filters approach, accessibility, collision, and panel-normal consistency. Open-on-edge samples outer child-link edges far from the hinge to increase moment arm. Close-by-push samples inset points on flat outer panels, again favoring large hinge distance, and aligns the pushing direction with closure. Each contact stores $T_0=(p_0,R_0)$ plus joint, child-link, approach, and contact-mode metadata.

\textbf{Analytic waypoints and validation.}
For a revolute joint with displacement $\Delta\theta^\star$ and uniformly sampled fractions $\alpha_k\in[0,1]$,
\[
G_k=\operatorname{Exp}(\alpha_k\Delta\theta^\star[a]_\times),\qquad
p_k=c+G_k(p_0-c),\qquad R_k=G_kR_0.
\]
For a prismatic joint with displacement $\Delta d^\star$,
\[
p_k=p_0+\alpha_k\Delta d^\star a,\qquad R_k=R_0.
\]
We prepend
\[
p_{\mathrm{pre}}=p_0-\delta_{\mathrm{app}}u_{\mathrm{app}},
\]
where $u_{\mathrm{app}}$ is the approach direction and $\delta_{\mathrm{app}}$ is a primitive- and geometry-dependent distance from $0.08$ to $0.70\mathrm{m}$. Typical rollouts use 100--200 approach or motion steps and 30--64 closing steps. If $q_T$ is the measured final joint state, a candidate passes when
\[
|q_T-q_0|\geq0.95|\Delta q^\star|
\]
and moves in the intended direction. Accepted waypoints are serialized as $[x,y,z,q_w,q_x,q_y,q_z]$ and stored with joint properties, contact and approach parameters, primitive type, and validation metadata.

\paragraph{Insertion and hanging annotation.}
These pairwise skills use active and passive anchors. For $s\in\{\mathrm{insert},\mathrm{hang}\}$,
\[
\mathcal{B}_{(h_{\mathrm{act}},h_{\mathrm{pas}}),s}
=\{a_{(h_{\mathrm{act}},h_{\mathrm{pas}}),s}^{(i)}\}_{i=1}^{N_{(h_{\mathrm{act}},h_{\mathrm{pas}}),s}},
\]
where each candidate stores both objects, anchors, affordances, targets, parameters, trajectory, validation, and diversity metadata. Functional and geometric compatibility determines valid pairings.

\begin{table*}[t]
\centering
\small
\renewcommand{\arraystretch}{1.15}
\caption{Active and passive roles for insertion and hanging. The passive side provides the receiving or supporting affordance, while the active side provides the object-side affordance that must be aligned or matched.}
\label{tab:supp_insert_hang_active_passive}
\begin{tabularx}{\textwidth}{p{0.14\textwidth}p{0.20\textwidth}p{0.24\textwidth}X}
\toprule
Skill & Passive side & Active side & Annotation goal \\
\midrule
Insertion &
Socket, hole, port, opening, container mouth, receptacle &
Peg, plug, cable connector, key, tool tip, object to be inserted &
Annotate where insertion is possible on the passive object and where the active object should enter, together with their insertion directions. \\

Hanging &
Hook, peg, rod, rail, support edge, hanger slot &
Mug handle, bag strap, clothes hanger loop, garment collar point, object loop &
Annotate where support is possible on the passive object and which active object region should engage with it. For rigid objects, vectors are aligned; for deformable objects, keypoints are matched. \\
\bottomrule
\end{tabularx}
\end{table*}

\textbf{Insertion.}
The passive frame is
\[
F_{\mathrm{pas}}^{\mathrm{ins}}=(p_{\mathrm{pas}},\mathbf{u}_{\mathrm{pas}},r_{\mathrm{pas}},\ell_{\mathrm{pas}},\mathrm{type}_{\mathrm{pas}}),
\]
and the active frame is
\[
F_{\mathrm{act}}^{\mathrm{ins}}=(p_{\mathrm{act}},\mathbf{u}_{\mathrm{act}},r_{\mathrm{act}},\ell_{\mathrm{act}},\mathrm{type}_{\mathrm{act}}).
\]
Here $p$ denotes the entry or tip, $\mathbf{u}$ the positive physical insertion direction, $r$ a cross-section descriptor (shown as a scalar size in the compatibility condition below), $\ell$ the available depth or effective length, and \texttt{type} an optional semantic connector type. Under this convention, a candidate orientation $R^{(i)}$ should satisfy
\[
R^{(i)}\mathbf{u}_{\mathrm{act}}\approx\mathbf{u}_{\mathrm{pas}};
\]
passive outward normals are sign-flipped before this check. Compatibility requires
\[
\mathrm{type}_{\mathrm{act}}\sim\mathrm{type}_{\mathrm{pas}},\qquad
r_{\mathrm{act}}+\epsilon_{\mathrm{clear}}\leq r_{\mathrm{pas}},\qquad
\angle(R^{(i)}\mathbf{u}_{\mathrm{act}},\mathbf{u}_{\mathrm{pas}})\leq\epsilon_{\mathrm{angle}}.
\]
Here $\epsilon_{\mathrm{clear}}$ and $\epsilon_{\mathrm{angle}}$ are clearance and angular tolerances. The approach-align-insert trajectory uses
\[
p_{\mathrm{pre}}=p_{\mathrm{pas}}-\delta_{\mathrm{pre}}\mathbf{u}_{\mathrm{pas}},\qquad
p_{\mathrm{ins}}=p_{\mathrm{pas}}+\delta_{\mathrm{ins}}\mathbf{u}_{\mathrm{pas}}.
\]
Here $\delta_{\mathrm{pre}}$ is the pre-entry offset and $\delta_{\mathrm{ins}}$ is the target insertion depth. Parameters store alignment, direction, depth, and clearance; validation records alignment error, penetration, achieved depth, and post-insertion stability.

\begin{table*}[t]
\centering
\small
\renewcommand{\arraystretch}{1.15}
\caption{Examples of insertion annotation. The passive object provides the receiving location and direction, while the active object provides the insertion tip and positive insertion direction.}
\label{tab:supp_insertion_examples}
\begin{tabularx}{\textwidth}{p{0.16\textwidth}p{0.23\textwidth}p{0.23\textwidth}X}
\toprule
Example & Passive annotation & Active annotation & Candidate meaning \\
\midrule
HDMI insertion &
HDMI port center, inward insertion direction, port size, port type &
HDMI plug tip, plug-forward direction, plug size, plug type &
Align the plug-forward vector with the port insertion direction and translate the plug into the port. \\

Peg-in-hole &
Hole center, hole axis, hole radius, hole depth &
Peg tip, peg axis, peg radius, peg length &
Align the peg axis with the hole axis and insert to a feasible depth with clearance. \\

Key insertion &
Keyhole entry point, keyhole direction, slot type &
Key tip, key-forward direction, key blade geometry &
Match key type and direction, align the key blade, and move along the keyhole axis. \\

Tool insertion &
Receptacle opening, insertion axis, opening size &
Tool tip or shaft endpoint, shaft direction, shaft diameter &
Align the tool shaft with the receptacle and insert without collision. \\
\bottomrule
\end{tabularx}
\end{table*}

\textbf{Hanging.}
For rigid supports,
\[
F_{\mathrm{pas}}^{\mathrm{hang}}=(p_{\mathrm{sup}},\mathbf{u}_{\mathrm{sup}},\mathbf{g},c_{\mathrm{clear}},\mathrm{type}_{\mathrm{sup}}),
\]
where $p_{\mathrm{sup}}$ and $\mathbf{u}_{\mathrm{sup}}$ specify the support point and entry direction, $\mathbf{g}$ is the gravity direction, and $c_{\mathrm{clear}}$ and $\mathrm{type}_{\mathrm{sup}}$ encode clearance and support type. The active loop or handle is
\[
F_{\mathrm{act}}^{\mathrm{hang}}=(p_{\mathrm{loop}},\mathbf{u}_{\mathrm{loop}},r_{\mathrm{loop}},\mathrm{type}_{\mathrm{loop}}),
\]
where $p_{\mathrm{loop}}$, $\mathbf{u}_{\mathrm{loop}}$, and $r_{\mathrm{loop}}$ specify the loop center, through-opening direction, and clearance. A candidate orientation $R^{(i)}$ aligns $R^{(i)}\mathbf{u}_{\mathrm{loop}}\approx\mathbf{u}_{\mathrm{sup}}$, engages the support, lowers under gravity, and releases. For deformable objects, the active side is instead
\[
F_{\mathrm{act,def}}^{\mathrm{hang}}=(p_{\mathrm{kp}},\mathrm{type}_{\mathrm{kp}},c_{\mathrm{conf}}),
\]
where $p_{\mathrm{kp}}$, $\mathrm{type}_{\mathrm{kp}}$, and $c_{\mathrm{conf}}$ denote the keypoint position, semantic role, and confidence. A semantic keypoint such as a collar or loop is then matched to the support without assuming a rigid active vector. The trajectory is
\[
\text{approach}\rightarrow\text{engage support}\rightarrow\text{lower under gravity}\rightarrow\text{release or retreat}.
\]

\begin{table*}[t]
\centering
\small
\renewcommand{\arraystretch}{1.15}
\caption{Examples of hanging annotation. Rigid objects use vector or frame alignment, while deformable objects can use keypoint-to-support matching.}
\label{tab:supp_hanging_examples}
\begin{tabularx}{\textwidth}{p{0.16\textwidth}p{0.23\textwidth}p{0.23\textwidth}X}
\toprule
Example & Passive annotation & Active annotation & Candidate meaning \\
\midrule
Mug on hook &
Hook point or support curve, hook entry direction, local clearance &
Mug handle center, vector through handle opening, handle clearance &
Align the handle-through vector with the hook direction, move the handle onto the hook, and release after gravity seating. \\

Bag on peg &
Peg tip or support edge, support direction, clearance &
Bag strap centerline or loop vector, strap opening size &
Place the strap loop over the peg and lower until the bag is supported. \\

Clothes hanger on rod &
Rod support line, approach side, gravity direction &
Hanger hook center and hook opening direction &
Move the hanger hook over the rod, lower onto the rod, and release. \\

\bottomrule
\end{tabularx}
\end{table*}

\textbf{Validation and metadata.}
Insertion checks semantic type, vector alignment, opening clearance, approach collision, achieved depth, and post-insertion alignment. Hanging checks support compatibility, clearance, contact formation, post-release gravity stability, and, for deformables, entanglement and over-stretching. We store
\[
v^{(i)}=\{\mathrm{align\_err},\mathrm{clearance},\mathrm{collision},\mathrm{depth},\mathrm{stability},\mathrm{release\_success},\mathrm{failure\_reason}\},
\]
while $d^{(i)}$ records angle, depth, approach offset, support point, loop orientation, keypoint choice, and release height.

\begin{table*}[t]
\centering
\small
\renewcommand{\arraystretch}{1.15}
\caption{Common validation checks and failure cases for insertion and hanging annotations.}
\label{tab:supp_insert_hang_validation_failures}
\begin{tabularx}{\textwidth}{p{0.16\textwidth}p{0.24\textwidth}p{0.26\textwidth}X}
\toprule
Skill & Check & Common failure case & Rejection criterion \\
\midrule
Insertion &
Semantic compatibility &
Plug and port types do not match, or active/passive roles are reversed. &
Connector type, category, or functional affordance is incompatible. \\

Insertion &
Vector alignment &
The active insertion direction is tilted relative to the passive insertion direction. &
Alignment error exceeds tolerance. \\

Insertion &
Clearance &
The active object is larger than the opening or socket. &
Cross-section plus clearance margin exceeds passive opening size. \\

Insertion &
Trajectory collision &
The active object collides with the rim, socket wall, or surrounding geometry. &
Collision or penetration exceeds tolerance before reaching required depth. \\

Insertion &
Insertion depth &
The object enters only shallowly or slips out immediately. &
Insertion depth or post-insertion stability is below threshold. \\

Hanging &
Support compatibility &
The active loop or handle cannot engage the passive support. &
Loop size, hook size, or support type is incompatible. \\

Hanging &
Vector or keypoint matching &
Rigid loop direction does not align with hook direction, or deformable keypoint misses the hook. &
Alignment error or keypoint-to-support distance exceeds tolerance. \\

Hanging &
Gravity stability &
Object slides off, rotates away, or falls after release. &
Post-release support is not maintained under gravity. \\

Hanging &
Clearance &
Object collides with hook, wall, rail, or nearby geometry during approach. &
Collision or insufficient clearance is detected. \\

Deformable hanging &
Entanglement or over-stretching &
Garment twists around the hook, rail, gripper, or arm. &
Cloth state violates deformation, support, or entanglement criteria. \\
\bottomrule
\end{tabularx}
\end{table*}

\paragraph{Garment and deformable-object trajectory annotation.}
Garment actions use semantic keypoints in a canonical garment frame and type-specific bimanual templates. The detected set $\mathcal{K}_{\mathrm{garment}}=\{k_1,\ldots,k_M\}$ depends on whether the asset is an upper-body garment, dress, pants, or rectangular cloth.

\begin{table*}[t]
\centering
\small
\renewcommand{\arraystretch}{1.15}
\caption{Garment keypoints used for deformable-object trajectory annotation. Keypoints are detected in a canonical garment frame and then converted back to the simulation or world frame for trajectory generation.}
\label{tab:supp_garment_keypoints}
\begin{tabularx}{\textwidth}{p{0.20\textwidth}p{0.32\textwidth}X}
\toprule
Garment type & Keypoints & Usage \\
\midrule
Upper-body garment / T-shirt &
\texttt{top\_left}, \texttt{top\_right}, \texttt{left\_shoulder}, \texttt{right\_shoulder}, \texttt{bottom\_left}, \texttt{bottom\_right}, \texttt{top\_point}, \texttt{bottom\_point}, \texttt{middle\_point} &
Sleeve folding, bottom-to-shoulder folding, fling from sleeve or bottom keypoints, bimanual lifting and stretching. \\

Dress &
Same as upper-body garment, with longer body extent &
Sleeve folding and bottom-up folding, with fold displacement adapted to garment length. \\

Pants &
\texttt{top\_left}, \texttt{top\_right}, \texttt{bottom\_left}, \texttt{bottom\_right}, \texttt{top\_point}, \texttt{bottom\_point} &
Side folding, leg folding, bottom-to-waist folding, compacting fold, bimanual lifting. \\

Plane / rectangular cloth &
\texttt{corner\_0}, \texttt{corner\_1}, \texttt{corner\_2}, \texttt{corner\_3} &
Corner grasping, folding, stretching, placement, and retrieval-style deformable manipulation. \\
\bottomrule
\end{tabularx}
\end{table*}

\textbf{Folding and flinging templates.}
For upper-body garments, $k_{\mathrm{lslv}},k_{\mathrm{rslv}}$ denote the \texttt{top\_left} and \texttt{top\_right} sleeve endpoints; $k_{\mathrm{lsh}},k_{\mathrm{rsh}}$ and $k_{\mathrm{lb}},k_{\mathrm{rb}}$ denote the shoulder points and bottom corners. Writing $\ell(a,b)$ for the line through keypoints $a,b$ and $\mathrm{Ref}_{\ell}$ for reflection across that line, the sleeve endpoints are reflected across shoulder--bottom fold lines,
\[
\ell_{\mathrm{L}}=\ell(k_{\mathrm{lsh}},k_{\mathrm{lb}}),\qquad
\ell_{\mathrm{R}}=\ell(k_{\mathrm{rsh}},k_{\mathrm{rb}}),
\]
\[
k_{\mathrm{lslv}}^\star=\mathrm{Ref}_{\ell_{\mathrm{L}}}(k_{\mathrm{lslv}}),\qquad
k_{\mathrm{rslv}}^\star=\mathrm{Ref}_{\ell_{\mathrm{R}}}(k_{\mathrm{rslv}}).
\]
A second stage nominally moves each bottom corner to its corresponding shoulder point, with the practical targets shifted or clamped as needed for stable bimanual execution. Pick points are shifted slightly inward from boundaries, and targets are clamped to each arm's side of the garment midline to reduce hand-hand collision.

For pants, $k_{\mathrm{top}}$ and $k_{\mathrm{bottom}}$ are the \texttt{top\_point} and \texttt{bottom\_point} centerline keypoints. The centerline $\ell_{\mathrm{mid}}=\ell(k_{\mathrm{top}},k_{\mathrm{bottom}})$ guides a side-to-center fold, followed by a bottom-to-waist fold and optional additional compacting fold. Targets adapt to leg length, waistband width, and bimanual reachability.

Flinging selects paired sleeve, bottom-corner, or shoulder keypoints. Left/right assignment follows world-frame lateral position. The synchronized sequence reaches, closes, lifts, moves forward, drops, and releases; lift height, forward displacement, drop distance, and drop height are stored as adjustable parameters.

\begin{table*}[t]
\centering
\small
\renewcommand{\arraystretch}{1.15}
\caption{Garment and deformable-object trajectory templates. Each template is generated from semantic keypoints and then validated with full-manipulator simulation when arm reachability or cloth entanglement is important.}
\label{tab:supp_garment_trajectory_templates}
\begin{tabularx}{\textwidth}{p{0.13\textwidth}p{0.24\textwidth}p{0.24\textwidth}X}
\toprule
Skill & Keypoints / anchors & Geometric rule & Generated trajectory \\
\midrule
Sleeve fold &
Sleeve endpoints, shoulders, bottom corners &
Use the shoulder--bottom line as the fold line and reflect each sleeve endpoint inward. &
Pick sleeve endpoints, lift, move to reflected targets, drop, release, and retract. \\

Bottom-to-shoulder fold &
Bottom corners and shoulder points &
Move bottom corners toward the corresponding shoulder region. &
Pick bottom corners, lift, fold upward toward shoulders, drop, release, and retract. \\

Pants side fold &
Left/right leg or side keypoints and centerline &
Fold side or leg regions toward the centerline. &
Pick side or cuff keypoints, lift, move inward, and place to overlap the two sides. \\

Pants compact fold &
Bottom/cuff keypoints and waistband region &
Fold the lower part upward toward the waistband; optionally repeat. &
Pick bottom/cuff points, lift, fold upward, place, and optionally perform another compacting fold. \\

Fling &
A paired keypoint group, e.g., sleeves, bottom corners, or shoulders &
Lift the two selected keypoints and move them forward with bimanual coordination. &
Reach, close grippers, lift, fling forward, drop, and release. \\

Plane / cloth fold &
Four corner keypoints &
Use selected corners or edges to define fold lines and place targets. &
Pick corners, lift, fold across a selected axis, place, and release. \\

Retrieval / washing / pushing &
Keypoints, edges, or local cloth regions &
Use keypoint or region targets together with full-arm motion constraints. &
Generate task-specific push, sweep, pull, or retrieval trajectories and validate with the full manipulator. \\
\bottomrule
\end{tabularx}
\end{table*}

\textbf{Bimanual representation and validation.}
Each waypoint stores paired end-effector poses and synchronized gripper commands,
\[
T_t=(T_t^{\mathrm{R}},T_t^{\mathrm{L}}),\qquad
u_t^{\mathrm{grip}}=(u_t^{\mathrm{R}},u_t^{\mathrm{L}}),
\]
\[
\tau^{(i)}=[(T_1,u_1^{\mathrm{grip}}),\ldots,(T_T,u_T^{\mathrm{grip}})].
\]
Trajectory distances are adapted from keypoint geometry, including sleeve length, hem-to-shoulder distance, and selected-pair separation. Full-manipulator validation is used when arm feasibility matters; trajectories are rejected for IK failure or singularity, large joint jumps, arm--cloth or table collision, entanglement, or failure to reach the intended geometric outcome. Metadata records IK, collisions, entanglement, keypoint displacement, completion, and failure reason.

\begin{table*}[t]
\centering
\small
\renewcommand{\arraystretch}{1.15}
\caption{Representative waypoint phases for garment skills. The same candidate-bank schema stores both the keypoint anchors and the generated bimanual trajectory.}
\label{tab:supp_garment_waypoint_phases}
\begin{tabularx}{\textwidth}{p{0.18\textwidth}p{0.24\textwidth}X}
\toprule
Skill & Phase order & Description \\
\midrule
Upper-body fold &
Pre-reach, reach, close, lift, move, drop, open, retract for sleeve; then repeat for bottom &
First folds sleeves inward using reflected targets, then folds bottom corners toward shoulder targets. \\

Fling &
Reach, close, lift, fling forward, drop, open &
Grasps a pair of keypoints, lifts the cloth, moves forward to spread it, drops, and releases. \\

Pants fold &
Reach, close, lift, side-fold move, drop, release; then bottom-fold move and release &
First aligns legs or sides, then folds the lower part toward the waistband. \\

Plane / cloth fold &
Reach, close, lift, fold across selected axis, drop, release &
Uses corner keypoints to define a fold axis and target corner positions. \\
\bottomrule
\end{tabularx}
\end{table*}

\paragraph{Navigation target and approach-pose annotation.}
Navigation candidates are interaction-ready base poses in traversable free space,
\[
x_{\mathrm{nav}}^{(i)}=q_{\mathrm{base}}^{(i)}=(x^{(i)},y^{(i)},\psi^{(i)}),
\]
stored in $\mathcal{B}_{h_{\mathrm{scene}},s_{\mathrm{nav}}}$ for scene-level anchor $h_{\mathrm{scene}}$.

\textbf{Object-centric target construction.}
For target object center $O$, orientation $R$, and current base position $B$, define
\[
\mathbf{u}=\frac{O-B}{\|O-B\|_2},\qquad
b_\ell=R^\top(B-O),\qquad
\mathbf{u}_\ell=R^\top\mathbf{u}.
\]
We intersect this local ray with the object's local AABB, obtain the entered-face center $f_\ell$ and outward normal $n_\ell$, and transform them back:
\[
f_w=O+Rf_\ell,\qquad n_w=Rn_\ell.
\]
The goal and yaw are
\[
g=f_w+\rho n_w,\qquad
\psi=\operatorname{atan2}((-n_w)_y,(-n_w)_x),
\]
with a face- or object-facing fallback when the horizontal normal is nearly vertical. If the local box or ray intersection is unavailable,
\[
g_{\mathrm{fallback}}=O-(r_{\mathrm{aabb}}(\mathbf{u})+\rho)\mathbf{u},
\]
where $r_{\mathrm{aabb}}(\mathbf{u})$ is the world-AABB extent of the object along $\mathbf{u}$.
Diversity is obtained by varying target object or part, approach side, standoff $\rho$, and yaw offset. Each candidate stores the approach face and normal, object-relative parameters, optional coarse path, validation, and diversity metadata.

\textbf{Planning and validation.}
Targets must lie in reachable free space with sufficient clearance and path connectivity. Invalid candidates are rejected or, when close, projected to a nearby navigable point. A* or another global planner provides a coarse path; DWB-style local refinement can enforce embodiment dynamics. Final poses must also support visibility and manipulation reachability.

\begin{table*}[t]
\centering
\small
\renewcommand{\arraystretch}{1.15}
\caption{Navigation target and approach-pose annotation. The navigation primitive generates interaction-ready base poses around objects or parts and validates them on the navigation mesh.}
\label{tab:supp_navigation_annotation}
\begin{tabularx}{\textwidth}{p{0.18\textwidth}p{0.28\textwidth}X}
\toprule
Component & Annotation & Description \\
\midrule
Target object or part &
$o(h)$ &
The object or part that the robot should approach, such as a cabinet, drawer, table, appliance, or object of interest. \\

Scene anchor &
$h_{\mathrm{scene}}$ &
A free-space or object-centric approach region around the target. This region defines where valid base poses may be sampled. \\

Functional affordance &
$\phi(h_{\mathrm{scene}})$ &
The scene-level affordance, such as navigation approach, interaction-ready base pose, or approach-for-manipulation. \\

Entry face center &
$f_w$ &
The center of the object face approached by the robot, computed from ray-AABB intersection in the object-local frame. \\

Outward normal &
$n_w$ &
The world-frame outward normal of the entry face. The base is placed outside this face and oriented toward $-n_w$. \\

Base target &
$q_{\mathrm{base}}=(x,y,\psi)$ &
The final navigation target, including base position and yaw. \\

Standoff margin &
$\rho$ &
The distance between the object face and the target base position. Larger margins create safer but farther approach poses. \\

Coarse path &
$\tau_{\mathrm{nav}}$ &
An optional A*-style path or waypoint sequence from the current base pose to the target base pose. \\

Validation metadata &
$v_{\mathrm{nav}}$ &
Reachability flag, nav-mesh validity, path connectivity, clearance, final pose error, and planner status. \\
\bottomrule
\end{tabularx}
\end{table*}

During execution,
\[
\mathrm{NavTo}(\mathrm{robot\_id},q_{\mathrm{base}})
\]
succeeds when
\[
\|p_{\mathrm{base}}-p_{\mathrm{target}}\|_2<\epsilon_{\mathrm{pos}},\qquad
|\Delta\psi|<\epsilon_{\mathrm{yaw}},
\]
where $p_{\mathrm{base}}$ is the achieved base position, $p_{\mathrm{target}}=(x,y)$ is the positional component of $q_{\mathrm{base}}$, and $\Delta\psi$ is the wrapped yaw error.
Metadata records completion, timeout, planner failure, reset invalidation, and object-relative face information so targets can be recomputed after object motion.

\begin{table*}[t]
\centering
\small
\renewcommand{\arraystretch}{1.15}
\caption{Common validation checks and failure cases for navigation target annotation.}
\label{tab:supp_navigation_validation_failures}
\begin{tabularx}{\textwidth}{p{0.20\textwidth}p{0.26\textwidth}X}
\toprule
Check & Failure case & Rejection or repair rule \\
\midrule
Nav-mesh validity &
Generated target lies outside the navigable mesh or inside occupied space. &
Reject the candidate or project it to the nearest valid navigable point if the projection remains close to the original target. \\

Clearance &
The base pose is too close to the object, wall, furniture, or obstacle. &
Reject poses with clearance below a safety margin. \\

Path connectivity &
The target is valid locally but disconnected from the robot's current reachable region. &
Reject if A* or the global planner cannot connect the current base pose to the target. \\

Yaw feasibility &
The target position is valid, but the desired yaw causes poor visibility or poor manipulation reachability. &
Perturb yaw or reject if the final orientation cannot face the target object or part. \\

Embodiment dynamics &
The A* path contains turns or passages infeasible for the robot base. &
Pass the path to local optimization or DWB-style rollout; reject if local rollout fails. \\

Manipulation reachability &
The robot can navigate to the base pose but cannot reach the target part from there. &
Reject if the manipulation anchor is outside the arm workspace or fails IK / reachability checks. \\

Stale target &
The object moves or the environment resets after the target is computed. &
Clear the cached target and recompute the object-centric approach pose. \\
\bottomrule
\end{tabularx}
\end{table*}

Navigation extends the schema from object contacts to interaction-ready scene poses while retaining geometric grounding, embodiment-aware validation, and candidate diversity.

\section{Additional Details of Large-scale Robot Data Collection}
\label{supp:robot-data-collection}

\begin{figure}[h]
    \centering
    \includegraphics[width=0.98\textwidth]{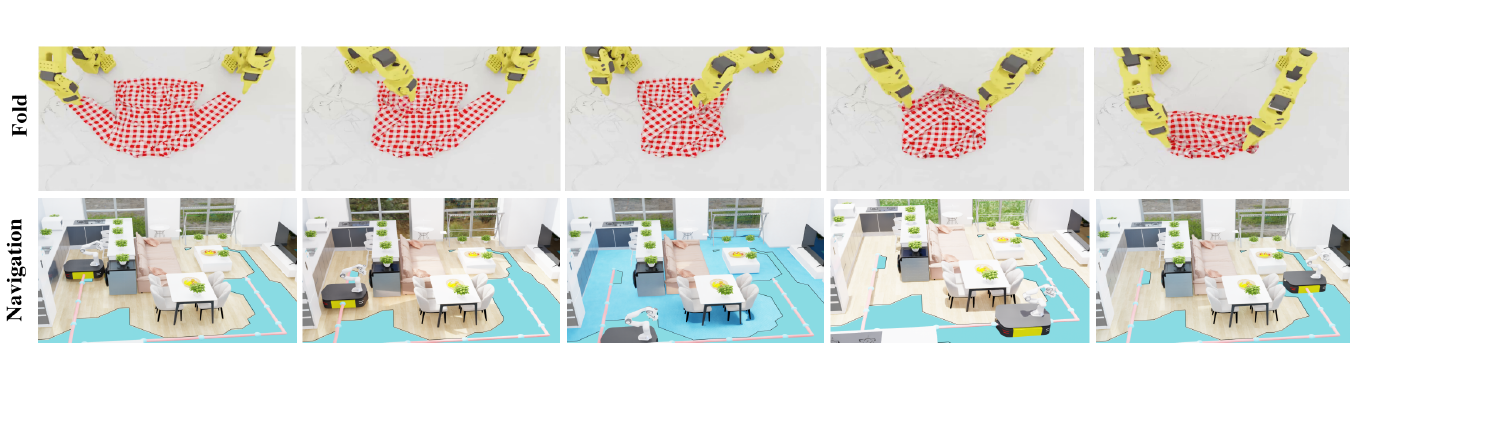}
    \caption{Closed-loop rollouts of policies trained from annotation-enabled simulation data: deformable-object folding on a real garment (top) and navigation in a furnished indoor scene (bottom).}
    \label{fig:supp_downstream_rollouts}
\end{figure}

\paragraph{Collection system.}
Our large-scale data-collection system treats AnnotateAnything annotations as executable task interfaces rather than object-specific scripted demonstrations. An annotation-aligned skill library covers tabletop and bimanual manipulation, whole-body and humanoid control, dexterous-hand manipulation, and mobile manipulation. Each skill maps compact fields---including grasp poses, target parts, waypoints, insertion directions, hanging anchors, and navigation goals---to controller or planner goals. Long-horizon tasks are formed by composing these skills according to task templates and object-state transitions.

Rollouts are generated in asynchronous parallel simulation environments. Each worker independently samples assets, tasks, robot embodiments, object poses, obstacles, and scene layouts, with domain randomization over object pose, robot--object configuration, camera viewpoint, lighting, materials, textures, distractor objects, and scene layout. We use cuRobo-v2 for goal-set IK, motion planning, and obstacle-aware execution; it provides GPU-native collision-aware motion generation for standard manipulators and high-DoF embodiments~\cite{curobo_v2}. Because randomization changes the relative robot--object pose, a worker retrieves ranked candidates from the relevant annotation bank. For each candidate, it solves goal-set IK using the complete task-compatible robot embodiment and selects the feasible IK solution with the lowest planning cost. If the highest-ranked candidate is infeasible under full-robot IK or planning constraints, the worker evaluates subsequent stored candidates until a feasible solution is found or the candidate set is exhausted. Rollouts are checked for task success, collision safety, contact consistency, articulation progress, and final-state stability when applicable. Candidates that repeatedly fail under randomized conditions are removed from the bank. This system demonstrates annotation executability and scalable data generation; a complete system-level analysis is deferred to companion work.

\subsection{Keypoint Generation and Affordance Grounding}
\label{supp:keypoint-affordance}

\begin{figure}[h]
    \centering
    \includegraphics[width=0.98\textwidth]{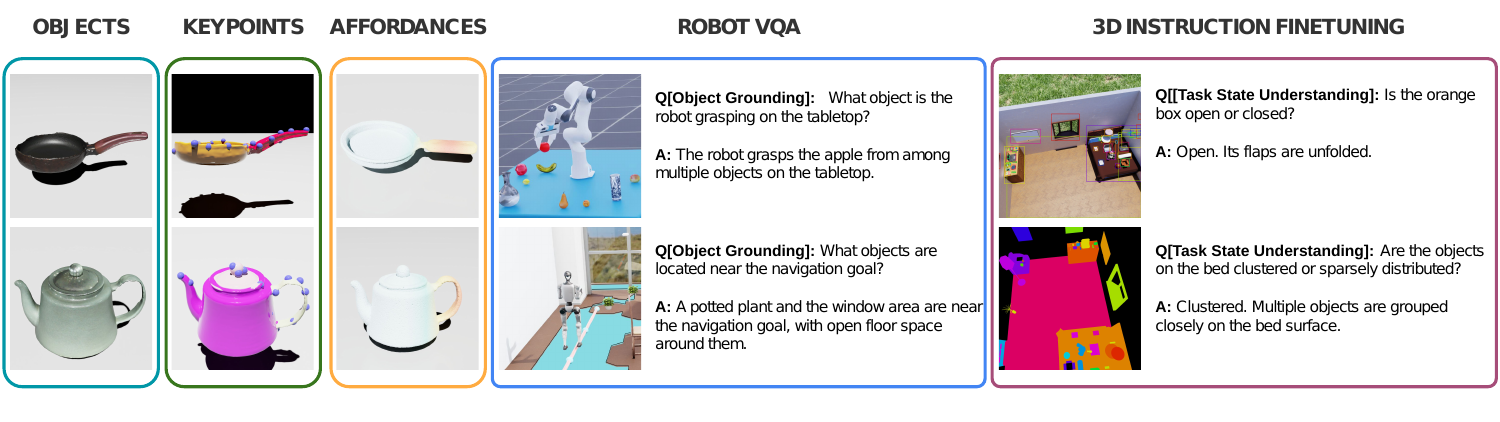}
    \caption{Downstream uses of the annotations. Object, keypoint, and affordance labels support robot-centric VQA for object grounding and navigation reasoning, and 3D instruction tuning for articulation-state and object-distribution reasoning.}
    \label{fig:supp_downstream_tasks}
\end{figure}

\paragraph{Keypoints and executable affordances.}
AnnotateAnything provides two related but distinct perception labels. Keypoints are sparse semantic or functional anchors produced by the visual annotation stage, whereas affordances describe where and how an interaction can be executed. As described in Sec.~\ref{sec:visual_annotation}, keypoint generation uses fused 3D observations reconstructed from rendered RGB-D views, samples spatially diverse candidate points (e.g., by farthest-point sampling), and uses a VLM to select meaningful locations such as handles, openings, corners, garment landmarks, or scene-level targets. A keypoint identifies a useful region but need not itself define an executable action.

Affordance supervision is derived from the validated candidate banks produced by the physics-based action pipeline in Sec.~\ref{sec:action_generation_pipeline}. Each positive label can include an anchor, skill, action direction or frame, contact configuration, trajectory, and validation metadata. Thus, it represents an interaction verified under the corresponding simulation, embodiment, and validation settings rather than only a visually plausible region. Multiple successful candidates are retained to preserve a multimodal label distribution.

\paragraph{Representations across asset types.}
For rigid objects, validated parallel-jaw or dexterous grasps provide grasp centers, approach directions, orientations, widths, palm poses, finger contacts, and contact modes. For articulated objects, labels pair a contact anchor with a task-conditioned motion trajectory, distinguishing actions such as pulling, pushing, rotating, and holding. Garment and deformable-object affordances are relational: pick points are stored with paired place points, second-hand grasp points, fold lines, or manipulation trajectories. Room-scale labels use occupancy or BEV maps to represent navigable regions, object-centric approach poses, interaction-ready base poses, and feasible navigation targets, connecting navigation to local manipulation.

\paragraph{Perception supervision.}
For image-based models, 3D anchors, contact regions, and trajectories can be projected into rendered RGB-D views to form keypoint targets, affordance masks, heatmaps, or directional fields. For point-cloud models, the labels can be attached to 3D points or local patches; for navigation models, room-level labels can be rasterized into BEV supervision for traversability, target poses, and interaction readiness. Successful candidates provide positive labels. Failed candidates can serve as hard negatives when failure reflects physical infeasibility, such as collision, unstable contact, unreachable geometry, or invalid motion, but we use them conservatively because some failures depend on the selected embodiment, planner, or randomized scene rather than the object's intrinsic affordance. This construction provides reusable keypoint and affordance supervision. We quantitatively evaluate the generated annotations within established part, affordance, and keypoint pipelines below.

\subsection{Utility as Supervision for Established Pipelines}
\label{supp:downstream-utility}

We evaluate whether generated annotations can replace reference supervision or inputs in established downstream pipelines: SAMPart3D~\cite{yang2024sampart3d} on a matched PartNeXt subset~\cite{wang2025partnext}, LASO~\cite{liu2024laso} on its matched evaluation subset, and Skeleton Merger~\cite{shi2021skeleton} on the Airplane and Chair categories of KeypointNet~\cite{you2020keypointnet}. For each experiment, the downstream pipeline and evaluation protocol are kept fixed; only the original supervision or input is replaced by AnnotateAnything outputs. This evaluation is distinct from the direct annotation-accuracy evaluation in Sec.~\ref{sec:exp_vl_quality}: here, the reported metric is the downstream pipeline's final performance.

\begin{table}[t]
\centering
\caption{
Using AnnotateAnything annotations as supervision or input for established downstream pipelines. ``Reference'' denotes our matched run using the pipeline's original supervision under the same subset and evaluation metric.
}
\label{tab:supp_downstream_utility}
\scriptsize
\setlength{\tabcolsep}{3.2pt}
\renewcommand{\arraystretch}{1.08}
\resizebox{\columnwidth}{!}{%
\begin{tabular}{llllcc}
\toprule
Annotation & Pipeline & Setting & Metric & Reference & Ours \\
\midrule
Part
& SAMPart3D
& 247 PartNeXt objects
& mIoU
& 46.70
& 43.9 \\

Affordance
& LASO
& 200 LASO objects
& mIoU/F1
& 43.6/57.2
& 40.7/54.1 \\

Keypoint
& Skeleton Merger
& KeypointNet Airplane/Chair
& DAS
& 77.7/76.8
& 74.5/73.6 \\
\bottomrule
\end{tabular}%
}
\end{table}

Across part segmentation, affordance grounding, and keypoint transfer, AnnotateAnything annotations yield downstream performance with modest gaps to the corresponding reference supervision. These experiments demonstrate downstream usability with modest gaps to reference supervision, rather than parity.

\subsection{Robot Reasoning VQA}
\label{supp:robot-vqa}

As an illustrative repurposing, simulation rollouts can be converted into robot-reasoning VQA using execution-level metadata, including the selected object, target part, anchor point, skill type, goal-set IK solution, approach pose, navigation target, articulation direction, garment manipulation landmark, and task-success state. These labels are determined at runtime because domain randomization can change which candidate is feasible even for the same asset.

We convert these records into VQA pairs using VLM-assisted language generation. The answers are derived from simulator records and selected candidates; the VLM only diversifies and naturalizes question wording. Questions cover object and part selection, approach direction, navigation progress, articulation motion, garment landmarks, and execution success. Because they concern executed candidates and task progress, these questions require trajectory-level traces rather than static asset annotations alone. This lightweight task demonstrates trajectory-grounded reasoning supervision rather than a full VQA benchmark.

\subsection{3D VLM Instruction-tuning Data}
\label{supp:3d-vlm-finetuning}

As another illustrative repurposing, AnnotateAnything supports instruction-tuning data from two complementary sources. Simulation provides 3D and projected 2D bounding boxes, instance masks, object poses, part identities, and object--instance associations, which yield grounding instructions for localization, instance identification, part reference, and 2D--3D association. Language annotations and the cross-level composition in Sec.~\ref{sec:vl_pipeline} provide spatial-reasoning supervision. Because object-level annotations are linked to room layouts through instance identities and 6D poses, they support questions about object relations, room structure and layout, containment, support, proximity, and object-to-room associations.

Structured templates ground answers in simulator or annotation metadata, while VLM-assisted paraphrasing increases linguistic diversity without supplying ground truth. The resulting data illustrates how the same annotated assets can support 3D visual grounding and spatial reasoning. Full-scale 3D VLM training and evaluation remain outside the scope of this paper.

\section{Additional Experimental Details}
\label{app:exp_details}
This section supplements Secs.~\ref{sec:exp_setup}--\ref{sec:real_world} with evaluation scopes, metrics, disaggregated results, validation criteria, rollout protocols, baselines, human evaluation, and timing definitions. Pipeline details appear in Secs.~\ref{app:language_pipeline}, \ref{app:visual_annotation}, \ref{app:action_annotation}, and~\ref{sec:primitive_specific_annotation}; the rollout system is described in Sec.~\ref{supp:robot-data-collection}.

\subsection{Evaluation Suites and Manual References}
\label{app:audited_suite}

We use three evaluation scopes. The \textbf{processed asset pool} supports source-coverage and scale statistics. The \textbf{audited suite} supports action-quality, rollout, human-evaluation, and reported per-skill throughput metrics. The suite was sampled before any annotation outcomes were computed and was manually checked only for asset validity, simulator compatibility, and independently defined skill applicability; no success- or failure-based filtering was applied. Metrics are computed over asset--skill pairs whose applicability is defined independently of annotation outcomes. The \textbf{manually annotated subset} is a separate balanced set containing reference language, parts, affordances, keypoints, and scene cues for visual-language and type-specific visual evaluation. Real-world trials are reported separately and are excluded from audited-suite averages. Because AnnotateAnything is not trained on the processed asset pool or audited suite, a conventional train/test split is not applicable.

\begin{table}[!ht]
\centering
\caption{
Audited evaluation suite. 
Simulation-based action-quality, rollout, and action/trajectory human-evaluation metrics in the main paper are computed over valid asset--skill pairs from this suite unless otherwise specified.
}
\label{tab:app_audited_suite}
\scriptsize
\setlength{\tabcolsep}{4pt}
\begin{tabular}{lccc}
\toprule
Asset Type & \#Assets & \#Valid Asset--Skill Pairs & Evaluated Skills \\
\midrule
Rigid objects & 480 & 1{,}700 & Grasp, DexGrasp, BiGrasp, Insertion, Hanging \\
Articulated objects & 320 & 740 & Grasp, Articulation, Navigation / Approach \\
Deformable / garments & 280 & 760 & BiGrasp, BiDexGrasp, Deformable, Hanging \\
Room-scale scenes & 80 & 498 & Navigation / Approach, Mobile Manipulation \\
\midrule
Total & 1{,}160 & 3{,}698 & -- \\
\bottomrule
\end{tabular}
\end{table}

Table~\ref{tab:app_audited_suite} defines the scale of the action-quality and rollout evaluation. The visual-language study uses a separate, balanced subset so that each asset type contributes equally to the manual-reference analysis.

\begin{table}[!ht]
\centering
\caption{
Manually annotated subset used for visual-language and visual annotation evaluation. 
Reference annotations include language descriptions, functional parts, affordance regions, keypoints, and scene-level spatial cues when applicable.
}
\label{tab:app_manual_subset}
\scriptsize
\setlength{\tabcolsep}{4pt}
\begin{tabular}{lcl}
\toprule
Asset Type & \#Examples & Reference Annotation Types \\
\midrule
Rigid objects & 90 & Language, parts, keypoints, affordances \\
Articulated objects & 90 & Language, parts, keypoints, affordances, articulation cues \\
Deformable / garments & 90 & Language, garment keypoints, parts, affordances \\
Room-scale scenes & 90 & Scene language, object layout, spatial cues, approach regions \\
\midrule
Total & 360 & -- \\
\bottomrule
\end{tabular}
\end{table}

The manually annotated subset contains $90$ examples from each of the four asset families and is used only for visual-language and visual-annotation evaluation. It is not used to select, filter, or tune the audited action-quality suite.

\subsection{Annotation Coverage and Conversion Statistics}
\label{app:coverage_details}

The statistics underlying Fig.~\ref{fig:annotation_coverage} characterize annotation conversion rather than a curated asset release. The processed pool contains $17{,}005$ assets from $9$ source families. The reported $100$M total includes original candidates and validated perturbation- and symmetry-derived variants. Failed raw proposals are excluded, and every augmented variant is counted only after passing the corresponding geometry and physics validation checks. The total therefore measures validated annotation instances rather than $100$M distinct manipulation strategies.

\begin{table}[htbp]
\centering
\caption{
Processed asset pool used for annotation generation. 
Source families are grouped coarsely because the goal is annotation conversion rather than curated asset-dataset release.
}
\label{tab:app_asset_pool}
\scriptsize
\setlength{\tabcolsep}{4pt}
\begin{tabular}{lccc}
\toprule
Asset Type & \#Assets & \#Ann./Obj./Skill & Main Applicable Skills \\
\midrule
Rigid objects & 12{,}538 & 869 & Grasp, DexGrasp, BiGrasp, Insertion, Hanging \\
Articulated objects & 2{,}094 & 304 & Articulation, Grasp, Navigation / Approach \\
Deformable / garments & 2{,}167 & 83 & BiGrasp, BiDexGrasp, Deformable, Hanging \\
Room-scale scenes & 206 & 527 & Navigation / Approach, Mobile Manipulation \\
\midrule
Total & 17{,}005 & -- & -- \\
\bottomrule
\end{tabular}
\end{table}

The $18$ atomic skills in Fig.~\ref{fig:atomic_skill_taxonomy} are grouped into the reporting families below.

\begin{table}[htbp]
\centering
\caption{
Mapping from the atomic skills in Fig.~\ref{fig:atomic_skill_taxonomy} to the skill families reported in the experimental tables.
}
\label{tab:app_skill_taxonomy_mapping}
\scriptsize
\setlength{\tabcolsep}{4pt}
\begin{tabular}{ll}
\toprule
Reported Skill Family & Atomic Skills Included \\
\midrule
Grasp & parallel-jaw grasp, functional grasp, lift grasp \\
DexGrasp & dexterous grasp, dexterous stable hold \\
BiGrasp & bimanual grasp, dual-end-effector lift, coordinated hold \\
BiDexGrasp & dual-hand dexterous grasp, bimanual dexterous hold \\
Articulation & open, close, push / pull articulated part, rotate joint \\
Insertion & peg-in-hole, place-in-container, insert-through-opening \\
Hanging & hang-on-hook, hang-on-edge, support-and-release \\
Deformable & garment pick, fold, spread, fling, deformable hanging \\
Navigation / Approach & navigation target, object-centric approach, interaction-ready base pose \\
\bottomrule
\end{tabular}
\end{table}

Using these reporting families, Table~\ref{tab:app_conversion_funnel}
summarizes how raw candidates are filtered through geometry, trajectory,
and physics checks before entering the retained annotation bank.

On the audited suite, each attempted pair produces on average $2{,}315$ raw candidates; $1{,}384$ pass geometry checks, $842$ pass IK or trajectory checks, $615$ pass physics validation, and $538$ enter the final bank. The corresponding ratios are reported below.

\begin{table}[!ht]
\centering
\caption{
Stage-wise candidate-to-annotation conversion on the audited suite. 
Counts are averaged per attempted asset--skill pair.
}
\label{tab:app_conversion_funnel}
\scriptsize
\setlength{\tabcolsep}{4pt}
\begin{tabular}{lccccc}
\toprule
Statistic & Generated & Geometry Feasible & IK / Traj. Feasible & Physics Validated & Retained Bank \\
\midrule
Avg. count / attempted pair & 2{,}315 & 1{,}384 & 842 & 615 & 538 \\
Retained ratio & 100.0\% & 59.8\% & 36.4\% & 26.6\% & 23.2\% \\
\bottomrule
\end{tabular}
\end{table}

Table~\ref{tab:app_skill_scale} reports pair counts, physics pass rate (Pass), asset--skill readiness (Ready), retained annotations per ready pair (Accept\#/Ready), and retained annotations per wall-clock minute (Ann./min). Macro averages weight skill families equally.

\begin{table}[!ht]
\centering
\caption{
Atomic-skill annotation statistics on the audited suite.
}
\label{tab:app_skill_scale}
\scriptsize
\setlength{\tabcolsep}{4pt}
\begin{tabular}{lccccc}
\toprule
Skill Family & \#Asset--Skill Pairs & Pass & Ready & Accept\#/Ready & Ann./min \\
\midrule
Grasp & 760 & 60.3 & 96.4 & 1{,}220 & 800 \\
DexGrasp & 520 & 43.8 & 88.7 & 460 & 200 \\
BiGrasp & 430 & 38.5 & 90.5 & 390 & 105 \\
BiDexGrasp & 320 & 27.4 & 82.4 & 290 & 95 \\
Articulation & 390 & 41.6 & 89.6 & 380 & 110 \\
Insertion & 290 & 33.2 & 84.1 & 240 & 70 \\
Hanging & 310 & 31.5 & 81.7 & 230 & 60 \\
Deformable & 260 & 30.1 & 78.9 & 120 & 45 \\
Navigation / Approach & 418 & 58.7 & 94.2 & 950 & 300 \\
\midrule
Macro Avg. & -- & 40.6 & 87.4 & 475.6 & 198 \\
\bottomrule
\end{tabular}
\end{table}

\paragraph{Candidate-bank diversity.}
Raw candidate counts include dense local variations and therefore do not directly measure distinct manipulation modes. We cluster candidates within each asset--skill pair using a $2$\,cm translational and $10^\circ$ rotational threshold for wrist poses. DexGrasp additionally uses a $0.15$\,rad finger-joint threshold. Articulation clustering additionally distinguishes the interacted part, motion direction, trajectory family, and waypoint geometry. The evaluated asset sets were sampled independently from the processed asset pool and are separate from the audited action-quality suite; their selection did not use annotation outcomes.

\begin{table}[!ht]
\centering
\caption{
Candidate-bank diversity after collapsing local variants. Values report median [IQR] over independently sampled assets.
}
\label{tab:app_candidate_diversity}
\scriptsize
\setlength{\tabcolsep}{3.5pt}
\renewcommand{\arraystretch}{1.08}
\begin{tabular}{lcccc}
\toprule
Skill & Assets & Candidates / asset & Clusters / asset & Cluster ratio \\
\midrule
Grasp
& 4{,}185
& 851 [560, 1{,}113]
& 176 [118, 239]
& 0.230 [0.142, 0.352] \\
DexGrasp
& 400
& 210 [158, 257]
& 153 [96, 193]
& 0.714 [0.501, 0.862] \\
Articulation
& 300
& 162 [63, 528]
& 21 [10, 42]
& 0.119 [0.084, 0.171] \\
\bottomrule
\end{tabular}
\end{table}

The median inter-cluster wrist-pose separations are $20.9$\,cm/$125.6^\circ$ for Grasp, $21.4$\,cm/$132.9^\circ$ for DexGrasp, and $13.8$\,cm/$78.6^\circ$ for Articulation. The median normalized inter-cluster trajectory distance for Articulation is $0.276$. These statistics indicate that the retained banks contain materially different pose and trajectory modes in addition to local perturbations.

\subsection{Visual-Language and Visual Annotation Evaluation}
\label{app:vl_eval_details}

Evaluation uses the balanced manual subset in Table~\ref{tab:app_manual_subset}. Complete bundles contain language, keypoints, part or affordance regions, and scene cues when applicable. We compare \textbf{Direct VLM}, which predicts the bundle directly from multi-view images; \textbf{Aff.-only + heur.}, which uses visual affordance anchors and geometric heuristics to construct regions and anchors; \textbf{w/o 3D refinement}, which removes point-cloud consistency and part-region refinement; and the complete method. All variants are converted to the same bundle and type-specific output schemas before scoring. Five raters score semantic correctness, 3D grounding, coverage, and actionability from $0$ to $100$. The overall bundle score averages these four dimensions, and reported uncertainty is the standard error over example-level scores after averaging across raters.

\begin{table}[!ht]
\centering
\caption{
Human evaluation rubric for complete visual-language annotation bundles. 
All dimensions are scored from $0$ to $100$.
}
\label{tab:app_vl_rubric}
\scriptsize
\begin{tabular}{lp{0.68\linewidth}}
\toprule
Dimension & Description \\
\midrule
Semantic correctness & Correctness of object, part, function, affordance, and interaction-constraint descriptions. \\
3D grounding accuracy & Whether keypoints, part regions, and affordance anchors lie on the intended 3D regions. \\
Coverage & Whether major interaction-relevant parts and regions are included. \\
Actionability & Whether the annotation provides useful anchors for physics-grounded action generation. \\
\bottomrule
\end{tabular}
\end{table}

For type-specific evaluation, let $\mathcal{R}$ and $\mathcal{P}$ be reference and predicted regions and let $\ell(\cdot)$ denote a compatible part or affordance label. We use bidirectional label-compatible IoU,
\[
S_{\mathrm{reg}}(\mathcal{P},\mathcal{R})
=
50\left(
\frac{1}{|\mathcal{R}|}\sum_{r\in\mathcal{R}}\max_{p\in\mathcal{P}:\ell(p)\sim\ell(r)}\mathrm{IoU}(p,r)
+
\frac{1}{|\mathcal{P}|}\sum_{p\in\mathcal{P}}\max_{r\in\mathcal{R}:\ell(p)\sim\ell(r)}\mathrm{IoU}(p,r)
\right),
\]
and normalized keypoint correctness,
\[
S_{\mathrm{kp}}
=
100\frac{1}{|\mathcal{K}^{\star}|}
\sum_{k^{\star}\in\mathcal{K}^{\star}}
\mathbf{1}\!\left[
\min_{k:\ell(k)\sim\ell(k^{\star})}
\frac{\|k-k^{\star}\|_2}{d_{\mathcal{A}}}<0.05
\right].
\]
Here $d_{\mathcal{A}}$ is the object bounding-box diagonal, or the local room scale for scene anchors. The region metric is applied separately with functional-part and affordance label spaces. These aggregate matching scores, unlike bundle ratings, are not subjective human scores.

\begin{table}[!ht]
\centering
\caption{
Type-specific visual annotation metrics used for the right block of Table~\ref{tab:vl_quality}. 
All scores are normalized to a $0$--$100$ scale.
}
\label{tab:app_visual_annotation_metrics}
\scriptsize
\setlength{\tabcolsep}{4pt}
\begin{tabular}{lp{0.68\linewidth}}
\toprule
Annotation Type & Metric Definition \\
\midrule
Part & Label-compatible bidirectional IoU between predicted and reference functional part regions. \\
Affordance & Label-compatible bidirectional IoU between predicted and reference affordance regions. \\
Keypoint & Normalized keypoint correctness under a label-compatible distance threshold. \\
\bottomrule
\end{tabular}
\end{table}

Using these definitions, we next report complete bundle quality by asset type before separating the automatic visual-annotation scores.

\begin{table}[htbp]
\centering
\caption{
Bundle-level visual-language annotation quality by asset type. 
Scores are human ratings on a $0$--$100$ scale.
}
\label{tab:app_vl_breakdown}
\scriptsize
\setlength{\tabcolsep}{3.5pt}
\begin{tabular}{llccccc}
\toprule
Asset Type & Method & Semantic & 3D Ground. & Coverage & Action. & Overall \\
\midrule
Rigid & Direct VLM & 85.4 & 63.8 & 71.5 & 66.0 & 71.7 \\
Rigid & Aff.-only + heur. & 78.0 & 71.8 & 74.2 & 68.4 & 73.1 \\
Rigid & w/o 3D refine & 88.2 & 75.6 & 79.4 & 77.3 & 80.1 \\
Rigid & Ours & 93.5 & 90.8 & 91.0 & 91.8 & 91.8 \\
\midrule
Articulated & Direct VLM & 83.5 & 60.2 & 67.1 & 62.8 & 68.4 \\
Articulated & Aff.-only + heur. & 76.5 & 69.0 & 72.8 & 66.2 & 71.1 \\
Articulated & w/o 3D refine & 87.0 & 72.1 & 76.5 & 73.2 & 77.2 \\
Articulated & Ours & 92.8 & 88.7 & 89.6 & 90.4 & 90.4 \\
\midrule
Deformable / garments & Direct VLM & 82.1 & 58.6 & 64.8 & 60.3 & 66.5 \\
Deformable / garments & Aff.-only + heur. & 75.0 & 68.5 & 70.6 & 65.0 & 69.8 \\
Deformable / garments & w/o 3D refine & 86.2 & 70.4 & 74.5 & 71.8 & 75.7 \\
Deformable / garments & Ours & 92.4 & 86.5 & 88.0 & 89.2 & 89.0 \\
\midrule
Room-scale scenes & Direct VLM & 85.8 & 63.4 & 71.4 & 67.3 & 72.0 \\
Room-scale scenes & Aff.-only + heur. & 77.8 & 71.5 & 75.0 & 70.5 & 73.7 \\
Room-scale scenes & w/o 3D refine & 88.4 & 75.5 & 80.8 & 78.5 & 80.8 \\
Room-scale scenes & Ours & 93.6 & 91.6 & 92.1 & 92.6 & 92.5 \\
\bottomrule
\end{tabular}
\end{table}

The asset-type breakdown tests whether complete annotation bundles remain reliable across rigid, articulated, deformable, and room-scale data. We separately report the automatic part, affordance, and keypoint scores to distinguish overall bundle quality from type-specific geometric matching.

\begin{table}[htbp]
\centering
\caption{
Type-specific visual annotation scores by method. 
These scores correspond to the right block of Table~\ref{tab:vl_quality}.
}
\label{tab:app_visual_annotation_scores}
\scriptsize
\setlength{\tabcolsep}{4pt}
\begin{tabular}{lccc}
\toprule
Method & Part & Affordance & Keypoint \\
\midrule
Direct VLM & 72.8 & 68.5 & 70.2 \\
Aff.-only + heur. & 70.5 & 79.3 & 68.8 \\
w/o 3D refine & 82.6 & 78.1 & 80.8 \\
Ours & \textbf{91.5} & \textbf{89.0} & \textbf{90.3} \\
\bottomrule
\end{tabular}
\end{table}

\paragraph{Grounding-interface reliability.}
We additionally evaluate each generated interface directly against manual references. Of $1{,}080$ evaluated part regions, $944$ ($87.4\%$) satisfy label compatibility and IoU $\geq 0.5$. Of $810$ affordance regions, $687$ ($84.8\%$) satisfy the corresponding criterion. Generated keypoints obtain $90.3\%$ PCK@0.05, where the threshold is normalized by the object bounding-box diagonal. For compatible-skill prediction, $300/360$ anchors ($83.3\%$) have exactly correct skill sets, with macro precision/recall/F1 of $88.4/84.6/86.5$.

\paragraph{Matched external annotation evaluation.}
On a matched $247$-object PartNeXt subset~\cite{wang2025partnext}, our part annotations obtain $44.1$ object-macro mIoU, compared with $46.70$ for SAMPart3D~\cite{yang2024sampart3d} and $41.01$ for P3-SAM~\cite{ma2025p3sam}. On a matched $200$-object LASO subset~\cite{liu2024laso}, our affordance annotations obtain $41.0/54.3$ mIoU/F1, compared with $43.6/57.2$ for the released LASO baseline and $35.8/48.9$ for Direct VLM. All methods are evaluated on identical subsets and metrics in these matched re-runs; these values should not be interpreted as direct comparisons with numbers reported under different original-paper protocols.

\paragraph{Detectable failures and semantic filtering.}
Across $3{,}240$ asset- and part-level VLM queries, $38$ fail structured parsing and $56$ cannot be linked to a valid segmentation identifier. All $94/3{,}240$ detectable failures ($2.9\%$) are rejected before action generation. Uncertain attributes may be represented as \texttt{unknown}, and anchors with unresolved interaction-defining attributes are not used for skill instantiation. These checks reject malformed or ungrounded outputs but do not identify every possible semantic error. In a separate $18$-object audit, skill-specific geometric and physics validation increases candidate-level intended-part selectivity from $50.0\%$ to $70.0\%$, with improvements on $16/18$ objects.

\subsection{Physics-grounded Action Annotation Evaluation}
\label{app:physics_eval_details}

For valid pair $(\mathcal{A},s)$, let $\mathcal{C}_{\mathcal{A},s}$ be generated candidates, $\mathcal{P}_{\mathcal{A},s}$ physics-valid candidates, and $\mathcal{V}_{\mathcal{A},s}$ retained annotations. The physics pass rate, readiness, and annotation density are
\[
R_{\mathrm{pass}}(s)=
\frac{\sum_{\mathcal{A}}|\mathcal{P}_{\mathcal{A},s}|}
{\sum_{\mathcal{A}}|\mathcal{C}_{\mathcal{A},s}|},
\qquad
R_{\mathrm{ready}}(s)=
\frac{\sum_{\mathcal{A}}\mathbf{1}[|\mathcal{V}_{\mathcal{A},s}|>0]}
{\#\text{ attempted asset--skill pairs for }s},
\]
\[
D_{\mathrm{accept}}(s)=
\frac{\sum_{\mathcal{A}}|\mathcal{V}_{\mathcal{A},s}|}
{\sum_{\mathcal{A}}\mathbf{1}[|\mathcal{V}_{\mathcal{A},s}|>0]}.
\]
Execution success is measured by retrieving ranked stored annotations, solving goal-set inverse kinematics and collision-aware planning with the complete task-compatible robot embodiment, and applying the skill-specific success criterion in Sec.~\ref{app:skill_validation}. Ann./min includes candidate generation, filtering, optimization, validation, and serialization. Macro averages weight reported skill families equally.

For object-centric skills, physics validation uses a floating end effector as a candidate filter. The reported \emph{Exec.} metric is distinct from floating validation and evaluates goal-set inverse kinematics, collision-aware motion planning, and execution with the complete task-compatible robot embodiment. Stored candidates are evaluated in rank order, with fallback to subsequent candidates when the highest-ranked candidate is infeasible under full-robot constraints.

\begin{table}[htbp]
\centering
\caption{
Full physics-grounded action annotation results on the audited suite. 
This table expands the compact metrics reported in Table~\ref{tab:physics_quality_and_collection}.
}
\label{tab:app_physics_full}
\scriptsize
\setlength{\tabcolsep}{3.5pt}
\begin{tabular}{lccccccc}
\toprule
Skill Family
& \#Pairs 
& Pass 
& Ready 
& Accept\#/Ready 
& Exec. 
& Ann./min 
& Human \\
\midrule
Grasp & 760 & 60.3 & 96.4 & 1{,}220 & 98.1 & 800 & 91.2 \\
DexGrasp & 520 & 43.8 & 88.7 & 460 & 93.9 & 200 & 85.7 \\
BiGrasp & 430 & 38.5 & 90.5 & 390 & 95.0 & 105 & 87.0 \\
BiDexGrasp & 320 & 27.4 & 82.4 & 290 & 88.5 & 95 & 82.4 \\
Articulation & 390 & 41.6 & 89.6 & 380 & 93.2 & 110 & 87.2 \\
Insertion & 290 & 33.2 & 84.1 & 240 & 89.3 & 70 & 84.1 \\
Hanging & 310 & 31.5 & 81.7 & 230 & 88.7 & 60 & 83.0 \\
Deformable & 260 & 30.1 & 78.9 & 120 & 84.1 & 45 & 80.2 \\
Navigation / Approach & 418 & 58.7 & 94.2 & 950 & 96.5 & 300 & 89.6 \\
\midrule
Macro Avg. & -- & 40.6 & 87.4 & 475.6 & 91.9 & 198 & 85.6 \\
\bottomrule
\end{tabular}
\end{table}
\vspace{-3mm}

Floating FrankaUMI candidates are rechecked on the full Franka--FrankaUMI system; floating Dex3-1 candidates are rechecked on G1--Dex3-1; and navigation or approach candidates are evaluated with the base-aware LocoManip embodiment. Deformable skills already use full-manipulator validation.

\paragraph{Floating-to-full-robot audit.}
For DexGrasp, we evaluate $520$ audited asset--skill pairs with $10$ candidates per pair. Of $5{,}200$ candidates, $2{,}278$ ($43.8\%$) pass floating-hand validation, leaving $461/520$ ready pairs ($88.7\%$). On the complete G1--Dex3-1 embodiment, the first-ranked floating-valid annotation succeeds for $402/461$ pairs ($87.2\%$). Allowing fallback to another stored candidate increases success to $433/461$ pairs ($93.9\%$). The latter is the annotation-bank execution protocol and corresponds to the DexGrasp Exec. value reported in the main paper.

\subsection{Skill-specific Validation Criteria}
\label{app:skill_validation}

Table~\ref{tab:app_skill_validation} summarizes the criteria used for the compact results in Table~\ref{tab:physics_quality_and_collection}. Primitive-specific thresholds are defined in Secs.~\ref{app:action_annotation} and~\ref{sec:primitive_specific_annotation} and are not repeated here. 

\begin{table}[!ht]
\centering
\caption{
Skill-specific validation and execution-success criteria used for evaluation.
}
\label{tab:app_skill_validation}
\scriptsize
\setlength{\tabcolsep}{3pt}
\begin{tabular}{p{0.14\linewidth}p{0.22\linewidth}p{0.29\linewidth}p{0.29\linewidth}}
\toprule
Skill Family & Candidate Representation & Physics Validation & Execution Success \\
\midrule
Grasp 
& 6D gripper pose, width, approach direction 
& Collision-free approach and closing; stable contact; bounded penetration; robustness under disturbance 
& Object remains supported through the primitive-specific lift-and-stability test \\

DexGrasp 
& Hand pose, joint configuration, contact set 
& Joint limits, self-collision, hand-object penetration, contact stability, and object support 
& Object remains stably constrained by the hand after closing and perturbation \\

BiGrasp 
& Two end-effector poses and coordinated contact targets 
& Both contacts feasible; no inter-arm collision; object remains supported by the pair 
& Object is controlled by both end-effectors without dropping or excessive drift \\

BiDexGrasp 
& Dual-hand poses, finger configurations, contact sets 
& Dual-hand joint limits, contact stability, hand-object and inter-hand collision checks 
& Dual-hand contact remains stable and task-relevant during execution \\

Articulation 
& Handle/contact point, motion direction, waypoint trajectory 
& Contact maintained along trajectory; joint progresses in intended direction; no jamming or invalid collision 
& Target joint reaches the primitive-specific progress threshold defined in Sec.~\ref{sec:primitive_specific_annotation} \\

Insertion 
& Object pose, insertion direction, approach waypoint 
& Semantic and geometric compatibility; alignment; collision-free approach; no jamming before insertion 
& Object reaches the primitive-specific insertion depth and alignment tolerance \\

Hanging 
& Hanging pose, support/contact point, release trajectory 
& Support clearance; contact formation; gravity stability; no immediate slip after release 
& Object remains supported after release under the primitive-specific stability test \\

Deformable 
& Garment keypoints, bimanual grasp points, target waypoints 
& Keypoint contacts feasible; full-arm reachability; no severe collision, overstretching, or entanglement 
& Target keypoints or shape state satisfy the task-specific deformable-object criterion \\

Navigation / Approach 
& Target base pose, approach direction, path waypoints 
& Traversable region; collision-free path; clearance; manipulation reachability at final pose 
& Robot reaches an interaction-ready target pose within the primitive-specific pose tolerance \\
\bottomrule
\end{tabular}
\end{table}

For object-centric skills, the Physics Validation column describes floating end-effector candidate filtering, while the Execution Success column describes evaluation with the complete task-compatible robot embodiment after inverse kinematics and collision-aware planning. Deformable and other strongly embodiment-dependent skills already use the full manipulator during physics validation.

\subsection{Annotation-enabled Rollout Evaluation}
\label{app:rollout_details}

The rollout system in Sec.~\ref{supp:robot-data-collection} consumes annotations as executable task interfaces. \textbf{No annotation / random} samples targets or parameters without stored labels; \textbf{VL-only annotation} uses visual-language anchors without physics-validated actions; and \textbf{Ours annotation bank} uses the complete bank. Data Succ. is successful rollouts divided by attempts; Traj./h is successful trajectories per wall-clock hour; Att./Succ. is attempts per successful trajectory; and Traj. Human is the $0$--$100$ human trajectory score. In the aggregate rows, Data Succ. and Traj./h are macro-averaged across skill families, Att./Succ. is computed as $100/\text{Data Succ.}$ because Data Succ. is expressed as a percentage, and Traj. Human is averaged over the sampled trajectory-level ratings for each method.
For each randomized scene, stored candidates are evaluated in annotation-bank rank order. For each candidate, the feasible inverse-kinematics solution with the lowest planning cost is selected; if no full-robot solution is feasible, the system proceeds to the next stored candidate.
\begin{table}[!ht]
\centering
\caption{
Per-skill annotation-enabled rollout collection results.
}
\label{tab:app_rollout_full}
\scriptsize
\setlength{\tabcolsep}{3.5pt}
\begin{tabular}{llcccc}
\toprule
Skill Family & Method & Data Succ. & Traj./h & Att./Succ. & Traj. Human \\
\midrule
Grasp & No annotation / random & 29.0 & 70 & 3.45 & 55.0 \\
Grasp & VL-only annotation & 67.5 & 150 & 1.48 & 72.6 \\
Grasp & Ours annotation bank & 90.5 & 300 & 1.11 & 90.1 \\
\midrule
DexGrasp & No annotation / random & 21.5 & 48 & 4.65 & 49.0 \\
DexGrasp & VL-only annotation & 59.4 & 110 & 1.68 & 68.0 \\
DexGrasp & Ours annotation bank & 86.7 & 170 & 1.15 & 86.0 \\
\midrule
BiGrasp & No annotation / random & 24.8 & 55 & 4.03 & 51.2 \\
BiGrasp & VL-only annotation & 61.2 & 105 & 1.63 & 69.1 \\
BiGrasp & Ours annotation bank & 87.8 & 145 & 1.14 & 86.5 \\
\midrule
BiDexGrasp & No annotation / random & 18.6 & 38 & 5.38 & 46.8 \\
BiDexGrasp & VL-only annotation & 55.0 & 88 & 1.82 & 65.0 \\
BiDexGrasp & Ours annotation bank & 82.4 & 115 & 1.21 & 82.0 \\
\midrule
Articulation & No annotation / random & 27.4 & 62 & 3.65 & 53.0 \\
Articulation & VL-only annotation & 64.8 & 122 & 1.54 & 71.2 \\
Articulation & Ours annotation bank & 88.6 & 150 & 1.13 & 87.5 \\
\midrule
Insertion & No annotation / random & 19.6 & 42 & 5.10 & 48.5 \\
Insertion & VL-only annotation & 58.1 & 96 & 1.72 & 67.5 \\
Insertion & Ours annotation bank & 84.2 & 125 & 1.19 & 84.0 \\
\midrule
Hanging & No annotation / random & 20.2 & 40 & 4.95 & 48.0 \\
Hanging & VL-only annotation & 57.9 & 90 & 1.73 & 67.2 \\
Hanging & Ours annotation bank & 83.4 & 110 & 1.20 & 83.4 \\
\midrule
Deformable & No annotation / random & 15.0 & 30 & 6.67 & 44.0 \\
Deformable & VL-only annotation & 52.6 & 78 & 1.90 & 64.0 \\
Deformable & Ours annotation bank & 80.3 & 90 & 1.25 & 80.4 \\
\midrule
Navigation / Approach & No annotation / random & 40.0 & 110 & 2.50 & 60.5 \\
Navigation / Approach & VL-only annotation & 79.4 & 195 & 1.26 & 79.0 \\
Navigation / Approach & Ours annotation bank & 90.1 & 280 & 1.11 & 89.4 \\
\midrule
Aggregate & No annotation / random & 24.0 & 55 & 4.17 & 51.8 \\
Aggregate & VL-only annotation & 61.8 & 115 & 1.62 & 69.4 \\
Aggregate & Ours annotation bank & 86.0 & 165 & 1.17 & 85.5 \\
\bottomrule
\end{tabular}
\end{table}

\subsection{Real-World Transfer Protocol}
\label{app:real_world}

We evaluate policies trained only from annotation-enabled simulation rollouts with domain randomization. A Franka Panda with its stock two-finger gripper performs Grasp, Insertion, Open Drawer, Close Lid, and long-horizon Pick-and-Place, while the same arm with an XHand attached through a custom mount performs DexGrasp. For each evaluated object or configuration, we fine-tune the pretrained $\pi_{0.5}$ policy~\cite{Intelligence202505AV} on $1{,}000$ automatically collected simulation demonstrations generated with the corresponding simulated embodiment. We use no real-world demonstrations, real-world fine-tuning, or test-time adaptation.

Grasp and DexGrasp use ten objects each; Insertion uses two peg configurations; Open Drawer uses two cabinets; Close Lid uses three lidded objects; and the long-horizon evaluation uses five five-object sets. The controlled evaluation contains $540$ single-task trials and $100$ long-horizon episodes. Each evaluated object, configuration, or long-horizon object set receives $20$ independently reset physical runs. Success requires the specified terminal state without manual intervention; failures include drops, jams, missed targets, incomplete articulation, or timeout.

\begin{table}[!ht]
\centering
\caption{
Controlled zero-shot real-world evaluation. All policies are trained only from annotation-enabled simulation rollouts. Confidence intervals are reported in Appendix~\ref{app:human_eval}.
}
\label{tab:app_real_world_protocol}
\scriptsize
\setlength{\tabcolsep}{3pt}
\renewcommand{\arraystretch}{1.05}
\begin{tabular}{lclc}
\toprule
Task & Trials & Success Criterion & Successes \\
\midrule
Grasp
& 200
& Object is lifted and held stably
& 186 \\
DexGrasp
& 200
& Object remains stably held by the dexterous hand
& 154 \\
Insertion
& 40
& Object reaches and remains in the target insertion region
& 28 \\
Open Drawer
& 40
& Drawer reaches the target open state
& 33 \\
Close Lid
& 60
& Lid reaches the target closed state
& 53 \\
Long-horizon Pick-and-Place
& 100
& All five objects are sequentially manipulated and placed successfully
& 61 \\
\bottomrule
\end{tabular}
\end{table}

The long-horizon task requires five objects to be manipulated and placed sequentially within each episode, and an episode is counted as successful only when the complete sequence succeeds. Because no matched physical baseline is included, these results provide evidence of zero-shot sim-to-real feasibility rather than comparative real-world superiority. Additional qualitative hardware demonstrations are excluded from the controlled totals.

\subsection{Baselines and Ablations}
\label{app:baselines}

Controlled ablations isolate semantic priors, 3D grounding, and physics validation. \textbf{Geometry-only} uses surface normals, curvature, free space, collision margins, reachability, and traversability without visual-language priors. \textbf{VL-only} uses semantic anchors without the complete physics optimization and validation pipeline. \textbf{w/o physics validation} generates and optimizes candidates but stores them without validator filtering; validation-dependent quantities are computed post hoc. \textbf{Ours full} combines visual-language priors, 3D grounding, embodiment constraints, optimization, physics validation, and physics-aware augmentation. The random variant is used only for rollout collection.

\begin{table}[!ht]
\centering
\caption{
Per-skill ablation summary. 
Each cell reports Pass / Exec.
}
\label{tab:app_per_skill_ablation}
\scriptsize
\setlength{\tabcolsep}{3pt}
\begin{tabular}{lcccc}
\toprule
Skill Family & Geometry-only & VL-only & w/o Phys. Val. & Ours \\
\midrule
Grasp & 40.5 / 69.8 & 46.0 / 75.2 & 39.8 / 70.5 & 60.3 / 98.1 \\
DexGrasp & 28.8 / 54.0 & 32.4 / 62.6 & 27.1 / 58.0 & 43.8 / 93.9 \\
BiGrasp & 30.2 / 56.4 & 34.0 / 64.5 & 28.8 / 60.1 & 38.5 / 95.0 \\
BiDexGrasp & 18.5 / 46.2 & 23.0 / 55.5 & 19.2 / 50.0 & 27.4 / 88.5 \\
Articulation & 32.5 / 58.5 & 35.7 / 67.0 & 31.0 / 62.6 & 41.6 / 93.2 \\
Insertion & 22.0 / 48.6 & 27.5 / 57.2 & 23.4 / 53.5 & 33.2 / 89.3 \\
Hanging & 20.8 / 47.5 & 25.8 / 56.0 & 22.7 / 52.2 & 31.5 / 88.7 \\
Deformable & 17.6 / 42.0 & 21.5 / 50.4 & 19.0 / 47.1 & 30.1 / 84.1 \\
Navigation / Approach & 42.0 / 71.0 & 50.5 / 82.2 & 32.0 / 74.0 & 58.7 / 96.5 \\
\bottomrule
\end{tabular}
\end{table}

Framework-adapted external comparison protocols and results have been moved to Sec.~\ref{sec:exp_physics_quality} and Table~\ref{tab:external_adapted_comparison}, adjacent to the main physics and rollout results. The appendix retains the controlled-ablation definitions and per-skill breakdowns; the moved comparison reports framework-adapted estimates rather than official or end-to-end reproduced metrics from the original systems.

\paragraph{Protocol-matched DexGrasp baseline.}
We additionally compare with BODex~\cite{chen2025bodexscalableefficientrobotic} using the same Dex3-1 hand, $520$ audited asset--skill pairs, $10$ candidates per pair, the same floating-hand validator, and the same G1 full-arm execution protocol. No output conversion or hand retargeting is applied. BODex obtains $40.1/84.8/90.2\%$ Pass/Ready/Exec., compared with $43.8/88.7/93.9\%$ for AnnotateAnything.

\subsection{Human Evaluation and Uncertainty Reporting}
\label{app:human_eval}

Human evaluation covers complete visual-language bundles, retained action annotations, and rollout trajectories. Each item is independently evaluated by five non-author raters with experience in robotics, perception, planning, or simulation. The raters did not participate in method development, output generation, or evaluation-sample selection. Before evaluation, all raters complete calibration on nine held-out examples that are excluded from the reported results.

Method identities and validation outcomes are hidden. Outputs are randomized and interleaved across methods, and raters see the input asset or scene, the generated annotations, and the corresponding action or rollout visualization when applicable. Ratings use a $0$--$100$ scale in discrete 10-point increments. We evaluate $360$ visual-language bundles, $540$ action annotations, and $450$ trajectories per rollout method. Generation-based evaluations use three independent seeds.

For each example, we first average the five raters' scores and then report the mean and standard error over example-level scores. Type-specific Part, Affordance, and Keypoint scores are automatic reference-matching metrics rather than human ratings.

\begin{table}[!ht]
\centering
\caption{
Human evaluation dimensions. 
All dimensions are scored from $0$ to $100$.
}
\label{tab:app_human_rubric}
\scriptsize
\setlength{\tabcolsep}{5pt}
\begin{tabular}{lll}
\toprule
Target & Dimensions & Overall Score \\
\midrule
Visual-language bundle 
& Semantic correctness; 3D grounding; coverage; actionability 
& Average over dimensions \\
Action annotation 
& Task relevance; physical plausibility; executability; diversity 
& Average over dimensions \\
Rollout trajectory 
& Task alignment; motion quality; safety plausibility; demonstration usefulness 
& Average over dimensions \\
\bottomrule
\end{tabular}
\end{table}

The rubric above defines the dimensions summarized by each overall human score. The disaggregated values below show the corresponding dimension-level results for visual-language bundles, physics action annotations, and collected trajectories.

\begin{table}[!ht]
\centering
\caption{
Per-dimension human evaluation scores.
}
\label{tab:app_human_breakdown}
\scriptsize
\setlength{\tabcolsep}{4pt}
\begin{tabular}{lccccc}
\toprule
Target & Dim. 1 & Dim. 2 & Dim. 3 & Dim. 4 & Overall \\
\midrule
Visual-language bundle & 93.1 & 89.4 & 90.2 & 91.0 & 90.9 \\
Physics action annotation & 86.8 & 84.9 & 85.2 & 85.5 & 85.6 \\
Collected trajectory & 86.2 & 85.0 & 84.7 & 86.1 & 85.5 \\
\bottomrule
\end{tabular}
\end{table}

\paragraph{Inter-rater agreement.}
Using interval-scale Krippendorff's $\alpha$, agreement is $0.76$ for visual-language bundles, $0.69$ for action annotations, and $0.72$ for rollout trajectories.

\paragraph{Confidence intervals.}
Paired confidence intervals use $10{,}000$ cluster-bootstrap resamples. For action and rollout rates, the asset--skill pair is the resampling cluster, so all candidates and repeated executions from the same pair remain together. For human-score comparisons, all seeded generations of the same matched item remain in the same cluster. Real-world success-rate intervals use Wilson binomial intervals computed from the exact physical-trial totals.

\begin{table}[!ht]
\centering
\caption{
Summary of paired $95\%$ confidence intervals for the main comparisons.
}
\label{tab:app_uncertainty_summary}
\scriptsize
\setlength{\tabcolsep}{4pt}
\begin{tabular}{lcc}
\toprule
Comparison & Improvement & $95\%$ CI \\
\midrule
Visual-language overall score & $+12.4$ & $[9.8,15.0]$ \\
Action-annotation human score & $+15.4$ & $[11.9,18.9]$ \\
Rollout-trajectory human score & $+16.1$ & $[12.6,19.6]$ \\
ManiTwin~\cite{Wang2026ManiTwinSD} matched Pass & $+4.5$ pp & $[3.6,5.4]$ \\
ManiTwin matched Ready & $+13.8$ pp & $[8.1,19.5]$ \\
ManiTwin matched Exec. & $+10.7$ pp & $[8.8,12.6]$ \\
\bottomrule
\end{tabular}
\end{table}

\paragraph{Real-world uncertainty.}
Wilson $95\%$ confidence intervals are $[88.6,95.8]\%$ for Grasp, $[70.7,82.3]\%$ for DexGrasp, $[54.6,81.9]\%$ for Insertion, $[68.1,91.3]\%$ for Open Drawer, $[77.8,94.2]\%$ for Close Lid, and $[51.2,70.0]\%$ for long-horizon Pick-and-Place.

\subsection{Timing Metrics}
\label{app:timing_metrics}

Annotation throughput is measured as retained annotations per wall-clock minute and includes candidate generation, geometry filtering, trajectory or pose optimization, physics validation, and serialization. The per-skill values are reported in Table~\ref{tab:app_skill_scale}. For variants without validator filtering, throughput is measured on the stored candidate bank and validation-dependent metrics are computed post hoc. Rollout throughput is successful trajectories per wall-clock hour and includes candidate retrieval, goal-set inverse kinematics, collision-aware motion planning, full-robot execution, and trajectory validation. Compute resources are listed in Appendix~\ref{app:compute_resources}.

\section{Related Work}
\label{app:related_work}

\subsection{Simulation-Based Automatic Data Collection}

Simulation-based data generation often relies on demonstration reuse or optimization in simulation. RLBench provides expert demonstrations generated through waypoint-guided motion planning, while MimicGen transforms a small set of human demonstrations to new scene configurations, object instances, and robot arms \cite{james2019rlbench,mandlekar2023mimicgen}. RoboGen uses foundation and generative models to propose tasks, construct scenes, and produce training supervision, whereas GenSim uses large language models to generate simulation task programs and expert demonstrations \cite{wang2024robogen,wang2024gensim}. Benchmark suites such as Meta-World and GPU simulators such as Isaac Gym support large-scale reinforcement-learning experimentation, and simulation-trained dexterous systems show that difficult skills can be learned without human demonstrations \cite{yu2019metaworld,makoviychuk2021isaacgym,openai2018dextrous}. Demonstration-based systems often depend on task programs or seed demonstrations; optimization-based systems require substantial reward, curriculum, and training design, and commonly produce task- or controller-specific outputs.

AnnotateAnything targets a complementary layer: reusable semantic and executable annotations that can initialize multiple data-collection and policy-learning pipelines, rather than demonstrations or policies specialized to one training setup.

\subsection{Automatic Manipulation Annotation and Affordance Generation}

Grasping has received especially extensive attention in automatic manipulation annotation. Dex-Net 2.0 and ACRONYM synthesize large grasp datasets using analytic metrics and physics simulation, respectively \cite{mahler2017dexnet2,eppner2021acronym}; 6-DOF GraspNet and Contact-GraspNet learn grasp proposals from point clouds and cluttered observations \cite{mousavian2019graspnet,sundermeyer2021contactgraspnet}. DexGraspNet extends large-scale annotation to dexterous hands, and recent work also addresses bimanual dexterous grasp synthesis \cite{wang2023dexgraspnet,shao2024bimangrasp}.

Beyond grasping, Where2Act and VAT-Mart learn where and how to interact with articulated objects \cite{mo2021where2act,wu2022vatmart}, while GAPartNet provides generalizable actionable-part semantics and poses \cite{geng2023gapartnet}. O2O-Afford learns object--object affordances, OmniHang estimates stable hanging configurations and collision-free paths, and IndustReal develops simulation-to-real methods for contact-rich assembly \cite{mo2021o2oafford,you2021omnihang,tang2023industreal}. 3D AffordanceNet benchmarks 3D affordance understanding over a predefined affordance vocabulary \cite{deng2021affordancenet3d}. These systems provide strong supervision but are generally specialized by skill family, embodiment, contact model, or affordance vocabulary. AnnotateAnything instead uses one candidate-bank interface to organize grasping, articulation, placement, insertion, hanging, deformable manipulation, and navigation annotations while retaining skill-specific generation and validation.

\subsection{Large-Scale 3D Assets and Generative Asset Creation}

ShapeNet established large-scale CAD repositories, and PartNet introduced hierarchical part annotations \cite{chang2015shapenet,mo2019partnet}. PartNet-Mobility provides articulated objects with motion annotations and is integrated with the SAPIEN ecosystem \cite{partnetmobility2020,xiang2020sapien}. Objaverse further expanded the scale and diversity of web-sourced 3D objects \cite{deitke2023objaverse}. Manipulation-oriented resources add narrower forms of supervision: ACRONYM and DexGraspNet provide grasp annotations, and GAPartNet supplies actionable-part semantics and poses \cite{eppner2021acronym,wang2023dexgraspnet,geng2023gapartnet}.

Text-to-3D systems such as DreamFusion and Shap-E generate novel 3D assets from language \cite{poole2022dreamfusion,jun2023shapee}. However, general-purpose raw or generated assets typically do not include the complete manipulation metadata required by robot-learning systems, such as joint and part structure when applicable, functional parts, contact semantics, affordances, and executable trajectories. AnnotateAnything serves as a conversion layer from heterogeneous assets to interaction-ready supervision rather than as another asset repository or generative model.

\subsection{Simulation Environments and Manipulation Benchmarks}

The simulator ecosystem spans general-purpose rigid-body dynamics, articulated interaction, and large-scale GPU execution. MuJoCo and PyBullet are widely used physics engines \cite{todorov2012mujoco,benelot2018}; Isaac Gym enables GPU-resident parallel learning, and Isaac Sim supports richer industrial and photorealistic workflows \cite{makoviychuk2021isaacgym,nvidia2025isaacsim}. SAPIEN emphasizes articulated-object interaction, while RoboSuite provides a modular MuJoCo-based manipulation framework \cite{xiang2020sapien,zhu2020robosuite}.

Benchmark suites emphasize different capabilities: Meta-World targets multitask and meta-reinforcement learning, RLBench provides a large task library with generated demonstrations, ManiSkill and ManiSkill2 broaden object and manipulation coverage, and BEHAVIOR focuses on household and longer-horizon embodied activities \cite{yu2019metaworld,james2019rlbench,mu2021maniskill,gu2023maniskill2,srivastava2021behavior}. These simulators and benchmarks provide essential execution and evaluation infrastructure, but task authoring and annotation are commonly benchmark-specific. AnnotateAnything is complementary infrastructure intended to populate such environments with transferable, interaction-ready annotations across assets, embodiments, and skill families.

\subsection{Skill-Centric Robot Learning and Atomic Manipulation Skills}

The options framework formalizes temporally extended actions and hierarchical decision making \cite{sutton1999options}. Recent robot systems instantiate related ideas through explicit skills or callable APIs: SayCan selects predefined skills by combining language-model scores with affordance-based feasibility, Code as Policies generates programs that call perception and control APIs, and Composable Part-Based Manipulation composes manipulation constraints using object-part decomposition and part correspondences \cite{ahn2022saycan,liang2022codeaspolicies,liu2023cpm}.

These systems still rely on a skill or API inventory and on mechanisms that ground those skills to objects through contacts, parts, trajectories, constraints, or correspondences. AnnotateAnything addresses this grounding layer by automatically generating candidates that instantiate atomic skills, allowing skill libraries and hierarchical planners to extend beyond manually authored primitive sets.

\section{Limitations, Broader Impact, and Asset Licensing}
\label{app:governance}

\subsection{Limitations}

The system remains limited by its upstream models, observations, and simulation fidelity. Visual-language annotations can be incomplete or inaccurate when task-relevant parts are occluded, segmentation is coarse, materials are reflective, or geometry is ambiguous. Grounding is generally stronger for rigid and moderately articulated objects than for highly deformable assets, small contact-rich components, or severely cluttered scenes. Action validation depends on collision geometry, articulation metadata, contact modeling, and embodiment-specific planning; simulated success therefore does not guarantee real-world success. The current benchmark also does not cover every manipulation family, embodiment, or deployment constraint, and scaling to very large asset banks requires substantial compute and simulator infrastructure. Structured parsing and segmentation-identifier checks reject detectable malformed or ungrounded outputs, but they do not identify or correct every retained semantic error. Performance therefore remains dependent on the selected VLM and segmentation pipeline, especially for unusual objects, poor segmentations, and cluttered scenes. Physics and geometric validation can remove a measurable subset of inappropriate candidates but cannot guarantee semantic correctness. In addition, framework-adapted comparisons may partly reflect output-conversion fidelity, and the real-world study provides feasibility evidence rather than comparative physical superiority.

\subsection{Compute Resources}
\label{app:compute_resources}

Visual-language annotation runs on NVIDIA RTX A6000 GPUs with 48\,GB of GPU memory and includes VLM inference, multi-view processing, part, keypoint, and affordance grounding, and evaluation-sample generation. Physics validation and rollout collection run in headless Isaac Sim on NVIDIA RTX 4090 GPUs with 24\,GB of GPU memory; CPU workers handle asset loading, scene orchestration, logging, and asynchronous scheduling. Jobs are dispatched to available simulator workers according to asset complexity, embodiment, and skill type, so a scheduled job is not necessarily an independent Isaac Sim process.

The reported system-level throughput is measured under this heterogeneous configuration. Ann./min includes candidate generation, geometry filtering, optimization, physics validation, and annotation-bank serialization; Traj./h includes candidate retrieval, goal-set IK, collision-aware motion planning, full-robot execution, and trajectory validation. Floating-gripper grasp validation is generally cheaper than dexterous, bimanual, articulated, deformable, or navigation-related evaluation. The reported action-annotation and rollout throughput values are measured on the audited action-quality suite and the rollout evaluation in Sec.~\ref{app:exp_details}. The manually annotated visual-language subset is evaluated separately and is not included in the action or rollout throughput denominators. Random seeds control asset, skill, embodiment, initial pose, scene layout, camera, material, lighting, and disturbance sampling, but GPU-parallel asynchronous Isaac Sim execution is not assumed to be bitwise deterministic. The reported evaluation totals exclude exploratory development runs, failed preliminary experiments, and hyperparameter-search compute; throughput values in Tables~\ref{tab:physics_quality_and_collection} and~\ref{tab:app_rollout_full} should not be interpreted as hardware-independent constants.

\subsection{Broader Impact}

The main potential benefit is reducing the cost of converting raw 3D assets into reusable robot-learning supervision, enabling broader benchmarks, more diverse asset coverage, and less repeated manual simulation authoring. The same annotations can support quantitatively evaluated part, affordance, and keypoint pipelines, together with illustrative robot reasoning VQA and 3D VLM instruction-tuning repurposings.

The pipeline can also propagate errors from upstream models, asset metadata, or simulators into downstream policies. In addition, scalable simulation can make it easier to train large manipulation systems whose deployment behavior is not fully understood. Generated annotations should therefore be treated as useful but fallible supervision, accompanied by human auditing, task-specific validation metrics, and additional checks before real-world transfer.

\subsection{External Asset Sources and License Handling}

The processed annotation pool contains $17{,}005$ assets retained after normalization and successful simulator import. These counts describe processed assets rather than the raw sizes of the upstream repositories. For every retained asset, we retain the upstream source family, asset type, upstream identifier when available, and any available license or redistribution metadata.

\begin{table*}[t]
\centering
\caption{
External asset sources, exact retained counts, and release handling based on applicable upstream terms and available per-asset metadata. Raw assets are redistributed only when permitted; otherwise, we release derived annotations, source identifiers, and acquisition instructions.
}
\label{tab:app_source_license_inventory}
\scriptsize
\setlength{\tabcolsep}{3pt}
\renewcommand{\arraystretch}{1.08}
\begin{tabularx}{\textwidth}{
p{0.16\textwidth}
p{0.18\textwidth}
p{0.09\textwidth}
X}
\toprule
Asset family & Source & Retained assets & Release handling \\
\midrule

Rigid
& Objaverse~\cite{deitke2023objaverse}
& 5{,}000
& Follow collection-level and per-object license metadata; preserve source identifiers and required attribution. \\

Rigid
& ShapeNet~\cite{chang2015shapenet}
& 4{,}500
& Release derived annotations and source mappings rather than restricted raw meshes. \\

Rigid
& OmniObject3D
& 3{,}038
& Preserve upstream identifiers and attribution under the applicable upstream terms. \\

Articulated
& PartNet-Mobility~\cite{xiang2020sapien,partnetmobility2020}
& 800
& Do not redistribute restricted raw geometry or articulation files; release derived annotations and source mappings. \\

Articulated
& PhysX-Mobility
& 1{,}294
& Preserve applicable attribution and non-commercial restrictions. \\

Deformable / garment
& GarmentLab~\cite{lu2024garmentlab}
& 1{,}500
& Release derived annotations and source mappings; do not assume blanket permission to redistribute third-party assets. \\

Deformable / garment
& ClothesNet~\cite{zhou2023clothesnetinformationrich3dgarment}
& 667
& Release derived annotations and source mappings; users obtain restricted raw assets from the upstream provider. \\

Room-scale
& AgentWorld
& 114
& Release derived annotations and source identifiers; do not assume blanket redistribution permission for underlying scene assets. \\

Room-scale
& GRScenes
& 92
& Preserve applicable attribution, non-commercial, and share-alike restrictions for redistributable material. \\

\midrule
\textbf{Total}
& \textbf{9 source families}
& \textbf{17{,}005}
& -- \\
\bottomrule
\end{tabularx}
\end{table*}

\paragraph{Filtering and exclusions.}
We apply common source-independent preprocessing before annotation. Assets are excluded when they contain corrupt, empty, or unreadable geometry; unresolved dependencies that prevent rendering or simulator import; unrecoverable unit, scale, orientation, or coordinate-frame errors; unusable collision geometry; invalid articulation axes, links, limits, or moving-part metadata; garment geometry that cannot be initialized stably; or scene packages with invalid references, unusable layouts, or no valid navigable and interaction-ready region. Skill-specific incompatibility does not necessarily remove an asset from the pool: for example, an object unsuitable for insertion may remain valid for grasping. Evaluation is therefore defined over valid asset--skill pairs.

\paragraph{Deduplication and rejection accounting.}
Repeated imports sharing the same upstream identifier or ingestion record are collapsed when this information is available. We do not perform exhaustive cross-source geometric nearest-neighbor deduplication and therefore do not claim that the processed pool is free of all near-duplicates; visually or geometrically related assets may still differ in scale, topology, collision geometry, articulation, or applicable manipulation annotations. Because failed downloads and imports were not logged uniformly across all upstream importers, we report exact retained processed counts and explicit exclusion criteria rather than estimating incomplete source-level rejection totals.

\paragraph{Release policy.}
For every retained asset, the planned public package records the source family, upstream identifier when available, asset type, applicable skills, conversion status, and available license or redistribution metadata; unresolved cases are marked explicitly rather than assigned an inferred license. Raw third-party assets are redistributed only when the applicable upstream terms permit it. Otherwise, the release contains derived annotations, source mappings, and instructions for obtaining the upstream assets directly from their providers. The construction of the audited suite and manual-reference subset is described separately in Appendix~H.1.

\subsection{Artifacts and Documentation}

All artifacts supporting the reported results are complete and included in the anonymous supplementary materials. These include the paper and detailed method appendix, qualitative videos, redistributable demo assets, a code snapshot covering the principal annotation, simulation data-generation, and full-robot execution paths, and representative language, visual, and action annotation packages with asset metadata.

The remaining work concerns public-release packaging rather than completion of the reported experiments. The planned public release will provide a cleaned and documented codebase, the packaged full-scale derived annotation collection, fixed manifests for the retained pool, audited asset--skill suite, and manual-reference subset, rollout metadata, evaluation and uncertainty scripts, prompts and schemas, and complete provenance and license mappings. Restricted raw third-party meshes are neither included in the anonymous supplement nor planned for redistribution; users will instead receive source identifiers and instructions for obtaining and processing the upstream assets.

Language prompts and schemas are documented in Sec.~\ref{supp:asset_language_annotation} and the remaining language-annotation subsections, visual grounding in Sec.~\ref{app:visual_annotation}, and action generation and validation in Sec.~\ref{app:action_annotation}. The corresponding public manifests will link these derived annotations and rollout metadata to their provenance and redistribution fields.

\end{document}